\documentclass{article}

\usepackage{arxiv}

\usepackage[utf8]{inputenc} 
\usepackage[T1]{fontenc}    
\usepackage{hyperref}       
\usepackage{url}            
\usepackage{booktabs}       
\usepackage{amsfonts}       
\usepackage{nicefrac}       
\usepackage{microtype}      
\usepackage{lipsum}

\usepackage{natbib} 
\usepackage{mathtools} 
\usepackage{amsthm}
\usepackage{amssymb}
\usepackage{booktabs} 
\usepackage{tikz} 
\usepackage{cancel}
\usepackage{graphicx}

\usepackage{algorithm}
\usepackage[noend]{algpseudocode}

\def\ci{\perp\!\!\!\perp}

\newcommand{\M}{\mathbf{M}}
\newcommand{\Opt}{\mathbf{O}}
\newcommand{\Optcmin}{\mathbf{O}_{\rm Cmin}}
\newcommand{\Optmin}{\mathbf{O}_{\rm min}}
\newcommand{\Optnpi}{\mathbf{O}_{\rm \pi^N_i}}
\newcommand{\ECol}{E_{\mathbf{C}}}
\newcommand{\Colmin}{\mathbf{C}_{\rm min}}
\newcommand{\Par}{\mathbf{P}}
\newcommand{\Col}{\mathbf{C}}

\newcommand{\Sel}{\mathbf{S}}
\newcommand{\X}{X} 
\newcommand{\x}{x} 
\newcommand{\Y}{Y}

\newcommand{\Z}{\mathbf{Z}}
\newcommand{\B}{\mathbf{B}}
\newcommand{\A}{\mathbf{A}}

\newcommand{\E}{\mathbf{E}}
\newcommand{\R}{\mathbf{R}}
\newcommand{\V}{\mathbf{V}}

\newcommand{\Gra}{\mathcal{G}}
\newcommand{\pa}{pa}
\newcommand{\ch}{ch}
\newcommand{\an}{an}
\newcommand{\ancs}{\mathbf{vancs}}
\newcommand{\Adj}{{\ancs}}
\newcommand{\des}{des}

\newcommand{\forb}{\mathbf{forb}}
\newcommand{\spouse}{sp}
\newcommand{\Prob}{\mathcal{P}} 

\newcommand{\tailhead}{{\rightarrow}}
\newcommand{\headtail}{{\leftarrow}}
\newcommand{\headhead}{{\leftrightarrow}}

\newcommand{\astast}{{\ast\!{\--}\!\ast}}
\newcommand{\asthead}{{\ast\!{\rightarrow}}}
\newcommand{\headast}{{{\leftarrow}\!\ast}}

\newtheorem{myassum}{Assumptions}
\newtheorem{mydef}{Definition}
\newtheorem{mythm}{Theorem}
\newtheorem{myprop}{Proposition}
\newtheorem{mycor}{Corollary}
\newtheorem{mylemma}{Lemma}
\theoremstyle{definition}
\newtheorem*{mythm*}{Theorem}
\newtheorem*{mylemma*}{Lemma}
\newtheorem*{mycor*}{Corollary}
\newtheorem*{myprop*}{Proposition}

\title{Necessary and sufficient graphical conditions for optimal adjustment sets in causal graphical models with hidden variables}

\author{ {\bf Jakob Runge} \\
German Aerospace Center \\
Institute of Data Science\\
07745 Jena, Germany \\
and\\
Technische Universit\"at Berlin \\
10623 Berlin, Germany \\
}

\begin{document}
\maketitle

\begin{abstract}
The problem of selecting optimal backdoor adjustment sets to estimate causal effects in graphical models with hidden and conditioned variables is addressed. Previous work has defined optimality as achieving the smallest asymptotic estimation variance and derived an optimal set for the case without hidden variables. For the case with hidden variables there can be settings where no optimal set exists and currently only a sufficient graphical optimality criterion of limited applicability has been derived. In the present work optimality is characterized as maximizing a certain adjustment information which allows to derive a necessary and sufficient graphical criterion for the existence of an optimal adjustment set and a definition and algorithm to construct it. Further, the optimal set is valid if and only if a valid adjustment set exists and has higher (or equal) adjustment information than the Adjust-set proposed in Perkovi{\'c} et~al. [Journal of Machine Learning Research, 18: 1--62, 2018] for any graph. The results translate to minimal asymptotic estimation variance for a class of estimators whose asymptotic variance follows a certain information-theoretic relation. Numerical experiments indicate that the asymptotic results also hold for relatively small sample sizes and that the optimal adjustment set or minimized variants thereof often yield better variance also beyond that estimator class. Surprisingly, among the randomly created setups more than 90\% fulfill the optimality conditions indicating that also in many real-world scenarios graphical optimality may hold.
Code is available as part of the python package \url{https://github.com/jakobrunge/tigramite}.
\end{abstract}


\section{Introduction}

A standard problem setting in causal inference is to estimate the causal effect between two variables given a causal graphical model that specifies qualitative causal relations among observed  variables \citep{Pearl2000}, including a possible presence of hidden confounding variables. 
The graphical model then allows to employ graphical criteria to identify valid adjustment sets, the most well-known being the \emph{backdoor criterion} \citep{pearl1993} and the \emph{generalized adjustment criterion}  \citep{shpitser2010validity,perkovic2015complete,perkovic2018complete}, providing a complete identification of all valid adjustment sets. Estimators of causal effects based on such a valid adjustment set as a covariate are then unbiased, but for different adjustment sets the estimation variance may strongly vary. An \emph{optimal adjustment set} may be characterized as one that has minimal asymptotic estimation variance. In \textbf{current work}, following 
\citet{kuroki2004selection} and \citet{kuroki2003covariate}, \citet{henckel2019graphical} (abbreviated HPM19 in the following) showed that graphical optimality always holds for linear models in the causally sufficient case where all relevant variables are observed. In \citet{witte2020efficient} an alternative characterization of the optimal adjustment set is discussed and the approach was integrated into the IDA algorithm \citep{maathuis2009estimating,Maathuis2010} that does not require the causal graph to be known.
\citet{rotnitzky2019efficient} extended the results in HPM19 to asymptotically linear non-parametric graphical models. 
HPM19's optimal adjustment set holds for the causally sufficient case (no hidden variables) and the authors  gave an example  with hidden variables where optimality does not hold in general, i.e., the optimal adjustment set depends on the coefficients and noise terms (more generally, the distribution), rather than just the graph. 
Most recently, \citet{smucler2020efficient} (SSR20) partially extended these results to the non-parametric hidden variables case together with \emph{dynamic treatment regimes}, i.e., conditional causal effects. SSR20 provide a sufficient criterion for an optimal set to exist and a definition based on a certain undirected graph-construction using a result by \citet{van2019separators}. 
However, their sufficient criterion is very restrictive and a current major open problem is a \emph{necessary} and sufficient condition for an optimal adjustment set to exist in the hidden variable case and a corresponding definition of an optimal set.

My \textbf{main theoretical contribution} is a solution to this problem. Optimality for conditional causal effects in the hidden variables case is fully characterized by an information-theoretic approach involving a certain difference of conditional mutual informations among the observed variables termed the adjustment information. Maximizing the adjustment information formalizes the common intuition to choose adjustment sets that maximally constrain the effect variable and minimally constrain the cause variable. This allows to derive a necessary and sufficient graphical criterion for the existence of an optimal adjustment set. The derived optimal adjustment set also has the property of minimum cardinality, i.e., no node can be removed without sacrificing optimality. Further, the optimal set is valid if and only if a valid adjustment set exists and has higher (or equal) adjustment information than the Adjust-set proposed in \citet{perkovic2018complete} for any graph, whether graphical optimality holds or not.
The results translate to minimal asymptotic estimation variance for a class of estimators whose asymptotic variance follows a certain information-theoretic relation that, at present, I could only verify theoretically for the linear case.
As \textbf{practical contributions} the paper provides extensive numerical experiments that corroborate the theoretical results and show that the optimal adjustment set or minimized variants thereof often yield better variance also beyond the theoretically analyzed estimator class. Code is available in the python package \url{https://github.com/jakobrunge/tigramite}. More detailed preliminaries, proofs, algorithms, and further numerical experiments are given in the Supplementary Material. 

\subsection{Preliminaries and problem setting}
\begin{figure}[t]  
\centering
\includegraphics[width=1.\linewidth]{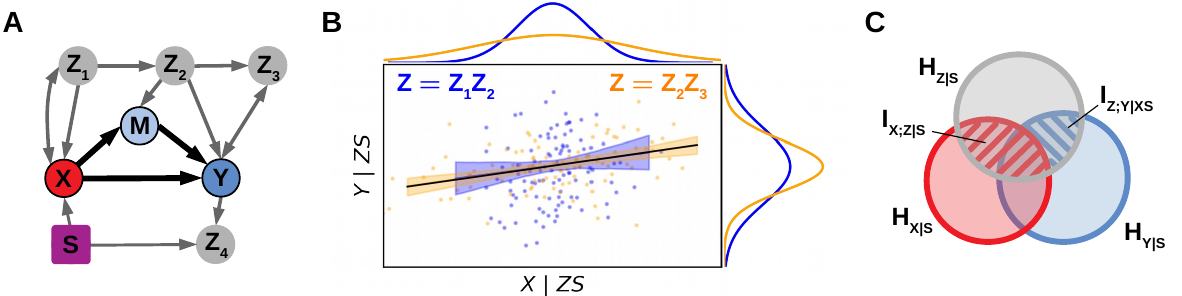}%
\caption{(\textbf{A}) Problem setting of optimal adjustment sets in causal graphs with hidden variables represented through bi-directed edges. 
The goal is to estimate the total causal effect of $\X$ on $\Y$ potentially through mediators $\M$, and given conditioned variables $\Sel$. The task is to select a valid adjustment set $\Z$ such that the estimator has minimal asymptotic variance. (\textbf{B}) illustrates two causal effect estimates for a linear Gaussian model consistent with the graph in \textbf{A}, see discussion in text. (\textbf{C}) For a certain class of estimators a minimal asymptotic estimation variance can be translated into an information-theoretical optimization problem, here visualized in a Venn diagram. An optimal adjustment set $\Z$ must maximize the adjustment information $I_{\Z;\Y|\X\Sel}-I_{\X;\Z|\Sel}$ (blue and red hatched, respectively).}
\label{fig:motivation_venn}
\end{figure}

We consider causal effects in causal graphical models over a set of variables $\V$  with a joint distribution $\Prob=\Prob(\V)$ that is consistent with an acyclic  directed  mixed  graph (ADMG) $\Gra=(\V,\mathcal{E})$. Two nodes can have possibly more than one edge which can be \emph{directed} ($\headtail$) or \emph{bi-directed} ($\headhead$). See Fig.~\ref{fig:motivation_venn}A for an example. Kinships are defined as usual: parents $\pa(X)$ for ``$\bullet\tailhead X$'', spouses $\spouse(X)$  for ``$X\headhead \bullet$'', children $\ch(X)$ for ``$X\tailhead\bullet$''. These sets all exclude $X$. Correspondingly descendants $\des(X)$ and ancestors $\an(X)$ are defined, which, on the other hand, both include $X$. The mediator nodes on causal paths from $X$ to $Y$ are denoted $\M=\M(X, Y)$ and exclude $X$ and $Y$. For detailed preliminaries, including the definition of open and blocked paths, see Supplementary Section~\ref{sec:setting}.
In this work we only consider a univariate intervention variable $\X$ and effect variable $Y$. We simplify set notation and denote unions of variables as $\{W\}\cup \M \cup \A=W\M\A$.

A (possibly empty) set of adjustment variables $\Z$ for the total causal effect of $\X$ on $\Y$ in an ADMG is called \emph{valid} relative to ($\X,\Y$) if the interventional distribution for setting $do(\X=\mathbf{x})$ \citep{Pearl2000} factorizes as $p(\Y|do(\X=\mathbf{x})) = \int p(\Y|\mathbf{x},\mathbf{z}) p(\mathbf{z}) d\mathbf{z}$ for non-empty $\Z$ and as $ p(\Y|do(\X=\mathbf{x})) = p(\Y|\mathbf{x})$ for empty $\Z$. Valid  adjustment sets, the set of which is here denoted $\mathcal{Z}$,  can  be  read  off  from  a  given  causal graph using the generalized  adjustment  criterion \citep{perkovic2015complete,perkovic2018complete} which generalizes Pearl's back-door criterion \citep{Pearl2000}. To this end define 
\begin{align} \label{eq:forb}
\forb(\X,\Y)=\X\cup\des(\Y\M)
\end{align}
(henceforth just denoted as $\forb$). A set $\Z$ is valid if both of the following conditions hold: 
(i) $\Z\cap \forb=\emptyset$, and (ii)~all non-causal paths  from $\X$ to $\Y$ are blocked by $\Z$. An adjustment set is called \emph{minimal} if no strict subset of $\Z$ is still valid. 
The validity conditions can in principle be manually checked directly from the graph, but, more conveniently, \citet{perkovic2018complete} define an adjustment set called `Adjust' that is valid if and only if a valid adjustment set exist. In our setting including conditioning variables $\Sel$ we call this set the \emph{valid ancestors} defined as
\begin{align} \label{eq:adjust}
 \ancs(\X,\Y,\Sel) &= \an(\X\Y\Sel)\setminus \forb\,
\end{align}
and refer to this set as $\ancs$ or Adjust-set.

Our quantity of interest is the average total causal effect of an intervention to set $\X$ to $x$ vs. $x'$ on the effect variable $\Y$ given a set of selected (conditioned) variables $\Sel=\mathbf{s}$
\begin{align} \label{eq:causal_effect}
 \Delta_{yxx'|\mathbf{s}} = E(\Y|do(x),\mathbf{s}) - E(\Y|do(x'),\mathbf{s})\,.
\end{align}
We denote an estimator given a valid adjustment set $\Z$ as $\widehat{\Delta}_{yxx'|\mathbf{s}.\mathbf{z}}$. 
In the linear case $\Delta_{yxx'|\mathbf{s}}$ for $x=x'+1$ corresponds to the regression coefficient $\beta_{\Y\X\cdot \Z \Sel}$ in the regression of $Y$ on $X$, $\Z$, and $\Sel$. The ordinary least squares (OLS) estimator $\hat{\beta}_{\Y\X\cdot \Z \Sel}$ is a consistent estimator of $\beta_{\Y\X\cdot \Z \Sel}$. 

Figure~\ref{fig:motivation_venn}A illustrates the \textbf{problem setting}: We are interested in the total causal effect of (here univariate) $\X$ on $\Y$ (conditioned on $\Sel$), which is here due to a direct link and an indirect causal path through a mediator $M$. There are six valid backdoor adjustment sets $\mathcal{Z}=\{Z_1,~Z_2,~Z_1Z_2,~Z_2Z_3,~Z_1Z_3,~Z_1Z_2Z_3\}$. $Z_4\in\forb$ cannot be included in any set because it is a descendant of $\Y\M$. Here $\ancs=Z_1Z_2S$. All valid adjustment sets remove the bias due to confounding by their definition. The question is which of these valid adjustment sets is statistically optimal in that it minimizes the asymptotic estimation variance?
More formally, the task is, given a graph $\Gra$ and $(\X,\Y,\Sel)$, to chose a valid optimal set $\Z_{\rm optimal}\in\mathcal{Z}$ such that the causal effect estimator's asymptotic variance $\text{Var}(\widehat{\Delta}_{yxx'|\mathbf{s}.\mathbf{z}})=E[(\Delta_{y\x\x'|\mathbf{s}} - \widehat{\Delta}_{y\x\x'|\mathbf{s}.\mathbf{z}})^2]$ is minimal:
\begin{align}
\Z_{\rm optimal}\in {\rm argmin}_{\Z \in \mathcal{Z} } \text{Var}(\widehat{\Delta}_{yxx'|\mathbf{s}.\mathbf{z}})\,.
\end{align}


My proposed approach to optimal adjustment sets is based on information theory \citep{Cover2006}. The main quantity of interest there is the conditional mutual information (CMI) defined as a difference $I_{X;Y|Z} = H_{Y|Z} - H_{Y|Z X}$ of two (conditional) Shannon entropies $H_{Y|X}=-\int_{x,y} p(x,y) \ln p(y|x)dxdy$.
Its main properties are non-negativity, $I_{X;Y|Z}=0$ if and only if $X \ci Y | Z$, and the chain rule $I_{XW;Y|Z}=I_{X;Y|Z}+I_{W;Y|Z X}$. All random variables in a CMI can be multivariate.

Throughout the present paper we will assume the following.

\begin{myassum}[General setting and assumptions] \label{assum:opt}
We assume a causal graphical model over a set of variables $\V$ with a joint distribution $\Prob=\Prob(\V)$ that is consistent with an ADMG $\Gra=(\V,\mathcal{E})$. We assume a non-zero causal effect from $\X$ on $\Y$, potentially through a set of mediators $\M$, and given selected conditioned variables $\Sel$, where $\Sel\cap \forb=\emptyset$. 
We assume that at least one valid adjustment set (given $\Sel$) exists and, hence, the causal effect is identifiable (except when stated otherwise). 
Finally, we assume the usual Causal Markov Condition (implicit in semi-Markovian models) and Faithfulness.
\end{myassum}

\section{Optimal adjustment sets}

\subsection{Information-theoretic characterization}
Figure~\ref{fig:motivation_venn}B illustrates two causal effect estimates for a linear Gaussian model consistent with the graph in Fig.~\ref{fig:motivation_venn}A. With $\Z=Z_1Z_2$ (blue) the error is much larger than with $\Opt=Z_2Z_3$ (orange) for two reasons: $\Z$ constrains the residual variance $Var(Y|\Z\Sel)$ of the effect variable $\Y$ less than $\Opt$ and, on the other hand, $\Z$ constrains the residual variance $Var(X|\Z\Sel)$ of the cause variable $\X$ more than $\Opt$. Smaller estimator variance also holds for $\Opt$ compared to any other valid set in $\mathcal{Z}$ here.

We information-theoretically formalize the resulting intuition to choose an adjustment set $\Z$ that maximally constrains the effect variable $\Y$ and minimally constrains the cause variable $\X$. In terms of CMIs and given selected fixed conditions $\Sel$ the quantity to maximize can be stated as follows.
\begin{mydef}[Adjustment information] \label{def:info_cmi}
Consider a causal effect of $\X$ on $\Y$ for an adjustment set $\Z$ given a condition set $\Sel$. The \emph{(conditional) adjustment (set) information}, abbreviated $J_\Z$, is defined as
\begin{align} 
J_{\X\Y|\Sel.\Z} &\equiv I_{\Z;\Y|\X\Sel}- I_{\X;\Z|\Sel} \label{eq:info_cmi} \\
&= \underbrace{H_{\Y|\X\Sel} - H_{\X|\Sel}}_{\text{not related to $\Z$}} - \underbrace{(H_{\Y|\X \Z \Sel} - H_{\X|\Z \Sel})}_{\text{adjustment entropy}}  \label{eq:jz}
\end{align}
\end{mydef}
$J_\Z$ is not necessarily positive if the dependence between $\X$ and $\Z$ (given $\Sel$) is larger than that between $\Z$ and $\Y$ given $\X\Sel$. Equation~\eqref{eq:jz} follows from the CMI definition.
Fig.~\ref{fig:motivation_venn}C illustrates the two CMIs in Eq.~\eqref{eq:info_cmi} in a Venn diagram.

Before discussing the range of estimators for which  maximizing the adjustment information $J_{\Z}$ leads to a minimal asymptotic estimation variance in Sect.~\ref{sec:estimator_class}, we characterize graphical optimality in an information-theoretic framework. Our goal is to provide graphical criteria for optimal adjustment sets, i.e., criteria that depend only on the structure of the graph $\Gra$ and not on the distribution.

\begin{mydef}[Information-theoretical graphical optimality] \label{def:def_gra_opt}
 Given Assumptions~\ref{assum:opt} we say that \emph{(information-theoretical) graphical optimality holds} if there is a $\Z\in\mathcal{Z}$ such that either there is no other $\Z'\neq \Z\in\mathcal{Z}$ or for all other $\Z'\neq\Z\in\mathcal{Z}$  and \emph{all} distributions $\mathcal{P}$ consistent with $\Gra$ we have $J_\Z \geq J_{\Z'}$.
\end{mydef}

My main result builds on the following lemma which relates graphical optimality to information-theoretic inequalities in a necessary and sufficient comparison condition for an optimal set to exist.

\begin{mylemma}[Necessary and sufficient comparison criterion for existence of an optimal set] \label{thm:existence}
Given Assumptions~\ref{assum:opt}, if and only if there is a $\Z\in\mathcal{Z}$ such that either there is no other $\Z'\neq \Z\in\mathcal{Z}$ or for all other $\Z'\neq\Z\in\mathcal{Z}$ and all distributions $\Prob$ consistent with $\Gra$ it holds that
\begin{align} \label{eq:decomp_zz}
\underbrace{I_{\Z\setminus\Z';\Y|\Z' \X\Sel}}_{\text{(i)}} \geq \underbrace{I_{\Z'\setminus\Z;\Y|\Z \X \Sel}}_{\text{(iii)}}~~~\text{and} ~~~
\underbrace{I_{\X;\Z'\setminus\Z|\Z \Sel}}_{\text{(ii)}}  \geq \underbrace{I_{\X;\Z\setminus\Z'|\Z' \Sel}}_{\text{(iv)}}\,,
\end{align}
then graphical optimality holds and $\Z$ is optimal implying $J_\Z \geq J_{\Z'}$.
\end{mylemma}

In SSR20 and HPM19 the corresponding conditional independence statements to the terms~(iii) and (iv) in the inequalities~\eqref{eq:decomp_zz} are used as a sufficient pairwise comparison criterion. However, Lemma~\ref{thm:existence} shows that for graphical optimality it is not necessary that terms~(iii) and (iv) vanish, they just need to fulfill the inequalities~\eqref{eq:decomp_zz} for a \emph{necessary} and sufficient criterion. 

In principle, Lemma~\ref{thm:existence} can be used to cross-compare all pairs of sets, but firstly, it is difficult to explicitly evaluate~\eqref{eq:decomp_zz} for all distributions $\Prob$ consistent with $\Gra$ and, secondly, iterating through all valid adjustment sets is computationally prohibitive even for small graph sizes. As an example, consider a confounding path consisting of $5$ nodes. Then this path can be blocked by $2^5-1$ different subsets.
In the main result of this work (Thm.~\ref{thm:graph_optimality}) a necessary and sufficient criterion based purely on graphical properties is given.


\subsection{Applicable estimator class} \label{sec:estimator_class}
The above characterization only relates optimality of adjustment sets to the adjustment information $J_{\Z}$ defined in Eq.~\eqref{eq:info_cmi}, but not to any particular estimator. 
Now the question is for which class of causal effect estimators $\widehat{\Delta}_{yxx'|\mathbf{s}.\mathbf{z}}$ the intuition of maximizing the adjustment information $J_{\Z}$ leads to a minimal asymptotic estimation variance. In its most general form this class is characterized as fulfilling
\begin{align} \label{eq:maxmin}
\Z_{\rm optimal} \in {\rm argmax}_{\Z \in \mathcal{Z} } J_{\Z}~\Leftrightarrow~\text{Var}(\widehat{\Delta}_{yxx'|\mathbf{s}.\mathbf{z}_{\rm optimal}})=\min_{\Z \in \mathcal{Z}}\text{Var}(\widehat{\Delta}_{yxx'|\mathbf{s}.\mathbf{z}})\,,
\end{align}
where we assume that $\widehat{\Delta}_{yxx'|\mathbf{s}.\mathbf{z}}$ is consistent due to a valid adjustment set and correct functional model specification.  
One can also further restrict the class to estimators whose (square-root of the) asymptotic variance can be expressed as
\begin{align} \label{eq:entropy_est}
\sqrt{\text{Var}(\widehat{\Delta}_{yxx'|\mathbf{s}.\mathbf{z}})} &= f(H_{\Y|\X \Z \Sel} - H_{\X|\Z \Sel})\,, 
\end{align}
for a real-valued, strictly monotonously increasing function of the adjustment entropy. Minimizing the adjustment entropy is by Eq.~\eqref{eq:jz} equivalent to maximizing the adjustment information. The following assumption and lemma then relates $J_\Z \geq J_{\Z'}$ to the corresponding asymptotic variances of a given estimator. 

\begin{myassum}[Estimator class assumption] \label{assum:estimator}
The model class of the estimator for the causal effect~\eqref{eq:causal_effect} is correctly specified and its asymptotic variance can be expressed as in relation~\eqref{eq:entropy_est}. 
\end{myassum}

\begin{mylemma}[Asymptotic variance and adjustment information] \label{thm:variance}
Given Assumptions~\ref{assum:opt} and an estimator fulfilling Assumptions~\ref{assum:estimator}, if and only if for two different adjustment sets $\Z,\Z'\in\mathcal{Z}$ we have $J_\Z \geq J_{\Z'}$, then the adjustment set $\Z$ has a smaller or equal asymptotic variance compared to $\Z'$.
\end{mylemma}
\textit{Proof.}
By Equations~\eqref{eq:jz} and \eqref{eq:entropy_est} $J_\Z \geq J_{\Z'}$ (for fixed $\X,\Y,\Sel$) is directly related to a smaller or equal asymptotic variance for $\Z$ compared to $\Z'$, and vice versa.
\hfill $\square$

The paper's theoretical results currently hold for estimators fulfilling relation~\eqref{eq:entropy_est}, but at least the main result on graphical optimality in Thm.~\ref{thm:graph_optimality}  can also be relaxed to estimators fulfilling the less restrictive relation~\eqref{eq:maxmin}. In this work, we leave the question of which general classes of estimators fulfill either relation~\eqref{eq:maxmin} or the more restricted relation~\eqref{eq:entropy_est} to further research and only show that it holds for the OLS estimator $\widehat{\beta}_{\Y X\cdot \Z \Sel}$ for Gaussian distributions. 

For Gaussians the entropies in \eqref{eq:entropy_est} are given by $H(Y|X\Z\Sel)=\frac{1}{2}+ \frac{1}{2}\ln(2\pi\sigma^2_{Y|X\Z\Sel})$ and $H(X|\Z\Sel)=\frac{1}{2}+ \frac{1}{2}\ln(2\pi\sigma^2_{X|\Z\Sel})$ where $\sigma(\cdot|\cdot)$ denotes the square-root of the conditional variance. Then
\begin{align} \label{eq:entropy_est_lin}
\sqrt{\text{Var}(\widehat{\Delta}_{yxx'|\mathbf{s}.\mathbf{z}})} &= \frac{1}{\sqrt{n}} e^{H_{\Y|\X \Z \Sel} - H_{\X|\Z \Sel}} = \frac{1}{\sqrt{n}} \frac{\sigma_{\Y|\X \Z \Sel}}{\sigma_{\X|\Z \Sel}}\,.
\end{align}
This relation is also the basis of the results for the causally sufficient case in \citet{henckel2019graphical} where it is shown that it holds more generally for causal linear models that do not require the noise terms to be Gaussian.


\subsection{Definition of O-set} \label{sec:Oset_def}

The optimal adjustment set for the causally sufficient case is simply $\Par = \pa(\Y\M) \setminus \forb$ and was derived in HPM19 and \citet{rotnitzky2019efficient}. In Section~\ref{sec:suff} the derivation is discussed from an information-theoretic perspective.
In the case with hidden variables we need to account for bidirected edges ``$\headhead$'' which considerably complicate the situation. Then the parents of $\Y\M$ are not sufficient to block all non-causal paths. Further, just like conditioning on parents of $\Y\M$ leads to optimality in the sufficient case since parents constrain information in $\Y\M$, in the hidden variables case also conditioning on spouses of $\Y\M$ constrains information about $\Y\M$.

\textbf{Example A.} A simple graph (ADMG) to illustrate this is $X\tailhead Y \headhead Z_1$ (shown with an additional $\Sel$ in Fig.~\ref{fig:examples}A below, or Fig. 4 in SSR20). Here $\Z'=\emptyset=\ancs$ is a valid set, but it is not optimal. Consider $\Opt=Z_1$, then term~(iii) $=0$ in the inequalities~\eqref{eq:decomp_zz} since $\Z'\setminus\Opt=\emptyset$. 
Even though not needed to block non-causal paths (there is none), $Z_1$ still constrains information in $Y$ while being independent of $X$ (hence, term~(iv) $=0$) which leads to $J_{\Opt}>J_\emptyset$ according to the inequalities~\eqref{eq:decomp_zz}.

Not only direct spouses can constrain information in $Y$ as Fig.~\ref{fig:examples}B below illustrates.
Since for $W\in \Y\M$ the motif ``$W\headhead \boxed{C_1} \headast C_2$'' (``$\ast$'' denotes either edge mark) is open, it holds that $I(C_1C_2;W) = I(C_1;Y)+I(C_2;W|C_1)\geq  I(C_1;Y)$ and we can even further increase the first term in the adjustment information by conditioning also on subsequent spouses. This chain of colliders only ends if we reach a tail or there is no further adjacency. However, we have to make sure that conditioning on colliders does not open non-causal paths. This leads to the notion of a \emph{valid collider path} (related to the notion of a \emph{district} in  \citet{evans2014markovian}).

\begin{mydef}[Valid collider paths] \label{def:def_collpath}
Given a graph $\Gra$, a \emph{collider path} of $W$ for $k\geq 1$ is defined by a sequence of edges $W \headhead C_{1}\headhead \cdots \headhead C_k$. We denote the set of path nodes (excluding $W$) along a path indexed by $i$ as $\pi^i_W$.
Using the set of valid ancestors $\ancs=\an(\X\Y\Sel) \setminus \forb$ for the causal effect of $\X$ on $\Y$ given $\Sel$ we call a collider path node set $\pi^i_W$ for $W\in\Y\M$ \emph{valid} wrt. to  $(\X,\Y,\Sel)$ if for each path node $C\in \pi^i_W$ both of the following conditions are fulfilled:
\begin{align}
(1)~C\notin \forb, ~~~\text{and}~~~ (2a)~C \in \ancs ~\text{or}~ (2b)~C \ci \X ~|~\ancs\,.
\end{align}
\end{mydef}

Condition~(1) is required for any valid adjustment set. If jointly (2a) and (2b) are not fulfilled, i.e. $C \notin \ancs$ and $C \cancel{\ci} \X ~|~\ancs$, then the collider path stops before $C$. 
Our candidate optimal adjustment set is now constructed based on the parents of $\Y\M$, valid collider path nodes of $\Y\M$, and their parents to `close' these collider paths.

\begin{mydef}[O-set] \label{def:def_opt}
Given Assumptions~\ref{assum:opt} and the definition of valid colliders in Def.~\ref{def:def_collpath}, define the set $\Opt(\X,\Y,\Sel) = \Par \cup \Col \cup \Par_\Col$ where
\begin{align*}
   \Par = \pa(\Y\M) \setminus \forb,~~
   \Col = \Cup_{W\in \Y\M} \Cup_{i} \{\pi^i_W:~ \text{$\pi^i_W$ is valid wrt. to $(\X,\Y,\Sel)$} \},~~
   \Par_\Col &= \pa( \Col)  \,.
\end{align*}
\end{mydef}

In the following we will abbreviate $\Opt=\Opt(\X,\Y,\Sel)$.
Algorithm~\ref{algo:opt} states efficient pseudo-code to construct the $\Opt$-set and detect whether a valid adjustment set exists. Since none of the conditions of Def.~\ref{def:def_collpath} for adding collider nodes depends on previously added nodes, the algorithm is order-independent. The statement occurring in lines 11 and 21 (``No valid adjustment set exists.'') is proven in Thm.~\ref{thm:validity}. If the graph is a DAG, then lines 4-22 can be omitted. 
The algorithm is of low complexity and the most time-consuming part is checking for a path in line 12, Def.~\ref{def:def_collpath}(2b)~$C \ci \X ~|~\ancs$, which can be implemented with (bi-directional) breadth-first search as proposed in \citet{van2019separators}.

Numerical experiments in Section~\ref{sec:numerics} will show that further interesting adjustment sets are the \emph{minimized} $\Opt$-set $\Opt_{\rm min}$, where $\Opt$ is minimized such that no subset can be removed without making $\Opt_{\rm min}$ invalid, and the  \emph{collider-minimized} $\Opt$-set $\Opt_{\rm Cmin}$ where only $\Col\Par_\Col\setminus\Par\subseteq\Opt$ is minimized such that no collider-subset can be removed without making $\Opt_{\rm Cmin}$ invalid. Both adjustment sets can be constructed with Alg.~\ref{algo:optmin} similar to the efficient algorithms in \citet{van2019separators}. Also the minimized sets are order-independent since the nodes are removed only after the for-loops.
Based on the idea in $\Opt_{\rm Cmin}$, in the numerical experiments we also consider Adjust$_{\rm Xmin}$, where only ${\rm Adjust}\setminus\pa(Y\M)$ is minimized and $\pa(Y\M)$ is always included. Finally, we also evaluate Adjust$_{\rm min}$ where Adjust is fully minimized.


Before discussing the optimality of the $\Opt$-set, we need to assure that it is a valid adjustment set. Similar to the proof given in \citet{perkovic2018complete} for the validity of the $\Adj$-set (for the case without $\Sel$), we can state that the $\Opt$-set is valid if and only if a valid adjustment set exists. 

\begin{mythm}[Validity of O-set] \label{thm:validity}
Given Assumptions~\ref{assum:opt} but \emph{without} a priori assuming that a valid adjustment set exists (apart from the requirement $\Sel\cap\forb=\emptyset$). If and only if a valid backdoor adjustment set exists, then $\Opt$ is a valid adjustment set.
\end{mythm}

\subsection{Graphical optimality}
\begin{figure*}[t]  
\centering
\includegraphics[width=1.\linewidth]{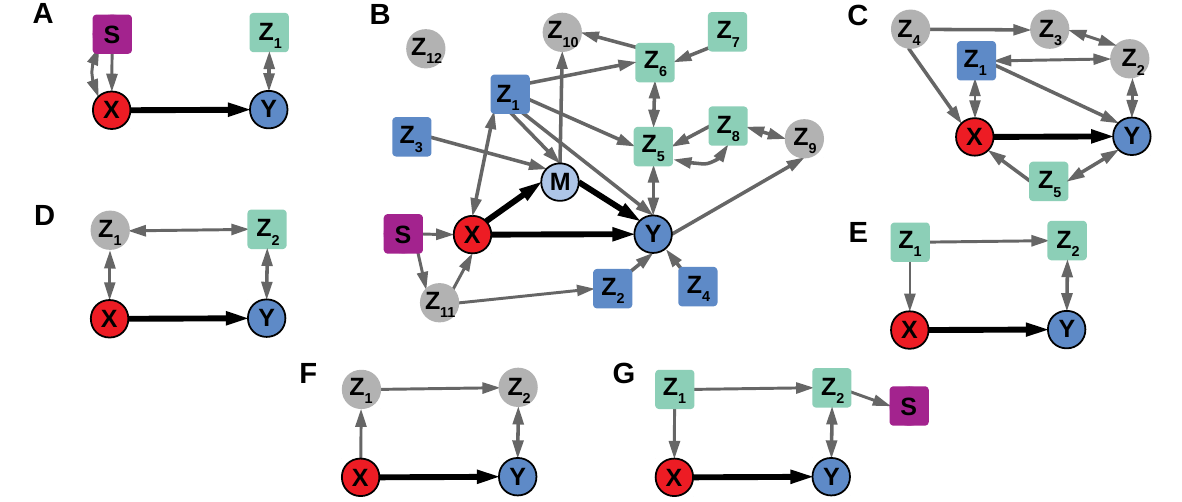}%
\caption{Examples illustrating (optimal) adjustment sets. In all examples the causal effect along causal paths (thick black edges) between $\X$ (red circle) and $\Y$ (blue circle) potentially through mediators $\M$ (light blue circle), and conditioned on some variables $\Sel$ (purple box), is considered. The adjustment set $\Opt$ consists of $\Par$ (blue boxes) and $\Col\Par_\Col\setminus \Par$ (green boxes).
See main text for details.}
\label{fig:examples}
\end{figure*}

We now move to the question of optimality. It is known that there are graphs where no graphical criterion exists to determine optimality. Examples, discussed later, are the graphs in Figs.~\ref{fig:examples}E,F. 

Before stating necessary and sufficient conditions for graphical optimality, I mention that next to the $\Opt$-set defined above and the Adjust set $\ancs$ \citep{perkovic2018complete}, I am not aware of any other systematically constructed set that will yield a valid adjustment set for the case with hidden variables. \citet{van2019separators} provide algorithms to list all valid adjustment sets, but the question is which of these a user should choose. As mentioned above, Lemma~\ref{thm:existence} can be used to cross-compare all pairs of sets, but this is not really feasible.
Hence, for automated causal effect estimation, rather than the question of whether graphical optimality holds, it is crucial to have a set with better properties than other systematically constructable sets. The following theorem states that the adjustment informations follow $J_\Opt\geq J_\ancs$ for \emph{any} graph (whether graphical optimality holds or not).

\begin{mythm}[O-set vs. Adjust-set
] \label{thm:opt_vs_adjust}
Given Assumptions~\ref{assum:opt} with $\Opt$ defined in Def.~\ref{def:def_opt} and  the Adjust-set defined in Eq.~\eqref{eq:adjust}, it holds that $J_\Opt\geq J_{\Adj}$ for any graph $\Gra$. We have $J_\Opt= J_{\Adj}$ only if (1)~$\Opt=\ancs$, or (2)~$\Opt\subseteq \ancs$ and $\X\ci \ancs\setminus \Opt~|~\Opt\Sel$.
\end{mythm}

In the following the $\Opt$-set is illustrated and conditions for graphical optimality are explored. SSR20 provide a sufficient condition for optimality, which states that either all nodes are observed (no bidirected edges exist) or for all observed nodes $\V\subset \ancs$. This is a very strict assumption and not fulfilled for any of the examples (except for Example G) discussed in the following.

\textbf{Example B.}
Figure~\ref{fig:examples}B  depicts a larger example to illustrate the $\Opt$-set $\Opt=\Par\Col\Par_\Col$ with $\Par=Z_1Z_2Z_3Z_4$ (blue boxes) and $\Col\Par_\Col\setminus \Par=Z_5Z_6Z_7Z_8$ (green boxes). We also have a conditioned variable $\Sel$. Among $\Par$, only $Z_1Z_2$ are needed to block non-causal paths to $\X$, $Z_3Z_4$ are only there to constrain information in $\Y$. Here the same holds for the whole set $\Col\Par_\Col\setminus \Par$ which was constructed from the paths $Y\headhead Z_5\headhead Z_6\headtail Z_7$ and $Y\headhead Z_5\headhead Z_8$ which does not include $Z_9$ since it is a descendant of $\Y\M$. Including an independent variable like $Z_{12}$ in $\Opt$ would not decrease the adjustment information $J_\Opt$, but then $\Opt$ would not be of minimum cardinality anymore (proven in Cor.~\ref{thm:minimality}). Here, again, the condition of SSR20 does not hold (e.g., $Z_5$ is not an ancestor of $XY\Sel$).
$\Opt$ is optimal here which can be seen as follows: For term~(iii) in the inequalities~\eqref{eq:decomp_zz} to even be non-zero, we would need a valid $\Z$ such that $\Z\setminus\Opt$ has a path to $Y$ given $\Opt\Sel X$. But these are all blocked. Note that while $Z_{10}$ or $Z_{9}\in\Z$ would open a path to $Y$, both of these are descendants of $\M$ or $\Y$ and, hence, cannot be in a valid $\Z$. For term~(iv) to even be non-zero $\Opt\setminus\Z$ would need to have a path to $X$ given $\Z\Sel$. But since any valid $\Z$ has to contain $Z_1$ and $Z_2$ (or $Z_{11})$, the only nodes in $\Opt$ with a path to $X$ are parents of $\Y\M$ and paths from these parents to $X$ all need to be blocked for a valid $\Z$. Hence, $\Opt$ is optimal here.

\textbf{Example C.}
In Fig.~\ref{fig:examples}C a case is shown where $\Opt=Z_1Z_5$. $Z_2$ is not part of $\Opt$ because none of the conditions in Def.~\ref{def:def_collpath}(2) is fulfilled: $Z_2\notin\ancs=Z_1Z_4Z_5$ and $Z_2 \cancel{\ci} \X ~|~\ancs$. Hence, we call $Z_2$ an N-node. But $Z_2$ cannot be part of any valid $\Z$ because it has a collider path to $X$ through $Z_1$ which is always open because it is part of $\ancs$. Hence, term~(iii) is always zero. Term~(iv) is zero because $\Opt\setminus\Z$ is empty for any valid $\Z$ here. Here even $J_\Opt> J_\Z$  since $\Opt$ is minimal and term~(ii) $I_{X;\Z\setminus\Opt|\Opt}>0$ for any $\Z\neq\Opt$ (generally proven in Corollary~\ref{thm:minimality}).

\textbf{Example D.}
The example in Fig.~\ref{fig:examples}D  depicts a case with $\Opt=Z_2$ where $Z_1$ is an N-node. Next to $\Z=\emptyset$ another valid set is $\Z=Z_1$. Then term~(iii) is non-zero and in the same way term~(iv) is non-zero. The sufficient  pairwise comparison criterion in SSR20 and HPM19 is, hence, not applicable.
However, it holds that always term~(iii)~$\leq$~(i) because the dependence between $Z_1$ and $Y$ given $X$ is always smaller than the dependence between $Z_2$ and $Y$ given $X$ and correspondingly term~(iv)~$\leq$~(ii). Hence, $\Opt$ is optimal here.
If a link $Y\tailhead Z_1$ exists, then the only other valid set is $\Z=\emptyset$ and both terms are strictly zero. 

\textbf{Example E.}
The example in Fig.~\ref{fig:examples}E (Fig.~3 in SSR20 and also discussed in HPM19) is not graphically optimal. Here $\Opt=Z_1Z_2$. Other valid adjustment sets are $Z_1$ or the empty set. From using $Z_1\ci Y|X$ and $X\ci Z_2|Z_1$ in the inequalities~\eqref{eq:decomp_zz} one can derive in information-theoretic terms that both $Z_1Z_2$ and $\emptyset$ are better than $\ancs=Z_1$, but since $J_{Z_1Z_2}=J_{\emptyset}+I_{Z_2;Y|XZ_1}-I_{X;Z_1}$, a superior adjustment set depends on how strong the link $Z_1\tailhead X$ vs. $Z_2\headhead Y$ is. The graph stays non-optimal also with a link $Z_1\headhead Z_2$. 

\textbf{Example F.}
The example in Fig.~\ref{fig:examples}F is also not graphically optimal. Here $\Opt=\emptyset$ and $Z_2$ is an N-node with a non-collider path to $X$. Other valid adjustment sets are $Z_1$ and $Z_1Z_2$. Higher adjustment information here depends on the distribution. Also the same graph with the link $Z_1\headhead X$ is non-optimal. If, however, there is another link $Z_1\tailhead Y$, then $\Opt=\emptyset$ is optimal (then $Z_1$ is a mediator).

\textbf{Example G.}
The example in Fig.~\ref{fig:examples}G is only a slight modification of Example E with an added selected condition $\Sel$. Then $Z_1,Z_2\in\ancs$. We still get $\Opt=Z_1Z_2$ and this is now optimal since $Z_2$ is always open and any valid set has to contain $Z_1$.

The main result of this work is a set of necessary and sufficient conditions for the existence of graphical optimality and the proof of optimality of the $\Opt$-set which is based on the intuition gained in the preceding examples. 

\begin{mythm}[Necessary and sufficient graphical conditions for optimality and optimality of O-set] \label{thm:graph_optimality}
Given Assumptions~\ref{assum:opt} and with $\Opt=\Par\Col\Par_\Col$ defined in Def.~\ref{def:def_opt}. Denote the set of N-nodes by $\mathbf{N}=\spouse(\Y\M\Col)\setminus (\forb\Opt\Sel)$. Finally, given an $N\in \mathbf{N}$ and a collider path $N\headhead \cdots \headhead C\headhead \cdots\headhead W$ (including $N\headhead W$) for $C\in \Col$ and $W\in \Y\M$ (indexed by $i$) with the collider path nodes denoted by $\pi^N_i$ (excluding $N$ and $W$), denote by $\Optnpi=\Opt(\X,\Y,\Sel'=\Sel N \pi^N_i)$ the O-set for the causal effect of $\X$ on $\Y$ given $\Sel'=\Sel\cup \{N\} \cup \pi^N_i$. 
If and only if exactly one valid adjustment set exists, or both of the following conditions are fulfilled, then graphical optimality holds and $\Opt$ is optimal:

(I)~For \emph{all} $N\in \mathbf{N}$ and all its collider paths $i$ to $W\in Y\M$ that are inside $\Col$ it holds that $\Optnpi$ does not block all non-causal paths from $\X$ to $\Y$, i.e., $\Optnpi$ is non-valid, 

and

(II)~for all $E\in \Opt \setminus \Par$ with an open path to $\X$ given $\Sel \Opt \setminus \{E\}$ there is a link $E \headhead W$ or an extended collider path $E\asthead C \headhead \cdots \headhead W$ inside $\Col$ for $W\in \Y\M$ where all colliders $C\in \ancs$.
\end{mythm}

Condition~(I) and (II) essentially rule out the two canonical cases in Examples F and E, respectively, on which non-optimality in any graph is based.
Applied to the examples, we obtain that in Example A Cond.~(I) holds since no N-node exists and Cond.~(II) holds since $X~\ci~Z_1~|~S$. In Example B also no N-node exists and Cond.~(II) holds as $X~\ci~E~|~\Sel \Opt\setminus \{E\}$ for every $E\in \Opt\setminus \Par$. In example C $Z_2$ is an N-node, but there is a collider path to $X$ through $Z_1$ which is in $\ancs$ such that Cond.~I is fulfilled. Further, while $X~\cancel{\ci}~Z_5~|~\Sel \Opt\setminus \{Z_5\}$, there is a link $Z_5\headhead Y$ such that Cond.~II holds. In example D $Z_1$ is an N-node, but it has a bidirected link with $X$ and Cond.~(II) holds since $X~\ci~Z_2~|~\Sel \Opt\setminus \{Z_2\}$. In Example E optimality does not hold, but Cond.~(I) actually holds since there is no N-node. Cond.~(II) is not fulfilled for $E=Z_1$, which has a path to $X$ given $\Opt$ and on the extended collider path $Z_1\tailhead Z_2\headhead Y$ $Z_2\notin \ancs$. For $\Z'=\emptyset$ and a distribution $\Prob'$ where the link $Z_2\headhead Y$ almost vanishes we then have $J_{\Opt}<J_{\Z'}$. Example F has an N-node $Z_2$ and $\Optnpi=\Opt(\X,\Y,\Sel'=Z_2)=Z_1Z_2$ is valid implying that Cond.~(I) does not hold, while Cond.~(II) is actually fulfilled with $\Opt=\emptyset$. For $\Z'=\Optnpi=Z_1Z_2$ and a distribution $\Prob'$ where the link $X\tailhead Z_1$ almost vanishes we then have $J_{\Opt}<J_{\Z'}$. Example G is optimal since there are no N-nodes and $Z_2\in \ancs$.

Similar to SSR20, HPM19, and \citet{witte2020efficient}, I also provide results regarding minimality and minimum cardinality for the hidden variables case in the Supplement.




\section{Numerical experiments} \label{sec:numerics}

We now investigate graphical optimality empirically to answer three questions: Firstly, whether for a linear estimator under Assumptions~\ref{assum:estimator} the asymptotically optimal variance also translates into better \emph{finite-sample} variance. Secondly, how the $\Opt$-set performs in non-optimal settings (according to Thm.~\ref{thm:graph_optimality}). Thirdly, how the $\Opt$-set and variants thereof perform for estimators not captured by the class for which the theoretical results were derived (Assumptions~\ref{assum:estimator}). 
To this end, we compare the performance of $\Opt$, Adjust, $\Optcmin$, $\Optmin$, Adjust$_{\rm Xmin}$, and Adjust$_{\rm min}$ (see definitions in Section~\ref{sec:Oset_def}) together with linear least squares estimation (LinReg) on linear models. In the Supplement we also investigate nonlinear models using nearest neighbor regression (kNN), a multilayer perceptron (MLP), random forest regression, and double machine learning for partially linear regression models (DML) \citep{chernozhukov2018double}. The experiments are based on a generalized additive model and described in detail in Section~\ref{sec:numerics_setup}. 
Among these $12{,}000$ randomly created configurations 93\% fulfill the optimality conditions in Thm.~\ref{thm:graph_optimality}.

\begin{figure*}[t]  
\centering
\includegraphics[width=1.\linewidth]{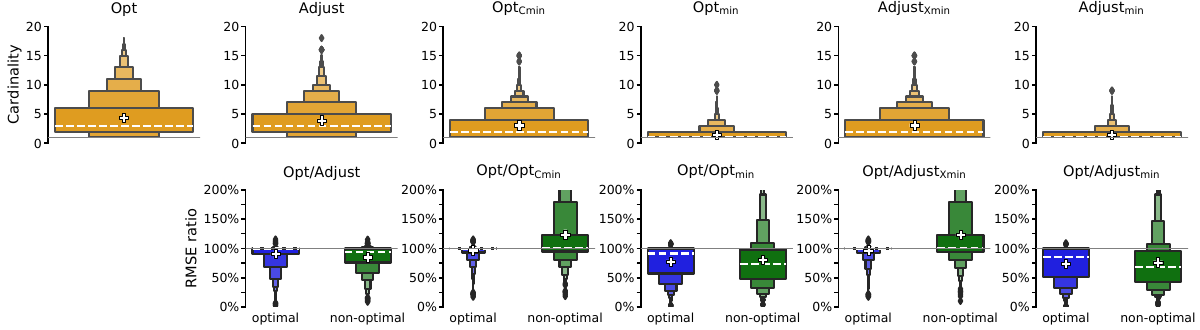}%
\caption{
Results of linear experiments with LinReg and sample size $n=100$. 
Shown are letter-value plots \citep{Hofmann2017}  of adjustment set cardinalities (top row), as well as RMSE ratios (bottom row) for the $\Opt$-set vs. other approaches for optimal configurations (left in blue) and non-optimal configurations (right in green). RMSE was estimated from $100$ realizations. The dashed horizontal line denotes the median of the RMSE ratios, and the white `plus' their average. The letter-value plots are interpreted as follows: The widest box shows the 25\%--75\% range. The next smaller box above (below) shows the 75\%--87.5\% (12.5\%--25\%) range and so forth.}
\label{fig:experiments_linear_100_short}
\end{figure*}

The results in Fig.~\ref{fig:experiments_linear_100_short} confirm our first hypothesis that for linear experiments with an estimator fulfilling Assumptions~\ref{assum:estimator} and in settings where graphical optimality is fulfilled (Thm.~\ref{thm:graph_optimality}) the $\Opt$-set either has similar RMSE or significantly outperforms all other tested variants. In particular, $\Optmin$ and Adjust$_{\rm min}$ are bad choices for this setting. Adjust is intermediate and $\Optcmin$ and Adjust$_{\rm Xmin}$ come closest to $\Opt$, but may still yield significantly higher variance. 

Secondly, in non-optimal settings (only 7\% of configurations) the $\Opt$-set still outperforms Adjust (as expected by Thm.~\ref{thm:opt_vs_adjust}). Compared to $\Optcmin$ and Adjust$_{\rm Xmin}$ the $\Opt$-set leads to worse results for about half of the studied configurations, while $\Optmin$ and Adjust$_{\rm min}$ are still bad choices. Cardinality is slightly higher for $\Opt$ compared to all other sets. In Fig.~\ref{fig:linear_cardinality_lineargaussian} we further differentiate the results by the cardinality of the $\Opt$-set and find that for small cardinalities (up to 4) the $\Opt$-set has the lowest variance in a majority of cases, but for higher cardinalities either $\Optcmin$ or again $\Opt$ have the lowest variance (slightly beating Adjust$_{\rm Xmin}$). Hence, either $\Opt$ or $\Optcmin$ performs best in non-optimal configurations.
For very small sample sizes $n=30$ (see Fig.~\ref{fig:experiments_linear_30}) that become comparable to the adjustment set cardinality, there tends to be a trade-off and smaller cardinality helps. Then $\Optcmin$  tends to be better than $\Opt$ for high cardinalities also in optimal settings, but here this effect is only present for $n=30$ and for $n=50$ already negligible compared to the gain in $J_\Opt$. 
In Appendix~\ref{sec:linear_appendix} are RMSE ratios for all combinations of adjustment approaches considered here and it is shown that, in general, results are very similar for other sample sizes.

Thirdly, we investigate non-parametric estimators on linear as well as nonlinear models (implementations described in Section~\ref{sec:numerics_setup}, results in the figures of Section~\ref{sec:nonparametric_appendix}) 
The different classes of estimators exhibit quite different behavior. For \textbf{kNN} (Figs.~\ref{fig:experiments_knn_1000},\ref{fig:experiments_knn_1000_nonlinear}) the $\Opt$-set has the lowest variance in around 50\% of the configurations followed by $\Optcmin$ and $\Optmin$. More specifically (Figs.~\ref{fig:cardinality_lineargaussian},\ref{fig:cardinality_nonlinearmixed}), for small $\Opt$-set cardinalities up to $2$ the $\Opt$-set and for higher either $\Optmin$ or $\Optcmin$ (the latter only in non-optimal configurations) perform best. For nonlinear experiments the results are less clear for $\Opt$-set cardinalities greater than $2$, but $\Optmin$ is still a good choice.
Regarding RMSE ratios, we see that, for the cases where $\Opt$ is not the best, the $\Opt$-set can have considerably higher variance, while $\Optmin$ seems to be most robust and may be a better choice if $\Opt$ is too large.
\textbf{MLP} (Figs.~\ref{fig:experiments_mlp_1000},\ref{fig:experiments_mlp_1000_nonlinear}) behaves much differently. Here in optimal cases neither method outperforms any other for small $\Opt$-set cardinalities, but for higher cardinalities (Figs.~\ref{fig:cardinality_lineargaussian},\ref{fig:cardinality_nonlinearmixed}) the $\Opt$-set is best in more than 50\% of configurations (slightly less for nonlinear experiments) and the others share the rest (except Adjust$_{\rm min}$). For non-optimal cases $\Opt$, $\Optcmin$ and Adjust$_{\rm Xmin}$ share the ranks. Regarding RMSE, for linear experiments the $\Opt$-results are almost as optimal as for the LinReg estimator in the optimal setting. 
However, for non-optimal cases $\Optcmin$ can have considerably smaller variance and seems to be a robust option then, similarly to Adjust$_{\rm Xmin}$. Also for nonlinear experiments $\Optcmin$ is more robust.
The \textbf{RF} estimator (Figs.~\ref{fig:experiments_rf_1000},\ref{fig:experiments_rf_1000_nonlinear}) is again different. Here no method clearly is top-ranked, $\Optmin$ and Adjust$_{\rm min}$ are slightly better for linear experiments and $\Opt$ for nonlinear experiments.
$\Optcmin$ and $\Optmin$ are more robust regarding RMSE ratios (similar to Adjust$_{\rm Xmin}$).
Finally, the \textbf{DML} estimator (Fig.~\ref{fig:experiments_doubleml_1000}) was here applied only to linear experiments since its model assumption does not allow for fully nonlinear settings. For optimal settings here $\Opt$ is top-ranked in a majority of cases, but closely followed by $\Optcmin$ and Adjust$_{\rm Xmin}$. In non-optimal cases for higher $\Opt$-set cardinalities  these two seem like a better choice. Quantitatively,  $\Optcmin$ and Adjust$_{\rm Xmin}$ are the most robust choices.

Overall, the $\Opt$-set and its variants seem to outperform or match the Adjust-variants and whether higher cardinality of the $\Opt$-set reduces performance depends strongly on the estimator and data.

\section{Discussion and Conclusions}
The proposed adjustment information formalizes the common intuition to choose adjustment sets that maximally constrain the effect variable and minimally constrain the cause variable. 
The main \textbf{theoretical contributions} are a necessary and sufficient graphical criterion for the existence of an optimal adjustment set in the hidden variables case and a definition and algorithm to construct it. To emphasize, graphical optimality implies that the $\Opt$-set is optimal for \emph{any distribution} consistent with the graph. 
Note that in cases where graphical optimality does not hold, there will still be distributions for which the $\Opt$-set has maximal adjustment information.

Further, the optimal set is valid if and only if a valid adjustment set exists and has smaller (or equal) asymptotic variance compared to the Adjust-set proposed in \citet{perkovic2018complete} for any graph, whether graphical optimality holds or not. This makes the $\Opt$-set a natural choice in automated causal inference analyses. 
\textbf{Practical contributions} comprise Python code to construct adjustment sets and check optimality, as well as extensive numerical experiments that demonstrate that the theoretical results also hold for relatively small sample sizes. 

The theoretical \textbf{optimality results are limited} to estimators for which the asymptotic variance becomes minimal for adjustment sets with maximal adjustment information (relation~\eqref{eq:maxmin}). This is fulfilled for least-squares estimators, where even the direct relation~\eqref{eq:entropy_est} holds, but it is unclear whether this also holds for more general classes. The numerical results show that the $\Opt$-set or minimized variants thereof often yield smaller variance also in non-optimal settings and beyond that estimator class. 
I speculate that further theoretical properties of maximizing adjustment information can be shown because relation~\eqref{eq:entropy_est} for $f(\cdot)=\frac{1}{\sqrt{n}} e^{H_{\Y|\X \Z \Sel} - H_{\X|\Z \Sel}}$ seems related to the lower bound of the estimation variance counterpart to Fano's inequality (Theorem 8.6.6 in \citet{Cover2006}). 
For estimators sensitive to high-dimensionality one may consider data-driven criteria or penalties to step-wisely minimize the $\Opt$-set. However, estimating, for example, the adjustment information from a potentially small sample size carries considerable errors itself.
Another current limitation is that relation~\eqref{eq:entropy_est} only holds for univariate singleton cause variables $X$. The information-theoretical results, however, also hold for multivariate $\mathbf{X}$ and preliminary results indicate that, while relation~\eqref{eq:entropy_est} does not hold for multivariate $\mathbf{X}$, the less restrictive relation~\eqref{eq:maxmin} still seems to hold.

The proposed information-theoretic approach can guide \textbf{further research}, for example, to theoretically study relations~\eqref{eq:maxmin},\eqref{eq:entropy_est} for other estimators and to address other types of graphs as emerge from the output of causal discovery algorithms and the setting where the graph is unknown \citep{witte2020efficient,maathuis2009estimating,Maathuis2010}. 
At present, the approach only applies to ADMGs and \emph{Maximal Ancestral Graphs} (MAG) \citep{richardson2002} without selection variables. 
Last, it remains an open problem to identify optimal adjustment estimands for the hidden variables case based on other criteria such as the front-door formula and Pearl's general do-calculus \citep{Pearl2000}.

The results may carry considerable \textbf{practical impact} since, surprisingly, among the randomly created configurations more than 90\% fulfill the optimality conditions indicating that also in many real-world scenarios graphical optimality may hold. Code is available in the python package \url{https://github.com/jakobrunge/tigramite}.

\paragraph{Acknowledgments} I thank Andreas Gerhardus for very helpful comments. This work was funded by the ERC Starting Grant CausalEarth (grant no. 948112).

\clearpage

\clearpage
\appendix
\counterwithin{mythm}{section}
\counterwithin{mylemma}{section}
\counterwithin{myprop}{section}
\counterwithin{mycor}{section}
\counterwithin{mydef}{section}
\counterwithin{algorithm}{section}
\renewcommand\thefigure{S\arabic{figure}}    
\setcounter{figure}{0}  
\renewcommand\theequation{S\arabic{equation}}    
\setcounter{equation}{0}  

\section{Problem setting and preliminaries} \label{sec:setting}
\subsection{Graph terminology}

We consider causal effects in causal graphical models over a set of variables $\V$  with a joint distribution $\Prob=\Prob(\V)$ that is consistent with an acyclic  directed  mixed  graph (ADMG) $\Gra=(\V,\mathcal{E})$. Two nodes can have possibly more than one edge which can be \emph{directed} ($\headtail$) or \emph{bi-directed} ($\headhead$). We use ``$\ast$'' to denote either edge mark. There can be no loops or directed cycles. See Fig.~\ref{fig:motivation_venn}A for an example. The results also hold for \emph{Maximal Ancestral Graphs} (MAG) \citep{richardson2002} without selection variables. 
A path between two nodes $X$ and $Y$ is a sequence of edges such that every edge occurs only once. A path between $X$ and $Y$ is called \emph{directed or causal} from $X$ to $Y$ if all edges are directed towards $Y$, else it is called \emph{non-causal}. A node $C$ on a path is called a \emph{collider} if ``$\asthead C \headast$''. Kinships are defined as usual: parents $\pa(X, \Gra)$ for ``$\bullet\tailhead X$'', spouses $\spouse(X, \Gra)$  for ``$X\headhead \bullet$'', children $\ch(X, \Gra)$ for ``$X\tailhead\bullet$'', and correspondingly descendants $\des$ and ancestors $\an$. We omit the $\Gra$ in the following since all relations are relative to the graph $\Gra$ in this paper. Our approach does not involve modified graph constructions as in \citet{van2019separators} and other works. A node is an ancestor and descendant of itself, but not a parent/child/spouse of itself. The mediator nodes on causal paths from $X$ to $Y$ are denoted $\M=\M(X, Y)$ and exclude $X$ and $Y$ (different from definitions in other works). 
For sets of variables the kinship relations correspond to the union of the individual variables. For parent/child/spouse-relationships these exclude the set of variables itself.
A path $\pi$ between $X$ and $Y$ in $\Gra$ is blocked (or closed) by a node set $\Z$ if (i)~$\pi$ contains a non-collider in $\Z$ or (ii)~$\pi$ contains a collider that is not in $\an(\Z)$. Otherwise the path $\pi$ is open (or active/connected) given $\Z$. 
Nodes $\X$ and $\Y$ are said to be m-separated given $\Z$  if every path between them is blocked by $\Z$, denoted as $\X \ci \Y | \Z$. In the following we will simplify set notation and denote unions of variables as $\{W\}\cup \M \cup \A=W\M\A$.

\section{Further theoretical results and proofs}

\subsection{Properties of adjustment information}
$J_\Z$ is not necessarily positive if the dependence between $\X$ and $\Z$ (given $\Sel$) is larger than that between $\Z$ and $\Y$ given $\X\Sel$. By the properties of CMI, it is bounded by 
\begin{align} 
- \min (H_{\X|\Sel}, H_{\Z|\Sel})\leq J_{\X\Y|\Sel.\Z} \leq \min (H_{\Y|\X\Sel}, H_{\Z|\X\Sel})\,.
\end{align}

\subsection{Causally sufficient case} \label{sec:suff}
\begin{figure}[h]  
\centering
\includegraphics[width=.25\linewidth]{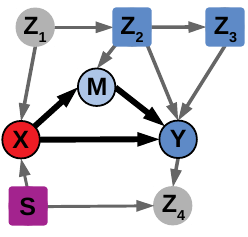}%
\caption{DAG version of graph in Fig.~\ref{fig:motivation_venn}A with $\Opt$-set shown as blue boxes.}
\label{fig:examples_suff}
\end{figure}

The optimal adjustment set for the causally sufficient case was derived in HPM19 and \citet{rotnitzky2019efficient}. Here the derivation is discussed from an information-theoretic perspective.

\begin{mydef}[O-set in the causally sufficient case] \label{def:def_opt_suff}
Given Assumptions~\ref{assum:opt} restricted to DAGs with no hidden variables, define the set
\begin{align*}
   \Opt &= \Par = \pa(\Y\M) \setminus \forb\,.
\end{align*}
\end{mydef}

In the causally sufficient case a valid adjustment set always exists and the $\Opt$-set is always valid since $\Opt$ contains no descendants of $\Y\M$ and all non-causal paths from $\X$ to $\Y$ are blocked since $\Par$ blocks all paths from $\X$ through parents of $\Y\M$. 

Figure~\ref{fig:examples_suff} shows an example DAG with a mediator $M$ and conditioned variable $S$. The $\Opt$-set $\Opt=Z_2Z_3$ is depicted by blue boxes. Compare $\Opt$ with $\ancs=Z_1Z_2Z_3\Sel$ (Adjust-set in \citet{perkovic2018complete}) in the inequalities~\eqref{eq:decomp_zz}. Since $Z_1~\ci~Y~|~\Opt XS$, term~(iii) is zero and since $\Opt\setminus\ancs=\emptyset$, also term~(iv) is zero. Further, terms~(i) and (ii) are both strictly greater than zero (under Faithfulness). Then $J_{\Opt}> J_{\ancs}$ and under Assumptions~\ref{assum:estimator} by Lemma~\ref{thm:variance} the $\Opt$-set has a smaller asymptotic variance than $\ancs$. Since the parents of $\Y\M$ block all paths from any other valid adjustment sets to $Y$ and because any valid adjustment set $\Z$ has to block paths from $X$ to $\pa(Y\M)\setminus\Z$, $J_{\Opt}\geq J_{\Z}$ holds in general for any valid set $\Z$ as proven from an information-theoretic perspective in Proposition~\ref{thm:optimality_suff}.

\begin{myprop}[Optimality of O-set in causally sufficient case] \label{thm:optimality_suff}
Given Assumptions~\ref{assum:opt} restricted to DAGs with no hidden variables  and with $\Opt=\Par$ defined in Def.~\ref{def:def_opt_suff}, graphical optimality holds for any graph and $\Opt$ is optimal.
\end{myprop}

Similar to HPM19 and \citet{witte2020efficient}, there also exist results regarding minimality and minimum cardinality which are covered for the hidden variables case in Corollary~\ref{thm:minimality}.

\subsection{Hidden variables case}
Here we provide some further theoretical results for the general hidden variables case in addition to the lemmas and theorems in the main text.

\begin{mycor}[Minimality and minimum cardinality] \label{thm:minimality}
Given Assumptions~\ref{assum:opt}, assume that graphical optimality holds, and, hence, $\Opt$ is optimal. Further it holds that:
\begin{enumerate}
\item  If $\Opt$ is not minimal, then  $J_{\Opt}> J_{\Z}$  for all \emph{minimal} valid $\Z\neq\Opt$,

\item If $\Opt$ is minimal valid, then $\Opt$ is the unique set that maximizes the adjustment information $J_\Z$ among all \emph{minimal} valid $\Z\neq\Opt$,

\item  $\Opt$ is of minimum cardinality, that is, there is no subset of $\Opt$ that is still valid and optimal.
\end{enumerate}
\end{mycor}

Another relevant Proposition states that $\Optcmin$ is a subset of $\ancs$, similar to corresponding Lemmas in \citet{van2019separators}.
\begin{myprop}[Collider-minimized O-set is a subset of Adjust.] \label{thm:minimal_ancs}
Given Assumptions~\ref{assum:opt} with $\Opt=\Par\Col\Par_{\Col}$ defined in Def.~\ref{def:def_opt} and the $\Optcmin$-set constructed with Alg.~\ref{algo:optmin} it holds that $\Optcmin\subseteq\ancs$.
\end{myprop}

\subsection{Proof of Lemma~\ref{thm:existence}}
\begin{mylemma*}[Necessary and sufficient comparison criterion for existence of an optimal set] 
Given Assumptions~\ref{assum:opt}, if and only if there is a $\Z\in\mathcal{Z}$ such that either there is no other $\Z'\neq \Z\in\mathcal{Z}$ or for all other $\Z'\neq\Z\in\mathcal{Z}$ and all distributions $\Prob$ consistent with $\Gra$ it holds that
\begin{align} \label{eq:decomp_zz_appendix}
\underbrace{I_{\Z\setminus\Z';\Y|\Z' \X\Sel}}_{\text{(i)}} &\geq \underbrace{I_{\Z'\setminus\Z;\Y|\Z \X \Sel}}_{\text{(iii)}},~~~\text{and} \nonumber \\
\underbrace{I_{\X;\Z'\setminus\Z|\Z \Sel}}_{\text{(ii)}}  &\geq \underbrace{I_{\X;\Z\setminus\Z'|\Z' \Sel}}_{\text{(iv)}}\,,
\end{align}
then graphical optimality holds and $\Z$ is optimal implying $J_\Z \geq J_{\Z'}$.
\end{mylemma*}

\textit{Proof.}
If there is no other $\Z'$, the statement trivially holds. Assuming there is another $\Z'$, we prove the two implications as follows by an  information-theoretic decomposition.

Define disjunct (possibly empty) sets $\R,\B,\A$ with $\Z=\A\B$ and $\Z'=\B\R$ with $\B=\Z\cap\Z'$. Note that if both $\R=\emptyset$ and $\A=\emptyset$, then $\Z=\Z'$.
Consider two different ways of applying the chain rule of CMI,
\begin{align}
& I_{\A\B\R;\Y|\X\Sel} -  I_{\X;\A\B\R|\Sel} \nonumber\\
&=  I_{\A\B;\Y|\X\Sel} +  I_{\R;\Y|\A\B \X \Sel} -I_{\X;\A\B|\Sel} - I_{\X;\R|\A\B\Sel}\\
&= I_{\B\R;\Y|\X\Sel} + I_{\A;\Y|\B\R \X\Sel} - I_{\X;\B\R|\Sel} - I_{\X;\A|\B\R \Sel}\,,
\end{align}

from which with $J_\Z=I_{\A\B;\Y|\X\Sel} - I_{\X;\A\B|\Sel}$ and $J_{\Z'}=I_{\R\B;\Y|\X\Sel} - I_{\X;\R\B|\Sel}$ it follows that
\begin{align} \label{eq:decomp}
 J_\Z &=J_{\Z'}  \nonumber \\
   &\phantom{=}+  \underbrace{I_{\A;\Y|\B\R \X\Sel}}_{\text{(i)}} +  \underbrace{I_{\X;\R|\A\B \Sel}}_{\text{(ii)}}  - \underbrace{I_{\R;\Y|\A\B \X \Sel}}_{\text{(iii)}} - \underbrace{I_{\X;\A|\B\R \Sel}}_{\text{(iv)}}\,.
\end{align}

The inequalities~\eqref{eq:decomp_zz_appendix} then read
\begin{align}
\underbrace{I_{\A;\Y|\B\R \X\Sel}}_{\text{(i)}} &\geq \underbrace{I_{\R;\Y|\A\B \X \Sel}}_{\text{(iii)}} ,~~~\text{and} \nonumber \\
\underbrace{I_{\X;\R|\A\B \Sel}}_{\text{(ii)}}  &\geq \underbrace{I_{\X;\A|\B\R \Sel}}_{\text{(iv)}}\,.
\end{align}

``if'': If term~(i) is greater or equal to term~(iii) and term~(ii) greater or equal to term~(iv), then trivially $J_\Z \geq J_{\Z'}$ for all distributions $\mathcal{P}$.

``only if'': We prove the contraposition that if for all valid $\Z$ there exists a valid $\Z'\neq\Z$ and a distribution $\Prob$ consistent with $\Gra$ such that
\begin{align}
\underbrace{I_{\A;\Y|\B\R \X\Sel}}_{\text{(i)}} &< \underbrace{I_{\R;\Y|\A\B \X \Sel}}_{\text{(iii)}} ,~~~\text{or}~~~ \underbrace{I_{\X;\R|\A\B \Sel}}_{\text{(ii)}}  < \underbrace{I_{\X;\A|\B\R \Sel}}_{\text{(iv)}}\,,
\end{align}
then there always exists a modification $\Prob'$ of the distribution $\Prob$ such that  $J_\Z < J_{\Z'}$. This is because, in both cases, we can always construct a distribution for which terms (ii) and (i), respectively, become arbitrary close to zero. Consider the two cases as follows:

1) there exists a distribution $\Prob$ with $I_{\A;\Y|\B\R \X\Sel} < I_{\R;\Y|\A\B \X \Sel}$: Since CMIs are always non-negative, it holds that $\R\neq \emptyset$ and there must exist at least one open path between $\R$ and $\Y$ where every collider is in $\A\B \X \Sel$ and no non-collider is in $\A\B \X \Sel$. No such open path can pass through $\X$ because if $\X$ is a non-collider (as for paths continuing on causal paths from $\X$ to $\Y$), then the path is blocked, and if $\X$ is a collider, then there would be a non-causal path from $\X$ to $\Y$ given $\Z\Sel$ which would make $\Z$ invalid while $\Z$ is assumed valid. Correspondingly, no open path from $\A$ (if $\A\neq \emptyset$) to $\Y$ given $\B\R \X\Sel$, if a path exists at all, can pass through $\X$ if $\Z'$ is assumed valid. 
Now we can construct a distribution $\Prob'$ with associated structural causal model (SCM) consistent with $\Gra$ where $I_{\A;\Y|\B\R \X\Sel} < I_{\R;\Y|\A\B \X \Sel}$ holds as in $\Prob$ and still all links ``$U\astast X$'' for $X\in\X$ and $U\notin \X\M\Y$ \emph{almost vanish}. Consider the three possible links and associated assignment functions in the SCM: (1)~``$X\tailhead U$'' with $U:=f_{U}(\ldots,X,\ldots)$, (2)~``$X\headtail U$'' with $X:=f_{X}(\ldots,U,\ldots)$, and (3)~``$X\headhead U$'' with $X:=f_{X}(\ldots,L^U,\ldots)$ where $L^U$ denotes one or more latent variables. In each case, to go from $\Prob$ to $\Prob'$, we can modify $f_{\cdot}\to f'_{\cdot}$ where in $f'_{\cdot}$ the dependence on the respective argument is replaced by $X\to c X$, $U\to c U$, or $L^U\to c L^U$ for $c\in \mathbb{R}$, and where we consider the limit $c\to 0$. This modification does not affect $I_{\A;\Y|\B\R \X\Sel} < I_{\R;\Y|\A\B \X \Sel}$ because the paths contributing to the two CMIs cannot pass through $\X$. On the other hand, then term~(ii) $I_{\X;\R|\A\B \Sel}\to 0$ because all paths passing through $\X$ contain almost zero links and there cannot be a path from $\R$ to $\X$ through $\M\Y$ for a valid $\Z$. Hence, since in Eq.~\eqref{eq:decomp} term~(i) is smaller than term~(iii) by assumption, and term~(ii) is almost zero, it holds that $J_\Z<J_{\Z'}$.

2) there exists a distribution $\mathcal{P}$ with $I_{\X;\R|\A\B \Sel}  < I_{\X;\A|\B\R \Sel}$: As before, since CMIs are always non-negative, it holds that $\A\neq \emptyset$ and there must exist at least one open path between $\A$ and $\X$ where every collider is in $\B\R \Sel$ and no non-collider is in $\B\R \Sel$. No such open path can pass through $\Y\M$ because if any node in $\Y\M$ is a collider, then the path is blocked, and no path can contain any node in $\Y\M$ as a non-collider since then either the graph is cyclic or $\Z'$ contains descendants of $\Y\M$ leading to  $\Z'\cap\forb\neq \emptyset$ while $\Z'$ is assumed valid. Correspondingly, no open path from $\R$ (if $\R\neq \emptyset$) to $\X$ given $\A\B \Sel$, if a path exists at all, can pass through $\Y\M$ if $\Z$ is assumed valid.   
Then, analogous to before, we can construct a $\Prob'$  with associated SCM consistent with $\Gra$  where $I_{\X;\R|\A\B \Sel}  < I_{\X;\A|\B\R \Sel}$ holds and where all links ``$U\astast W$'' for $W\in\Y\M$ and $U\notin \X\M\Y$ \emph{almost vanish}. Then term~(i) $I_{\A;\Y|\B\R\X \Sel}\to 0$ because all paths contain almost zero links and there cannot be a path from $\A$ to $\Y$ where $\X$ contains a collider for a valid $\Z'$ since this would constitute a non-causal path. Hence, since in Eq.~\eqref{eq:decomp} term~(ii) is smaller than term~(iv) by assumption, and term~(i) is almost zero, it holds that $J_\Z<J_{\Z'}$.
\hfill $\square$

\subsection{Proof of Proposition~\ref{thm:optimality_suff}}
\begin{myprop*}[Optimality of O-set in causally sufficient case] 
Given Assumptions~\ref{assum:opt} restricted to DAGs with no hidden variables  and with $\Opt=\Par$ defined in Def.~\ref{def:def_opt_suff}, graphical optimality holds for any graph and $\Opt$ is optimal.
\end{myprop*}

\textit{Proof.}
The proof is based on Lemma~\ref{thm:existence} and relation~\eqref{eq:decomp}. We will prove that for any DAG $\Gra$ term~(i)$\geq$(iii) and term~(ii)$\geq$(iv) from which optimality follows by Lemma~\ref{thm:existence}.

We have to show that $I_{\A;\Y|\B\R \X\Sel} \geq I_{\R;\Y|\A\B \X \Sel}$ and $I_{\X;\R|\A\B \Sel} \geq I_{\X;\A|\B\R \Sel}$ where $\Opt=\A\B$ and $\Z'=\R\B$ with $\B=\Opt\cap\Z'$. 

Any path from $\X$ or $\V\setminus  \Y\M\Opt\Sel\X$ to $\Y\M$ given $\Opt\Sel$ (denoted by $\boxed{\cdot}$), excluding the causal path from $\X$ to $\Y$, features at least one of the following motifs: ``$X,V\astast\boxed{P}\tailhead W$'' (excluding ``$X\tailhead\boxed{P}\tailhead W$''), or ``$V\headtail W$'' where, hence, $V\in \forb$.

Now all paths from a valid adjustment set $\Z'$ with $\Z'\in\mathcal{Z}$ to $\Y$ are blocked given $\Opt\Sel$: Motif ``$X,V\astast\boxed{P}\tailhead W$'' contains a non-collider in $\Opt\Sel$ and is, hence, blocked. In motif ``$V\headtail W$'' $V\in \forb$. Since $\X\notin \des(\Y)$ (acyclicity) and $\Z'\cap \des(\Y)=\emptyset$ (validity of $\Z'$), the paths from $\Z'$ to $V$ either end with a head at $V$  or there must be a collider $K$ that is a descendant of $V$ and hence, $K\in \forb$. Then $K\notin  \an(\Opt\Sel)$ and $K\notin \Z'$ and the path is therefore blocked.
Hence, with $\R\subseteq\Z'$, term (iii) is zero by Markovity. 
 
Term (iv) $I_{\X;\A|\Z' \Sel}=0$ for any valid $\Z'$ because $\A\subseteq \pa(\Y\M)$ and then otherwise there would be a non-causal path from $\X$ through $\A$ to $\Y\M$.
\hfill $\square$

\subsection{Further Lemmas} \label{sec:lemmas}

\begin{mylemma}[Relevant path motifs wrt. the O-set] \label{thm:motifs}
Given Assumptions~\ref{assum:opt}  but \emph{without} a priori assuming that a valid adjustment set exists (apart from the requirement $\Sel\cap\forb=\emptyset$). With $\Opt=\Par\Col\Par_{\Col}$ defined in Def.~\ref{def:def_opt} any path from $\X$ or $\V\setminus  \Y\M\Opt\Sel\X$ to $\Y\M$ given $\Opt\Sel$ (denoted by $\boxed{\cdot}$), excluding the causal path from $\X$ to $\Y$, features at least one of the following motifs with certain constraints as indicated. We denote $V\in\V\setminus  \Y\M\Opt\Sel\X$ and further differentiate nodes in $\Y\M$ as $W\in \Y\M$ and in $\Opt=\Par\Col\Par_{\Col}$ as $C\in \Col$ or $P\in \Par$ or $P_C\in \Par_\Col$. Last, we denote those collider path nodes not included in the $\Opt$-set in Alg.~\ref{algo:opt} due to not sufficing Def.~\ref{def:def_collpath}(1) as $F$ with $F\in \forb$ and those not sufficing Def.~\ref{def:def_collpath}(2a,b) as $N$ with $N\notin \forb$, $N\notin \ancs$, and $N~\cancel{\ci}~\X~|~\ancs$:

(1a)~``$\astast X\tailhead \boxed{C} \headhead$'' 

(1b)~``$\astast X\tailhead \boxed{P_C}\tailhead \boxed{C} \headhead$'' 

(2a)~``$X,V\astast\boxed{P}\tailhead W$'' excluding ``$X\tailhead\boxed{P}\tailhead W$''

(2b)~``$X,V\astast\boxed{P_C}\tailhead \boxed{C} \headhead$'' excluding (1b)

(3a)~``$V\headtail W$'' where, hence, $V\in \forb$ 

(3b)~``$X,V\headtail \boxed{C} \headhead$''

(4a)~``$\astast F\headhead W$''  with the constraint $F\notin\ancs$

(4b)~``$\astast F\headhead\boxed{C}\headhead$'' with the constraints $F\notin \pa(C)$ and $F\notin\ancs$

(5a)~``$\astast N\headhead W$''  with the constraints $N\notin \pa(W)$ and $W\notin \pa(N)$

(5b)~``$\astast N\headhead \boxed{C}\headhead$'' with the constraint $N\notin \pa(C)$

Further it holds that $F, N, X\notin \Sel$.
\end{mylemma}

\textit{Proof.}
Any path from $\X$ or $\V\setminus \Y\M\Opt\Sel\X$ to $\Y\M$ has to contain a link ``$A\astast B$'' where $A=\X$ or $A\in \V\setminus  \Y\M\Opt\Sel\X$ and $B\in \Y\M\Opt$ where $\astast\in\{\tailhead, \headtail, \headhead\}$. If we differentiate the left node by $X$ or $V\in \V\setminus  \Y\M\Opt\Sel\X$ and the right node by $W\in \Y\M$ or $C\in \Col$ or $P\in \Par$ or $P_C\in \Par_\Col$, we can in principle have $2\cdot4\cdot3=24$ link types which are motifs if we consider the adjacent links to $A$ and $B$. These are listed in the Lemma except for ``$\astast X\tailhead W$'' which is part of the causal path from $\X$ to $\Y$, ``$X\tailhead\boxed{P}\tailhead W$'' which cannot occur since then $P\in\M$, ``$V\tailhead W$'' which cannot occur since $\Par$ would contain $V$ or $V\in \des(\Y\M)$ leading to a cyclic graph, ``$V\tailhead C$'' which cannot occur since $\Par_\Col$ would contain $V$, and ``$X\headtail W$'' which cannot occur since this implies a cyclic graph.

Regarding the constraints listed in motifs (4a,b) for $F\in\forb$ it holds that $F\notin\ancs$ because $\ancs=\an(\X\Y\Sel)\setminus \forb$ by definition. Further, in (4b) $F\notin \pa(C)$ holds because otherwise $C\in \forb$. In motif (5a) $N\notin \pa(W)$ holds because $N\notin \ancs$ and  $W\notin \pa(N)$ holds because $N\notin \forb$. In motif (5b) $N\notin \pa(C)$ holds because $C\in\ancs$ contradicts $N\notin\ancs$ and $N~\cancel{\ci}~\X~|~\ancs$  with $N\tailhead C$ contradicts $C \ci \X ~|~\ancs$.
Last, it holds that $F, N, X\notin \Sel$ because $\Sel\cap\forb=\emptyset$, $\Sel\cap\X=\emptyset$ by Assumptions~\ref{assum:opt} and $N\notin \ancs$ while $\Sel\subseteq \ancs$.
\hfill $\square$

\begin{mylemma}[Sufficient condition for non-identifiability] \label{thm:nonidentifiability}
Given Assumptions~\ref{assum:opt}  but \emph{without} a priori assuming that a valid adjustment set exists (apart from the requirement $\Sel\cap\forb=\emptyset$). With $\Opt=\Par\Col\Par_{\Col}$ defined in Def.~\ref{def:def_opt},  if on any non-causal path from $\X$ to $\Y$ given $\Opt\Sel$ any of the motifs (1a) or (4a) or (4b) for $F=\X$ occurs as listed in Lemma~\ref{thm:motifs}, then the causal effect of $\X$ on $\Y$ (potentially through $\M$) is \emph{not} identifiable by backdoor adjustment.
\end{mylemma}

\textit{Proof.}
If motif (4a)~``$X\headhead W$'' for $W\in\Y\M$ occurs, the case is trivial \citep[Thm. 4.3.1]{Pearl2000}. In motifs (1a)~``$X\tailhead \boxed{C} \headhead$'' and (4b)~``$X\headhead\boxed{C}\headhead$'' we have that since Def.~\ref{def:def_collpath}(2b) $C \ci \X ~|~\ancs$ is not fulfilled, Def.~\ref{def:def_collpath}(2a) $C\in\ancs$ must be the case. Then every $C_k$ on collider paths to $W$ also fulfills $C_k\in\ancs$ because for all of them $C_k \ci \X ~|~\ancs$ does not hold since each collider is opened. Hence, there exists a collider path $X\asthead C \headhead \cdots \headhead W$ where every collider $C\in\ancs=\an(\X\Y\Sel)\setminus\forb$. This path cannot be blocked by any adjustment set (given $\Sel$): colliders with $C\in\an(\Sel)$  are always open. For colliders with $C\in\an(\X)$ or $C\in\an(\Y)$ there is a directed path to $\X$ or $\Y$ and either this path is open leading to a non-causal path, or an adjustment set contains a non-collider on that directed path which opens the collider $C$.
\hfill $\square$

In Theorem~\ref{thm:validity} we will prove that the condition in Lemma~\ref{thm:nonidentifiability} is also necessary for non-identifiability by backdoor adjustment. To this end, consider the following Lemmas.

\begin{mylemma}[Collider parents fulfill Def.~\ref{def:def_collpath}] \label{thm:colliderparents}
Given Assumptions~\ref{assum:opt}. With $\Opt=\Par\Col\Par_{\Col}$ defined in Def.~\ref{def:def_opt}, for every $P\in\Par_\Col$ conditions (1), and (2a) or (2b) in Def.~\ref{def:def_collpath} hold.
\end{mylemma}
\textit{Proof.}
Denote a pair $P_C\tailhead C$ for $C\in\Col$ fulfilling conditions (1), and (2a) or (2b) in Def.~\ref{def:def_collpath}.
Firstly, (1)~$P_C\notin \forb$ since if $P_C\in\des(\Y\M)$ also $C\in \des(\Y\M)$ and if $P_C=\X$, then by Lemma~\ref{thm:nonidentifiability} no valid adjustment set exists, contrary to Assumptions~\ref{assum:opt}. Secondly, it cannot be that (2a) $P_C \notin \ancs$ and (2b)~$P_C \cancel{\ci} \X ~|~\ancs$ because then the path from $\X$ to $P_C$ would extend to $C$ and would not be blocked because $P_C\notin \ancs$. But then also $C \notin \ancs$ and $C$ would not fulfill the conditions in Def.~\ref{def:def_collpath}. 
\hfill $\square$

\begin{mylemma}[Blockedness of parent-child-motifs] \label{thm:parentchild_blocked}
Given Assumptions~\ref{assum:opt} with $\Opt=\Par\Col\Par_{\Col}$ defined in Def.~\ref{def:def_opt}. Any path from $\X$ or a valid adjustment set $\Z$ with $\Z\in\mathcal{Z}$ to $\Y$ containing the motifs (1b), (2a), (2b), (3a), (3b) is blocked given $\Opt\Sel$.
\end{mylemma}
\textit{Proof.}
Motifs (1b), (2a), (2b), and (3b) contain a non-collider in $\Opt\Sel$ and are, hence, all blocked. In motif (3a) $V\in \forb$. Since $\X\notin \des(\Y)$ (acyclicity) and $\Z\cap \des(\Y)=\emptyset$ (validity of $\Z$), the paths from $\Z$ to $V$ either end with a head at $V$  or there must be a collider $K$ that is a descendant of $V$ and hence, $K\in \forb$. Then $K\notin  \an(\Opt\Sel)$ and $K\notin \Z$ and the path is therefore blocked.
\hfill $\square$

\begin{mylemma}[Blockedness of F-motifs] \label{thm:desYM_blocked}
Given Assumptions~\ref{assum:opt} with $\Opt=\Par\Col\Par_{\Col}$ defined in Def.~\ref{def:def_opt}. Firstly, any path from $\X$  to $\Y$ containing the motifs (4a) or (4b) for $F\in\des(\Y\M)$ is blocked given $\Opt\Sel$. Secondly, any path from  a valid adjustment set $\Z$ with $\Z\in\mathcal{Z}$ to $\Y$ containing the motifs (4a) or (4b) for $F\in\des(\Y\M)$ is blocked given $\X\Opt\Sel$.
\end{mylemma}

\textit{Proof.}
First statement: $F\notin \ancs$ by Lemma~\ref{thm:motifs} and, hence, in particular $F\notin\an(\X)$. Then, if a path exists, either the paths from $\X$ to $F$ end with a head at $F$ or there must be at least one collider $K$  with $F\in\an(K)$ on a path to $\X$. Now $F,K\notin \an(\Opt\Sel)$ because $\Opt\Sel\cap \forb=\emptyset$ and the path is blocked.
Secondly, $F\notin \an(\Z)$ since $\Z$ is valid. Then similarly, if a path exists, either the paths from $\Z$ to $F$ end with a head at $F$  or there must be at least one collider $K$ on a path to $\Z$ with $F\in\an(K)$. Now $F,K\notin \an(\X\Opt\Sel)$ because $\Opt\Sel\cap \forb=\emptyset$ and $F\notin \ancs$ by Lemma~\ref{thm:motifs} and the path is blocked.
\hfill $\square$

\begin{mylemma}[Blockedness of N-motifs] \label{thm:N_blocked}
Given Assumptions~\ref{assum:opt} with $\Opt=\Par\Col\Par_{\Col}$ defined in Def.~\ref{def:def_opt}. Firstly, any path from $\X$  to $\Y$ containing the motifs (5a) or (5b) is blocked given $\Opt\Sel$. Secondly, any path from  a valid adjustment set $\Z$  to $\Y$ containing the motifs (5a) or (5b) is blocked given $\X\Opt\Sel$ if $\Z$ does not contain any descendants of $N$ ($\Z\cap\des(N)=\emptyset$).
\end{mylemma}
\textit{Proof.}
First statement: $N\notin \ancs$ by definition of $N$ and, hence, in particular $N\notin\an(\X)$. Then, if a path exists, either the paths from $\X$ to $N$ end with a head at $N$ or there must be at least one collider $K$  with $N\in\an(K)$ and $K\notin\ancs$ on a path to $\X$. Now $N,K\notin \an(\Opt\Sel)$ can be seen by considering the different parts of $\Opt$: $N,K\notin \an(\Par\Sel)$ since $N,K\notin\ancs$ and $N,K\notin \an(C)$ for $C\in\ancs\cap\Col\Par_{\Col}$. Finally, $N,K\notin \an(C)$ for for $C\in\Col\Par_{\Col}$ with $C \ci \X ~|~\ancs$ because $N,K~\cancel{\ci}~\X~|~\ancs$. Hence, the path is blocked.
Second statement: 
If $\Z$ does not contain any descendants of $N$, then $N\notin\an(\Z)$. Then any path from a $\Z$ is blocked by the same reasoning as in the first part with the addition that $N\notin\an(\X)$ and hence the motif is blocked given $\X\Opt\Sel$.
\hfill $\square$

\begin{mylemma}[Existence of X-N-path and its openness given O-set] \label{thm:XN_path}
Given Assumptions~\ref{assum:opt} with $\Opt=\Par\Col\Par_{\Col}$ defined in Def.~\ref{def:def_opt}. There must be at least one path from $\X$ to $N$ (defined in the motifs (5a) or (5b)) that ends with a head at $N$ and where every collider is in $\ancs$ and every non-collider is not in $\ancs$, hence $\X \cancel{\ci} N|\ancs$. Further, for $N'\in \des(N)$ there is an open path from $N'$ to $\X$ given $\Opt\Sel$, hence $\X \cancel{\ci} N'|\Opt\Sel$.
\end{mylemma}
\textit{Proof.} First statement: By definition of the N-node, $N~\cancel{\ci}~\X~|~\ancs$. Now all paths that end with a tail at $N$ are blocked given $\ancs$ because $N\notin\an(\X)$ and the first collider $K$ coming from $N$ must be blocked because $K\notin\ancs$. Hence, there must be an open path that ends with a head at $N$ and where every collider is in $\ancs$ and every non-collider is not in $\ancs$ as stated. 
Second statement: We have to show that for $N'\in \des(N)$ there is a path to $\X$ where every collider is open given $\Opt\Sel$ and no non-collider is in $\Opt\Sel$. Consider the path in the first part from $\X$ to $N$, possibly extended by a directed path to $N'$. 
For all colliders $K$ we have that if $K\in \an(\Sel)$, the collider is always opened and if $K\in \an(\X)$, then either there exists an open path to $\X$ or, if a non-collider in this path is in $\Opt$, then the collider is open. Finally, if $K\in \an(\Y)$, the collider is open because every directed path from $K$ to $\Y$ either goes through $\Par\subseteq \Opt$ or $K\in \Par\subseteq \Opt$. It cannot be that $K\in \Y\M$ since then $K\notin \ancs$ and the path would be blocked while we consider the path from the first part to be open given $\ancs$.
Further, no non-collider $D$ on this path can be in $\Opt\Sel$: $D\notin\Par$ since $D\notin \ancs$  for the path from $\X$ to $N$ and a non-collider on the directed path $N\tailhead \cdots \tailhead N'$ cannot be in $\Par$ because $N\notin \ancs$. Finally, $D\notin \Col\Par_\Col$ because $D$ then has to fulfill either $D\in \ancs$, which cannot be as $D\notin \ancs$, or $D\ci \X | \ancs$ (Def.~\ref{def:def_collpath} part (2a,b)). The latter cannot be for $D$ occurring on the path from $\X$ to $N$ since this was shown to be open given $\ancs$, and for $D$ occurring on the directed path $N\tailhead \cdots \tailhead N'$  this cannot be because $N \cancel{\ci} \X | \ancs$ and no node on this directed path can be in $\ancs$ since $N\notin \ancs$.
Hence, $\X \cancel{\ci} N'|\Opt\Sel$.
\hfill $\square$

\subsection{Proof of Theorem~\ref{thm:validity}}
\begin{mythm*}[Validity of O-set] 
Given Assumptions~\ref{assum:opt} but \emph{without} a priori assuming that a valid adjustment set exists (apart from the requirement $\Sel\cap\forb=\emptyset$). If and only if a valid backdoor adjustment set exists, then $\Opt$ is a valid adjustment set.
\end{mythm*}
\textit{Proof.}
\textbf{``if''}: Given that a valid backdoor adjustment set exists, we need to prove that (i)~$\Opt\cap \forb=\emptyset$ with $\forb=\X\cup\des(\Y\M)$ and (ii)~all non-causal paths from $\X$ to $\Y$ are blocked by $\Opt$ (given $\Sel$). (i) is true by the construction of $\Opt$ in Def.~\ref{def:def_opt} and Alg.~\ref{algo:opt} where nodes $\in \des(\Y\M)$ are not added and nodes that are $\X$ indicate non-identifiability (see Lemma~\ref{thm:nonidentifiability}). By Lemma~\ref{thm:colliderparents} also $\Par_{\Col}\cap \des(\Y\M)=\emptyset$  and $\X\notin\Par_{\Col}$ because otherwise no valid adjustment set exists by Lemma~\ref{thm:nonidentifiability}.

Lemma~\ref{thm:motifs} lists all possible motifs on non-causal paths. By Lemma~\ref{thm:nonidentifiability} the occurrence of the motifs (1a) or (4a) or (4b) for $F=\X$ renders the effect non-identifiable, contrary to the assumption. Hence only the remaining motifs can occur. By Lemma~\ref{thm:parentchild_blocked} the motifs (1b), (2a), (2b), (3a), (3b) are blocked given $\Opt\Sel$. By Lemma~\ref{thm:desYM_blocked} (part one) the motifs (4a,b) for $F\in\des(\Y\M)$ are blocked given $\Opt\Sel$. By Lemma~\ref{thm:N_blocked} (part one) motifs (5a) and (5b) are blocked given $\Opt\Sel$.

\textbf{``only if''} is trivially true since  $\Opt$ is then assumed valid.
\hfill $\square$

\subsection{Proof of Theorem~\ref{thm:opt_vs_adjust}}
\begin{mythm*}[O-set vs Adjust-set
] 
Given Assumptions~\ref{assum:opt} with $\Opt$ defined in Def.~\ref{def:def_opt} and  the Adjust-set $\ancs$ defined in Eq.~\eqref{eq:adjust}, it holds that $J_\Opt\geq J_{\Adj}$ for any graph $\Gra$. We have $J_\Opt= J_{\Adj}$ only if $\Opt=\ancs$ or $\Opt\subseteq \ancs$ and $\X\ci \ancs\setminus \Opt~|~\Opt\Sel$.
\end{mythm*}
\textit{Proof.}
We directly use the decomposition in Eq.~\eqref{eq:decomp} with $\Z=\Opt=\A\B$ and $\Z'=\Adj=\B\R$ with $\Adj= \an(\X\Y\Sel)\setminus \forb$ and the definitions of $\R,\B,\A$ as in Eq.~\eqref{eq:decomp}. For term (iii), $I_{\R;\Y|\Opt \X \Sel}$, to be non-zero, there must be an active path from $\R\subseteq \ancs$ to $\Y$ given $\X\Opt\Sel$. By Lemma~\ref{thm:motifs}, Lemma~\ref{thm:parentchild_blocked}, Lemma~\ref{thm:desYM_blocked} (second part), and Lemma~\ref{thm:N_blocked} (second part), the only possibly open motifs on paths from $\R$ to $\Y$ given $\Opt \X \Sel$ are ``$\headtail N\headhead W$'' or ``$\headtail N\headhead \boxed{C}\headhead$'' where $\R\cap \des(N)\neq\emptyset$. But since $\R\subseteq\ancs $ and $N\notin \ancs$, $\R$ cannot contain descendants of $N$. Hence, term (iii) is zero. 
For term (iv), $I_{\X;\A|\B\R \Sel}=I_{\X;\A|\ancs}$, note that $\A=\Opt\setminus \ancs$ and, hence, for all $A\in\A$ it holds that $A\ci \X ~|~\ancs$ since all $A\in\A$ then fulfill Def.~\ref{def:def_collpath}(2b) (for $A\in \Par_\Col$ see Lemma~\ref{thm:colliderparents}). Hence, $I_{\X;\A|\ancs}=0$ by Markovity. This proves that $J_\Opt\geq J_{\Adj}$.

We are now left with terms (i) and (ii) in Eq.~\eqref{eq:decomp}. By construction of the collider path nodes, $\A\subseteq \Col\Par_\Col$ is connected to $\Y$ (potentially through $\M$) conditional on $\ancs\X$ since $\ancs$ contains all remaining collider nodes in $\Col$. Then by Faithfulness term (i) $I_{\A;\Y|\B\R \X\Sel}=I_{\A;\Y|\ancs\X}$ can only be zero if $\A=\emptyset$. Then $\Opt\subseteq \ancs$. Term (ii), $I_{\X;\R|\Opt\Sel}=0$ if $\R=\ancs\setminus\Opt=\emptyset$ or $\X\ci \ancs\setminus \Opt~|~\Opt\Sel$ together with Faithfulness.
\hfill $\square$

\subsection{Proof of Proposition~\ref{thm:minimal_ancs}}
\begin{myprop*}[Collider-minimized O-set is a subset of Adjust.] 
Given Assumptions~\ref{assum:opt} with $\Opt=\Par\Col\Par_{\Col}$ defined in Def.~\ref{def:def_opt} and the $\Optcmin$-set constructed with Alg.~\ref{algo:optmin} it holds that $\Optcmin\subseteq\ancs$.
\end{myprop*}
\textit{Proof.}
Define $\Colmin=\Optcmin\setminus\Par$. We need to show that $C\in\Colmin \Rightarrow C\in \ancs$ for all $C\in \Opt\setminus\Par$. Assume $C\notin \ancs$. Since then all $C\in \Opt\setminus\Par$ fulfill Def.~\ref{def:def_collpath}(2b) (for $C\in \Par_\Col$ see Lemma~\ref{thm:colliderparents}), it holds that $C \ci \X ~|~\ancs$ implying that no link $X\astast C$ exists. If a path exists at all, either (i)~there must be at least one collider $K$  with $C\in\an(K)$ and $K\notin\ancs$ on a path to $\X$ or (ii)~$C\in \des(\X)$. We now show that for case (i) $C$ has no open path to $\X$ given $\Sel\Opt\setminus\{C\}$. $K\notin \an(\Opt\Sel)$ can be seen by considering the different parts of $\Opt\Sel$: $K\notin \an(\Par\Sel)$ since $K\notin\ancs$ and $\an(\Par\Sel)\subseteq \ancs$. Further, $K\notin \an(\ancs\cap\Col)$. Finally, $K\notin \an(\Col\Par_\Col\setminus\ancs)$ since $C'\in\Col\Par_\Col\setminus\ancs$ fulfill (by  Def.~\ref{def:def_collpath}(2b)) $C' \ci \X ~|~\ancs$ and $K~\cancel{\ci}~\X~|~\ancs$. Hence, $\X ~\ci C ~|~\Sel\Opt\setminus\{C\}$ implying that $C$ would be removed in the first loop of Alg.~\ref{algo:optmin} and $C\notin \Colmin$, contrary to assumption.

In case (ii) the directed path from $\X$ to $C$ for $C\in\Col\setminus\Par_\Col$ is blocked because $\Par_\Col\subseteq\Opt$ contains all parents of $C$ and $X\notin\Par_\Col$ since we assume identifiability. This implies that $C$ would be removed in the first loop of Alg.~\ref{algo:optmin} and $C\notin \Colmin$, contrary to assumption. Finally, if there exists a directed path from $\X$ to $C=P_C\in\Par_\Col\setminus\Col$ for $P_C\notin\ancs$ we know that all children $C\in\ch(P_C)\cap \Col\Par$ were removed in the first loop of Alg.~\ref{algo:optmin}. Denote the remaining nodes after the first loop of Alg.~\ref{algo:optmin} by $\Optcmin'$. $P_C\notin\ancs$ has no directed path to $\Y$ and is separated from $\Y$ given $\Sel\Optcmin'$ because the motif $P_C\tailhead C\headhead $ is blocked since $C\notin\an(\Optcmin')$. This implies that $P_C$ would be removed in the second loop of Alg.~\ref{algo:optmin} and $P_C\notin \Colmin$, contrary to assumption. 
\hfill $\square$

\subsection{Proof of Theorem~\ref{thm:graph_optimality}}
\begin{mythm*}[Necessary and sufficient graphical conditions for optimality and optimality of O-set] 
Given Assumptions~\ref{assum:opt} and with $\Opt=\Par\Col\Par_\Col$ defined in Def.~\ref{def:def_opt}. Denote the set of N-nodes by $\mathbf{N}=\spouse(\Y\M\Col)\setminus (\forb\Opt\Sel)$. Finally, given an $N\in \mathbf{N}$ and a collider path $N\headhead \cdots \headhead C\headhead \cdots\headhead W$ (including $N\headhead W$) for $C\in \Col$ and $W\in \Y\M$ (indexed by $i$) with the collider path nodes denoted by $\pi^N_i$ (excluding $N$ and $W$), denote by $\Optnpi=\Opt(\X,\Y,\Sel'=\Sel N \pi^N_i)$ the O-set for the causal effect of $\X$ on $\Y$ given $\Sel'=\Sel\cup \{N\} \cup \pi^N_i$. 

If and only if exactly one valid adjustment set exists, or both of the following conditions are fulfilled, then graphical optimality holds and $\Opt$ is optimal:

(I)~For \emph{all} $N\in \mathbf{N}$ and all its collider paths $i$ to $W\in Y\M$ that are inside $\Col$ it holds that $\Optnpi$ does not block all non-causal paths from $\X$ to $\Y$, i.e., $\Optnpi$ is non-valid,   

and

(II)~for all $E\in \Opt \setminus \Par$ with an open path to $\X$ given $\Sel \Opt \setminus \{E\}$ there is a link $E \headhead W$ or an extended collider path $E\asthead C \headhead \cdots \headhead W$ inside $\Col$ for $W\in \Y\M$ where all colliders $C\in \ancs$.
\end{mythm*}

\textit{Proof.}
If exactly one valid adjustment set exists, then optimality holds by Def.~\ref{def:def_gra_opt} and  then this set is $\Opt$ because $\Opt$ is always valid if a valid set exists (Lemma~\ref{thm:validity}).

The proof is based on Lemma~\ref{thm:existence} and relation~\eqref{eq:decomp}. We will first prove the ``if''-statement by showing that Cond.~(I) leads to term~(i)$\geq$(iii) and Cond.~(II) leads to term~(ii)$\geq$(iv) from which optimality follows by Lemma~\ref{thm:existence}. Then we prove the ``only if''-statement  by showing that if either of the two conditions is not fulfilled, then for every adjustment set there exists an alternative set such that (i)$<$(iii) or (ii)$<$(iv) for some distribution $\Prob$ consistent with $\Gra$. This implies that graphical optimality does not hold.

\textbf{``if''}: We have to show that if both conditions hold, then $I_{\A;\Y|\B\R \X\Sel} \geq I_{\R;\Y|\A\B \X \Sel}$ and $I_{\X;\R|\A\B \Sel} \geq I_{\X;\A|\B\R \Sel}$ where $\Opt=\A\B$ and $\Z'=\R\B$ with $\B=\Opt\cap\Z'$. Further, we use $\A_\Par=\A\cap\Par$ and $\A_\Col=(\A\cap \Col\Par_\Col)\setminus\A_\Par$ where $\A=\A_\Par\cup \A_\Col$. 

Condition~(I) directly leads to $I_{\A;\Y|\B\R \X\Sel} \geq I_{\R;\Y|\A\B \X \Sel}$ as follows. 

We subdivide condition~(I) into two cases where the former implies the latter: (I.1)~There are no N-nodes, i.e., $\mathbf{N}=\emptyset$, or (I.2)~for \emph{all} $N\in \mathbf{N}$ and all its collider paths $i$ it holds that $\Optnpi$ does not block all non-causal paths from $\X$ to $\Y$.

If condition (I.1) holds, then there are no N-nodes. If there are no N-motifs on any path from $\R$ to $\Y$, then by Lemma~\ref{thm:motifs}, Lemma~\ref{thm:parentchild_blocked}, and Lemma~\ref{thm:desYM_blocked} (second part) all paths given $\X\Opt\Sel$ are blocked and term (iii) is zero by Markovity.

If condition~(I.2) holds, then there are N-nodes. By Lemma~\ref{thm:N_blocked} (second part) the only possibly open motifs on paths from $\R$ to $\Y$ given $\Opt \X \Sel$ are ``$\headtail N\headhead W$'' or ``$\headtail N\headhead \boxed{C}\headhead$'' where $\R\cap \des(N)\neq\emptyset$. Term (iii), $I_{\R;\Y|\B \X \Sel \A}=I_{\R;\Y|\X \Sel \Opt}$, is then always non-zero since, by definition of the N-nodes, there exists at least one collider path $N\headhead \cdots \headhead \boxed{C}\headhead \cdots\headhead W$ (including $N\headhead W$) for $C\in \Col$ and $W\in \Y\M$. To see under which conditions still term~(i)$\geq$(iii) consider two ways of decomposing the following CMI:
\begin{align} \label{eq:condI_decomp}
  I_{\A\R;\Y|\B \X \Sel} &= \underbrace{I_{\A;\Y|\B \X \Sel}}_{\text{term (i')}} + \underbrace{I_{\R;\Y|\B \X \Sel \A}}_{\text{term (iii)}} \nonumber \\ 
  &= \underbrace{I_{\R;\Y|\B \X \Sel}}_{\text{term (iii')}} + \underbrace{I_{\A;\Y|\B \X \Sel \R}}_{\text{term (i)}}\,.
\end{align}
From this decomposition we see that term~(i)$\geq$(iii) if and only if term~(i')$\geq$(iii'). Paths from $\R$ to $\Y$ via $\X$ given $\Sel\X\Z'\setminus\R=\B\Sel\X$ are blocked because if $\X$ is a collider, then there would be a non-causal path rendering $\Z'$ invalid. Therefore, for term (iii') to be non-zero $\Z'\Sel$ must contain at least descendants of an N-node $N$ and all its collider path nodes towards $W$, denoted $\pi^N_i$, for at least one path $i$. Then $\R\cap \des(N)\neq\emptyset$ and $\pi^N_i \subseteq \B\Sel$ such that there exists an open path ``$ N\headhead \boxed{C}\headhead \cdots \headhead \boxed{C}\headhead W$'' (or $N\headhead W)$.

Condition~(I.2) now guarantees that for all $N\in \mathbf{N}$ and all collider paths indexed by $i$ the O-set $\Optnpi$, which includes $N\pi^N_i$ as a subset, does \emph{not} block all non-causal paths. By Theorem~\ref{thm:validity}, if $\Optnpi$ is not valid, then no valid adjustment set $\Z'$ containing $N\pi^N_i$ as a subset exists. And this in turn implies that no valid set with $\R\cap \des(N)\neq\emptyset$ exists. To show this, assume the contraposition: If there was such a valid set $\Z'$ with $\R\cap \des(N)\neq\emptyset$ and $\pi^N_i\subset\Z'$, then it would open the collider motif $\asthead N \headhead$ since $\R$ contains descendants of $N$ and lead to an open path ``$ N\headhead \boxed{C}\headhead \cdots \headhead \boxed{C}\headhead W$'' (or $N\headhead W)$. If $\Z'$ is still valid, it must block all paths from $\X$ that end with an arrowhead at $N$. But then also $\Z'\cup \{N\}$ is valid.
Note that since $N\notin \forb$, $\Sel\cap\forb=\emptyset$, and $\pi^N_i\cap\forb=\emptyset$ since $\pi^N_i\subseteq \Col$, the validity of $\Optnpi$ depends only on its ability to block non-causal paths. Hence, term~(iii') is zero and by Eq.~\eqref{eq:condI_decomp} term~(i)$\geq$(iii).

Condition~(II) directly leads to $I_{\X;\R|\A\B \Sel} \geq I_{\X;\A|\B\R \Sel}$ as follows. 

Define $\E=\{E\in \Opt \setminus \Par: \X \cancel{\ci} E ~|~\Sel \Opt \setminus \{E\}\}$. By condition~(II) there exists a link $E \headhead W$ or an extended collider path $E\asthead C \headhead \cdots \headhead W$ inside $\Col$ for $W\in \Y\M$ where all colliders $C\in \ancs$. There are two types: (1)~$E\tailhead C \headhead \cdots \headhead W$ (then $E\in \Par_{\Col}$) and (2)~ $E \headhead W$ or $E\headhead C \headhead \cdots \headhead W$. We consider two cases:

Case~(1): $E\in \E$ for which there exists \emph{at least one} path of type~(1). Any valid $\Z'$ with $E\notin \Z'$ has to block paths from $\X$ to $E$ since otherwise there is a non-causal open path from $\X$ to $\Y$ through the motif chain $\astast E\tailhead C \headhead \cdots \headhead W$ for $W\in \Y\M$: $E$ is open since $E\notin \Z'$ and the part from $E$ to $W$ is open since all colliders $C\in \ancs$: if $C\in \an(\Sel)$, the collider is always opened and if $C\in \an(\X\Y)$  then either the directed path to $\X$ or $\Y$ is open, or $C$ is opened if $\Z'$ contains a node on that path. 

Case~(2): $E\in \E$ for which \emph{all paths} are of type~(2). Firstly, all paths from $\X$ to $E$ that end with a tail at $E$ must be blocked by $\Z'$ since otherwise there is a non-causal path as for case~(1). The same holds for paths that end with a head at $E$ if $E\in \ancs$. Consider paths that end with a head at $E$ and $E\notin \ancs$ which implies $E\ci \X ~|~\ancs$ by Def.~\ref{def:def_collpath}. Then it follows that $E\ci \X~|~\Sel \Opt\setminus \{E\}$ and, hence, $E\notin \E$ which we can show by considering where $E$ can occur with respect to the different motifs listed in Lemma~\ref{thm:motifs} as follows (see the definitions of $W,V,F,N,C,P_C$ there): Motif~(1a)~``$\astast X\tailhead \boxed{C} \headhead$'' is not relevant since then non-identifiability holds and motif (2a)~``$X,V\astast\boxed{P}\tailhead W$'' is not relevant since $\E\notin \Par$. Motifs (3a)~``$V\headtail W$'', (4a)~``$\astast F\headhead W$'', and (5a)~``$\astast N\headhead W$'' are not relevant since no $E\in \Opt$ is involved. For the motifs (1b)~``$\astast X\tailhead \boxed{P_C}\tailhead E\headhead$'', (2b)~``$X,V\astast\boxed{P_C}\tailhead \boxed{C} \headhead\cdots \headhead E$'', and (3b)~``$X,V\headtail \boxed{C} \headhead \cdots \headhead E$'' the path to $\X$ is blocked by $\Sel \Opt\setminus \{E\}$. For motif (4b)~``$\astast F\headhead\boxed{C}\headhead\cdots \headhead E$'' or ``$\astast F\headhead E$'', since  $\Sel \Opt\cap \forb = \emptyset$ and $\X\cap \des(\forb)=\emptyset$, there must exist a collider $\in \forb$ or $\asthead F \headhead$ on a path to $\X$ which is then blocked. Hence, $E\notin \E$. Finally, for (5b)~``$\astast N\headhead \boxed{C}\headhead\cdots \headhead E$'' or ``$\astast N\headhead E$'' with $N\notin \ancs$ and $\X \cancel{\ci} N ~|~\ancs$ we either have $\asthead N\headhead$ or there exists a collider on any path to $\X$ with $K\in \des(N)$ and, hence, $K\notin \ancs$. $E \cancel{\ci} \X~|~\Sel \Opt\setminus \{E\}$ would only be possible if $N$ or $K\in \an(\Opt\setminus \ancs)$. The subset $\Opt\setminus \ancs$ fulfills $\Opt\setminus \ancs\ci \X ~|~\ancs$ by Def.~\ref{def:def_collpath}. However, $N$ or $K\in \an(\Opt\setminus \ancs)$ implies that there is a path from $\Opt\setminus \ancs$ to $N$. Then $\X \cancel{\ci} N ~|~\ancs$ contradicts $\Opt\setminus \ancs\ci \X ~|~\ancs$ implying that $N,K\notin \an(\Opt\setminus \ancs)$ and, hence, $E \ci \X~|~\Sel \Opt\setminus \{E\}$ and  $E\notin \E$.

Both cases taken together, it holds that $\X \ci E~|~\Sel\Z'\setminus \{E\}$ for any valid $\Z'$.
Furthermore, $\X \ci P~|~\Sel\Z'\setminus \{P\}$ with $P\in \Par$ for any valid $\Z'$ since $\Par$ is directly connected to $\Y$ and, therefore, a valid $\Z'$ has to block a non-causal path between $\X$ and $\Y$ through $\Par$. 

Now decompose term (iv) as 
\begin{align} \label{eq:termiv}
I_{X;\A|\Z' \Sel}&=\underbrace{I_{X;\A_\Par\A_\E|\Z' \Sel}}_{=0} + I_{X;\A\setminus (\A_\Par\A_\E)|\Z' \Sel \A_\Par\A_\E}
\end{align}
with $\A_\Par=\A\cap\Par$ and $\A_\E=(\A\cap \E)\setminus\A_\Par$. The preceding derivations imply $\X \ci \A_\Par\A_\E|\Z' \Sel$ for any valid $\Z'$ and, hence, the first term vanishes.

Consider the set $\E'=\{E'\in \Opt \setminus \Par: \X \ci E' ~|~\Sel \Opt \setminus \{E'\}\}$. This implies that $\A_{\E'}=\A\setminus (\A_\Par\A_\E)$ fulfills $\A_{\E'} \ci \X ~|~\Sel \Opt\setminus \A_{\E'}$ and since $\Sel \Opt\setminus \A_{\E'}=\Sel\B \A_\Par\A_\E$ we have 
\begin{align} \label{eq:eprime_vanishes}
 I_{\X;\A_{\E'} |~\Sel\B \A_\Par\A_\E} = 0\,.
\end{align}
This now leads to term~(ii) $\geq$ term~(iv) by considering two ways of decomposing the following CMI:
\begin{align}
I_{\X;\R \A_{\E'}|\Sel\B \A_\Par\A_\E} 
    &= \underbrace{I_{\X;\A_{\E'}|\Sel\B \A_\Par\A_\E}}_{= 0 ~~\text{by Eq.~\eqref{eq:eprime_vanishes}}} + \underbrace{I_{\X;\R|\Sel\B \A_\Par\A_\E\A_{\E'}}}_{\text{term (ii)}} \\ 
  &= \underbrace{I_{\X;\R|\Sel\B \A_\Par\A_\E}}_{\geq0} + \underbrace{I_{\X;\A_{\E'}|\Sel\B \A_\Par\A_\E\R}}_{\text{term (iv) by Eq.~\eqref{eq:termiv}}}\,.
\end{align}

\textbf{``only if''}: 
We need to prove that if either Condition (I) or Condition (II) or both are not fulfilled, then for every valid adjustment set $\Z$ (including $\Opt$) there exists a valid set $\Z'$ and a distribution $\Prob$ compatible with $\Gra$ such that $J_{\Z}<J_{\Z'}$, i.e., with a \emph{strictly larger} adjustment information, implying that graphical optimality does not hold. We separate the proof into adjustment sets $\Z$ with $\Opt\setminus \Z\neq \emptyset$ for which the adjustment set $\Z'=\Opt$ together with a suitably constructed distribution $\Prob$ fulfills $J_{\Z}<J_{\Z'}$ and $\Opt\setminus \Z = \emptyset$ for which either $\Z'=\Opt$ or $\Z'=\Optnpi$ fulfills $J_{\Z}<J_{\Z'}$, depending on further case distinctions as discussed below.

We first consider adjustment sets $\Z$ with $\Opt\setminus \Z\neq \emptyset$, i.e. adjustment sets that are \emph{not} supersets of the $\Opt$-set. Consider $\Z'=\Opt$, which is valid by Thm.~\ref{thm:validity}, and implies $\R\neq \emptyset$ in the notation used throughout this paper. For all $\Z$ we can actually show that there always exists a distribution $\Prob$ such that $J_{\Z}<J_{\Z'}$, irrespective of whether Condition~(I) and/or (II) holds or not. 

Since $\R \subseteq \Opt=\B\R$, at least one $R\in\R$ has an open path to $\Y$ given $\A\B \X \Sel$, either because $R\in \Par$, or, if $R\in \Col \Par_{\Col}$, then, for at least one $R\in\R$, there is an open  collider path (or link) $R\asthead \boxed{C} \headhead \cdots \headhead \boxed{C} \headhead W$ for $W\in\Y\M$ since all colliders $C\in\B\Sel$. Hence it holds that $I_{\R;\Y|\A\B \X \Sel}>0$. 
Further, if $\A=\emptyset$, then immediately term~(i), $I_{\A;\Y|\B \R \X \Sel}=0$. Alternatively, if $\A\neq\emptyset$, we construct a distribution $\Prob$ with associated SCM consistent with $\Gra$ where all links ``$U\astast A$'' for $A\in\A$ \emph{almost vanish} and, hence, term~(i), $I_{\A;\Y|\B \R \X \Sel}\to 0$: Consider the three possible links and associated arbitrary assignment functions in the SCM: (1)~``$A\tailhead U$'' with $U:=f_{U}(\ldots,c A,\ldots)$, (2)~``$A\headtail U$'' with $A:=f_{A}(\ldots,c U,\ldots)$, and (3)~``$A\headhead U$'' with $A:=f_{A}(\ldots,c L^U,\ldots)$ where $L^U$ denotes one or more latent variables and $c\in \mathbb{R}$. We then consider the limit $c\to 0$ leading to term~(i), $I_{\A;\Y|\B \R \X \Sel}\to 0$ and then term~(i)$<$(iii). By Lemma~\ref{thm:existence}, where $\Prob$ is further modified to $\Prob'$ without affecting term~(i)$<$(iii), then graphical optimality does not hold.

Secondly, we consider adjustment sets $\Z$ with $\Opt\setminus \Z = \emptyset$, i.e. adjustment sets that are supersets of the $\Opt$-set (or the $\Opt$-set itself) and separately consider the cases that either Condition (I) or Condition (II) are not fulfilled. If both are not fullfilled, then either of the alternative adjustment sets $\Z'$ considered below can be used.

For the case that Condition~(I) does not hold, there exists at least one N-node. Now further divide the valid adjustment sets into those with $\Z\cap \des(N)\neq \emptyset$ and $\Z\cap \des(N)= \emptyset$.

In case of the former (where $\Z\neq\Opt$), consider the $\Z'=\Opt$ (valid by Thm.~\ref{thm:validity}). Then $\R=\emptyset$ and term~(ii), $I_{\X;\R|\A\B \Sel}=0$. On the other hand, $\A\neq \emptyset$ and by Lemma~\ref{thm:XN_path} there exists an open path from $N'\in \A \cap \des(N)$ to $\X$ given $\B\R\Sel=\Opt\Sel$ such that $I_{\X;\A|\R\B \Sel}>0$. Then term~(ii)$<$(iv) for all $\Prob$ that are faithful to $\Gra$. By Lemma~\ref{thm:existence}, where $\Prob$ is modified to $\Prob'$ without affecting term~(ii)$<$(iv), then graphical optimality does not hold.

In case of the latter, $\Z\cap \des(N)= \emptyset$ (which includes $\Opt$), consider $\Z'=\Optnpi$.  $\Optnpi$ is the O-set for the causal effect of $\X$ on $\Y$ given $\Sel'=\Sel\cup \{N\} \cup \pi^N_i$. By the negation of Condition~(1) there exists at least one N-node with at least one collider path $N\headhead \cdots \headhead C\headhead \cdots\headhead W$ (including $N\headhead W$) for $C\in \Col$ and $W\in \Y\M$ (indexed by $i$) with collider path nodes denoted $\pi^N_i$ such that $\Optnpi$  blocks all non-causal paths from $\X$ to $\Y$. Since also $N\notin \forb$, $\Sel\cap\forb=\emptyset$, and $\pi^N_i\cap\forb=\emptyset$, $\Z'=\Optnpi$ is valid. 
Since $N\in \Optnpi$ while $N\notin \Z$, we have $N\in\R\neq \emptyset$, and since $\pi^N_i\subseteq \Optnpi$ we have $\pi^N_i \subseteq \B\R\Sel$ and there exists an open path $ N\headhead \boxed{C}\headhead \cdots \headhead \boxed{C}\headhead W$ (or $N\headhead W$) where every $C\in\B$ or $C\in\R$ such that $I_{\R;\Y|\A\B \X \Sel}>0$. 
As above, either $\A=\emptyset$ and term~(i) $I_{\A;\Y|\B \R\X \Sel}=0$, or we can construct a distribution $\Prob$ with associated SCM consistent with $\Gra$ where all links ``$U\astast A$'' for $A\in\A$ \emph{almost vanish} and, hence, term~(i), $I_{\A;\Y|\B \R\X \Sel}\to 0$. Since $\A\cap N\pi^N_i=\emptyset$ this does not affect the collider path $N\headhead \boxed{C}\headhead \cdots \headhead \boxed{C}\headhead W$ (or $N\headhead W$) such that $I_{\R;\Y|\B\R \X \Sel}>0$. Then term~(i)$<$(iii) and by Lemma~\ref{thm:existence}, where $\Prob$ is further modified to $\Prob'$ without affecting term~(i)$<$(iii), then graphical optimality does not hold. As mentioned above, for this distribution also $J_{\Opt}<J_{\Optnpi}$.

Now consider the case that Condition~(II) does not hold. We have $\Z=\Opt \cup \A'$ which includes $\Z=\Opt$ for $\A'=\emptyset$. By the negation of Condition~(II) there exists an $E\in \Opt \setminus \Par$ with $\X \cancel{\ci} E ~|~\Sel \Opt \setminus \{E\}$ such that there is no link $E \headhead W$ and all extended collider paths $E\asthead C \headhead \cdots \headhead W$ inside $\Col$ for $W\in \Y\M$ contain at least one collider $C\notin \ancs$. Define the set of these non-ancestral colliders as
\begin{align} \label{eq:ColE}
\Col_E= \{C\in \Col: E\asthead \cdots \headhead C \headhead \cdots \headhead W \}\setminus \ancs\,.
\end{align}
We define $\ECol=\{E\}\cup(\des(\Col_E)\cap \Opt)$ and choose $\Z'=\Opt \setminus \ECol$ implying $\A=\ECol\cup \A'$, $\B=\Opt \setminus \ECol$, and $\R=\emptyset$. We need to show that (1)~$\Z'$ is valid and (2)~$I_{\X;\A|\B\R \Sel}=I_{\X;\ECol\A'|\Sel \Opt \setminus \ECol} > I_{\X;\R|\A\B \Sel}=0$ (since $\R=\emptyset$). 

Ad~(1): As a subset of $\Opt$ we have that $\Z'\cap \forb=\emptyset$. We investigate whether $\Z'$ blocks all non-causal paths between $\X$ and $\Y$ by considering the motifs in Lemma~\ref{thm:motifs}. In addition to all those motifs listed there, there are modified motifs where unconditioned $C$-nodes and $P_C$-nodes occur (denoted without a $\boxed{\cdot}$) due to removing $\ECol$ from $\Opt$. 

Firstly, the unmodified motifs are blocked as before (see Theorem~\ref{thm:validity}): Motif~(1a)~``$\astast X\tailhead \boxed{C} \headhead$'' is not relevant since then non-identifiability holds. By Lemma~\ref{thm:parentchild_blocked} the motifs (1b), (2a), (2b), (3a), (3b) all contain a non-collider in $\Sel\Opt \setminus \ECol$ and are blocked. By Lemma~\ref{thm:desYM_blocked} (part one) the motifs (4a,b) for $F\in\des(\Y\M)$ are blocked because $\Z'\cap \forb=\emptyset$. By Lemma~\ref{thm:N_blocked} (part one) motifs (5a) and (5b) are blocked given $\Sel\Opt \setminus \ECol$ because the proof in Lemma~\ref{thm:N_blocked} requires that on paths to $\X$ either $N$ is a collider or there exists a descendant collider $K$ and that $N,K\notin \an (\Opt\Sel)$. The latter is fulfilled because $\Sel\Opt \setminus \ECol$ is a subset of $\Sel\Opt$.

Secondly, all paths from $\X$ through the removed node $E$ to $W\in\Y\M$ are blocked by $\Sel\Opt \setminus \ECol$: Paths through $\Par$ are blocked since $E\notin \Par$ and $\des(\Col_E)\cap \ancs=\emptyset$ and, hence, $\Par \subseteq \Sel\Opt \setminus \ECol$. Paths through colliders are blocked by the negation of condition~(II): there is no link $E \headhead W$ and all extended collider paths $E\asthead C \headhead \cdots \headhead W$ inside $\Col$ for $W\in \Y\M$ contain at least one collider $C\notin \ancs$. By construction, $\ECol=\{E\}\cup(\des(\Col_E)\cap \Opt)$, implying that all these non-ancestral colliders are blocked. 

Thirdly, we consider the modified motifs with unconditioned $C,P_C\in(\des(\Col_E)\cap \Opt)$. By definition of $\Col_E$ in \eqref{eq:ColE}, $C,P_C\notin \ancs$. (As a remark, $E$ can potentially be in $\ancs$.)
Motif~(1a)~``$\astast X\tailhead \boxed{C} \headhead$'' cannot occur since then non-identifiability holds.
Modified motifs (1,2b')~``$X,V\astast\boxed{P_C}\tailhead C \headhead$'' and (1,2b'')~``$X,V\astast\boxed{P_C}\tailhead \boxed{C}\headhead \cdots \headhead C \headhead$'' are blocked since they contain a conditioned non-collider. 
Motifs (1,2b''')~``$X,V\astast P_C\tailhead C\headhead$'' are blocked since $\Sel\Opt \setminus \ECol= \Sel\Opt \setminus (\{E\}\cup(\des(\Col_E)\cap \Opt))$ does not contain any descendant of $C$. 
Motif (2a')~``$X,V\astast P\tailhead W$'' is not possible since $P\in \Par\subseteq\ancs$. 
Motifs (3a)~``$V\headtail W$'', (4a)~``$\astast F\headhead W$'', and (5a)~``$\astast N\headhead W$'' are not modified since no conditioned node occurs. 
Motif~(3b')~``$X,V\headtail C \headhead$'' is blocked because due to $C\notin \ancs$ there must exist a descendant of $C$  that is a collider $K\notin \ancs$ on the path to $\X$. Since $\ECol$ contains all descendants of $C$, also $K$ and all its descendants are not in $\Sel\Opt \setminus \ECol$ and $K$ is blocked. 
Finally, motifs (4b')~``$\astast F\headhead C \headhead$'' and (5b')~``$\astast N\headhead C \headhead$'' are blocked since $\Sel\Opt \setminus \ECol$ does not contain any descendant of $C$. This proves the validity of $\Z'$.

Ad~(2): To show that $I_{\X;\A|\B\R \Sel}=I_{\X;\ECol\A'|\Sel \Opt \setminus \ECol}>I_{\X;\R|\A\B \Sel}=0$, we decompose $I_{\X;\ECol\A'|\Sel \Opt \setminus \ECol}=I_{\X;\ECol|\Sel \Opt \setminus \ECol}+I_{\X;\A'|\Sel \Opt}$ and prove that at least the first term is non-zero. We start from the assumption in the negation of Condition~(II) that $E \cancel{\ci} \X~|~\Sel \Opt\setminus \{E\}$. This implies that there exists a path from $\X$ to $E$ where no non-collider  is in $\Sel \Opt\setminus \{E\}$ and for every collider $K$ it holds that $\des(K)\cap \Sel \Opt\setminus \{E\} \neq \emptyset $. With $\Z'=\Opt \setminus \ECol$ all non-colliders are still open. Consider those colliders $K$ with $\des(K)\cap (\Opt\setminus \{E\}) \subseteq \ECol\setminus \{E\}$. Then these colliders are closed on the path from $E$ to $\X$. However, for each such $K$ there is a $C\in \ECol\setminus \{E\}$ with $C\in \des(K)$. Then the path from $\X$ through $\asthead K \tailhead \cdots \tailhead C$ is open given $\Sel\Opt \setminus \ECol$. Hence, at least for the last such collider on the path from $E$ to $\X$ there is an open path from $C\in \ECol\setminus \{E\}$ to $\X$ given $\Sel\Opt \setminus \ECol$. 
Then Faithfulness implies that $I_{\X;\ECol|\Sel \Opt \setminus \ECol} > 0$ and, hence, term~(ii) $<$ term~(iv) holds for all distributions $\Prob$ consistent with $\Gra$.
By Lemma~\ref{thm:existence}, where the distribution $\Prob$ is modified to $\Prob'$ without affecting term~(ii)$<$(iv), then graphical optimality does not hold. Note that this also proves that $J_{\Opt}<J_{\Opt\setminus\ECol}$.

This concludes the proof of Theorem~\ref{thm:graph_optimality}.
\hfill $\square$

\subsection{Proof of Corollary~\ref{thm:minimality}}

\begin{mycor*}[Minimality and minimum cardinality] 
Given Assumptions~\ref{assum:opt}, assume that graphical optimality holds, and, hence, $\Opt$ is optimal. Further it holds that:
\begin{enumerate}
\item  If $\Opt$ is not minimal, then  $J_{\Opt}> J_{\Z}$  for all \emph{minimal} valid $\Z\neq\Opt$,

\item If $\Opt$ is minimal valid, then $\Opt$ is the unique set that maximizes the adjustment information $J_\Z$ among all \emph{minimal} valid $\Z\neq\Opt$,

\item  $\Opt$ is of minimum cardinality, that is, there is no subset of $\Opt$ that is still valid and optimal.
\end{enumerate}
\end{mycor*}
\textit{Proof.}
We again define disjunct sets $\R,\B,\A$ with $\A=\Opt\setminus \Z$, $\R=\Z \setminus \Opt$, and $\B=\Opt\cap\Z$, where any of them can be empty, but not both $\R$ and $\A$ since then $\Z=\Opt$. Hence $\Opt=\A\B$ and $\Z=\B\R$. Consider relation~\eqref{eq:decomp} in this case,
\begin{align}
 J_\Opt &=J_{\Z}  \nonumber \\
   &\phantom{=}+  \underbrace{I_{\A;\Y|\B\R \X\Sel}}_{\text{(i)}} +  \underbrace{I_{\X;\R|\A\B \Sel}}_{\text{(ii)}}  - \underbrace{I_{\R;\Y|\A\B \X \Sel}}_{\text{(iii)}} - \underbrace{I_{\X;\A|\B\R \Sel}}_{\text{(iv)}}\,.
\end{align}

Part 1 and 2: Since graphical optimality holds, we know that $J_{\Opt}=J_{\Z}$ can only be achieved if  term~(i) $=$ term~(iii) and term~(ii) $=$ term~(iv). From Eq.~\eqref{eq:condI_decomp} we know that term~(i) $=$ (iii) can only hold if $I_{\A;\Y|\B\X\Sel}=0$. But this implies $\A=\emptyset$ by Faithfulness since, by construction,  $\A\subset\Opt$ is always connected to $\Y$ (potentially through $\M$) given $\X\Sel\Opt\setminus\A$. Then term~(iv) $=0$ and, by optimality, $I_{\X;\R|\emptyset\B \Sel}=0$. But the latter would imply that $\Z=\B\R$ is either not minimal anymore since $\R$ is not connected to $\X$ and, hence, does not block any non-causal path not already blocked by $\B$. Then $J_{\Opt} > J_{\Z}$ among all minimal valid $\Z$ (Part 1). Or $\Z$ is minimal and $\R=\emptyset$, for which $\Z=\Opt$ is the unique set maximizing $J_\Z$ among all minimal valid $\Z\neq\Opt$ (Part 2).   

Part 3, i.e., that removing any subset from $\Opt$ decreases $J_\Opt$ follows directly from setting $\R=\emptyset$ and considering $\A\neq \emptyset$ (since otherwise nothing would be removed). Then term~(ii) and term~(iii) are both zero and by optimality term~(iv), which must be smaller or equal to term~(ii), is zero. Since $\A$ is connected to $\Y$ (see Part 1) by Faithfulness we have $J_{\Opt} > J_{\Opt\setminus\A}$.
\hfill $\square$

\clearpage
\section{Algorithms}

\begin{algorithm}[h!]
\caption{Construction of $\Opt$-set and test for backdoor-identifiability.}
\begin{algorithmic}[1]
\Require Causal graph $\Gra$, cause variable $\X$, effect variable $\Y$, mediators $\M$, conditioned variables $\Sel$
\State Initialize $\Par=\emptyset$, $\Col=\emptyset$ and $\Par_{\Col}=\emptyset$
\For {$W\in\Y\M$}
    \State $\Par = \Par \cup \pa(W)\setminus \forb$
\EndFor
\For {$W\in\Y\M$}
    \State Initialize nodes in this level $\mathcal{L}=\{W\}$
    \State Initialize ignorable nodes $\mathcal{N}=\emptyset$  
    \While{$|\mathcal{L}|>0$}
        \State Initialize next level $\mathcal{L}'=\emptyset$
        \For {$C\in \spouse(\mathcal{L})\setminus\mathcal{N} $}
            \If {$C=\X$}
                \State \Return No valid backdoor adjustment set exists.
            \EndIf
            \If {$C\notin \Col$ and Def.~\ref{def:def_collpath} (1)~$C\notin \forb$ and ((2a)~$C \in \ancs$ or (2b)~$C \ci \X ~|~\ancs$)}
                \State $\Col = \Col \cup \{C\}$
                \State $\mathcal{L}' = \mathcal{L}' \cup \{C\}$
            \Else 
               \If {$C\notin \Col$ }
                    \State $\mathcal{N}=\mathcal{N} \cup \{C\}$
               \EndIf 
            \EndIf
        \EndFor
        \State $\mathcal{L}=\mathcal{L}'\setminus \mathcal{N}$
    \EndWhile
\EndFor
\For {$C\in\Col$}
    \If {$\X \in \pa(C)$}
      \State \Return No valid backdoor adjustment set exists.
    \EndIf
    \State $\Par_{\Col} = \Par_{\Col} \cup \pa(C)$ 
\EndFor
\State \Return $\Opt=\Par\Col\Par_{\Col}$
\end{algorithmic}
 \label{algo:opt}
\end{algorithm}

\begin{algorithm}[h!]
\caption{Construction of $\Opt_{\rm min}$ and $\Opt_{\rm Cmin}$-sets. The relevant code for $\Opt_{\rm Cmin}$ is indicated in parentheses.}
\begin{algorithmic}[1]
\Require Causal graph $\Gra$, cause variable $\X$, effect variable $\Y$, mediators $\M$, conditioned variables $\Sel$, $\Opt=\Par\Col\Par_{\Col}$-set
\State Initialize $\Opt_{\rm min}=\Opt$ ($\Col_{\rm min}=\Col\Par_{\Col}\setminus\Par$)
\For {$Z\in \Opt_{\rm min}$ ($Z\in \Col_{\rm min}$)}
 \If {$Z$ has no active path to $\X$ given $\Sel\Opt\setminus\{Z\}$}
    \State Mark $Z$ for removal
 \EndIf      
\EndFor
\State Remove marked nodes from $\Opt_{\rm min}$ ($\Col_{\rm min}$)
\For {$Z\in \Opt_{\rm min}$ ($Z\in \Col_{\rm min}$)}
 \If {$Z$ has no active path to $\Y$ given $\X\Sel\Opt_{\rm min}\setminus\{Z\}$ (given $\X\Sel\Par\Col_{\rm min}\setminus\{Z\}$)}
    \State Mark $Z$ for removal 
 \EndIf      
\EndFor
\State Remove marked nodes from $\Opt_{\rm min}$ ($\Col_{\rm min}$)
\State \Return  $\Opt_{\rm min}$ ($\Opt_{\rm Cmin}=\Par\Col_{\rm min}$)
\end{algorithmic}
 \label{algo:optmin}
\end{algorithm}

\clearpage
\section{Further details and figures of further numerical experiments}\label{sec:numerics_setup}

\subsection{Setup} 

We compare the following adjustment sets (see definitions in Section~\ref{sec:Oset_def}):
\begin{itemize}
 \item $\Opt$
 \item Adjust
 \item $\Optcmin$
 \item $\Optmin$
 \item Adjust$_{\rm Xmin}$
 \item Adjust$_{\rm min}$
\end{itemize} 
To investigate the applicability of different estimators, we use above adjustment sets together with the following estimators from \texttt{sklearn} (version 0.24.2) and the \texttt{doubleml} (version 0.4.0) package (see instantiated class for parameters):
\begin{itemize}
 \item Linear ordinary least squares (LinReg) regressor \texttt{LinearRegression()}
 \item $k$-nearest-neighbor (kNN) regressor \texttt{KNeighborsRegressor(n\_neighbors=3)}
 \item Multilayer perceptron (MLP) regressor \texttt{MLPRegressor(max\_iter=2000)}
 \item Random forest (RF) \citep{breiman2001random} regressor \texttt{RandomForestRegressor()}
 \item Double machine learning for partially linear regression models (DML) \citep{chernozhukov2018double} \texttt{DoubleMLPLR(data, ml\_g, ml\_m)} from \texttt{doubleml} with \texttt{ml\_g=ml\_g=MLPRegressor(max\_iter=2000)} from \texttt{sklearn} 
\end{itemize} 
Sklearn \citep{Pedregosa2011} and \texttt{doubleml} \citep{DoubleML2021Python} are both available under an MIT license.

As data generating processes we consider linear and nonlinear experiments generated with the following generalized additive model:
\begin{align} \label{eq:numericalmodel}
V^j &= \textstyle{\sum_i} c_i f_{i}(V^i) + \eta^j\quad \text{for}\quad j\in\{1,\ldots,\tilde{N}\}\,.
\end{align}
To generate a structural causal model among  $\tilde{N}$ variables we randomly choose $L$ links whose functional dependencies are linear for linear experiments and one half is $f_i(x)=(1+5 x e^{-x^2/20})x$ for nonlinear experiments. Coefficients $c_i$ are drawn uniformly from $\pm[0.1, 2]$. For linear experiments we use normal noise $\eta^j \sim \mathcal{N}(0,\sigma^2)$ and, in addition, for nonlinear models $\frac{1}{3}$ of the noise terms is Weibull-distributed, both with standard deviation $\sigma$ drawn uniformly from $[0.5, 2]$. From the $\tilde{N}$ variables of each dataset we randomly choose a fraction $\lambda$ as unobserved and denote the number of observed variables as $N$. For each combination of $N\in\{5, 10, 15, 20\}$, $L\in\{2\tilde{N}, 3\tilde{N}\}$, and $\lambda\in\{30\%, 40\%, 50\%\}$ we randomly create a structural causal model and then randomly pick an observed pair $(X=V^i,Y=V^j)$ connected by a causal path, set $\Sel=\emptyset$, and consider the intervention $do(V^i = V^i+1=x)$ relative to the unperturbed data ($x'$) as ground truth, which corresponds to the linear regression coefficient in the linear case. We further assert that the following criteria hold: (1)~the effect is identifiable, (2)~the minimal adjustment cardinality is $|\Adj_{\rm min}(X,Y)|>0$, and (3)~the (absolute) causal effect is $\geq10^{-3}$ to make sure that Faithfulness holds (if these criteria cannot be fulfilled, another model is generated). We create 500 models for each combination of $N, L, \lambda$.
Surprisingly, among in total $12{,}000$ randomly created configurations 93\% fulfill the optimality conditions in Thm.~\ref{thm:graph_optimality}. This may indicate that also in many real-world scenarios graphical optimality actually holds. Here we do not consider the effect of a selected conditioning variable $\Sel$ since it would have a similar effect on all methods considered.

For the considered graphs the computation time to construct adjustment sets is very short and arguably negligible to the actual cost of fitting methods that use these adjustment sets.
The results were evaluated on Intel Xeon Platinum 8260.

\clearpage
\subsection{Figures for linear least squares estimator} \label{sec:linear_appendix}

\begin{figure*}[h]  
\centering
\includegraphics[width=1.\linewidth]{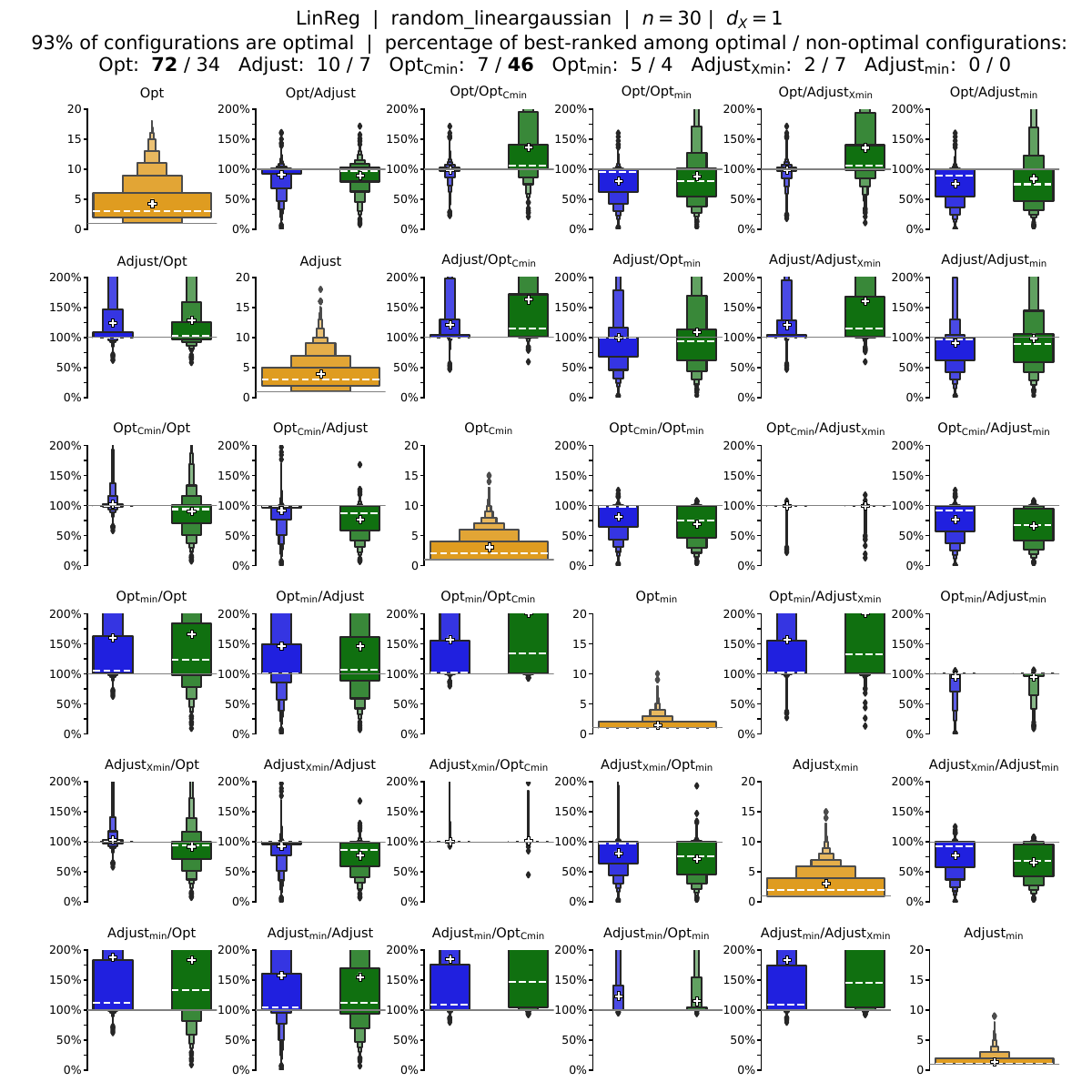}%
\caption{
Results of linear experiments with linear estimator and sample size $n=30$. The diagonal depicts letter-value plots \citep{Hofmann2017} of adjustment set cardinalities and the off-diagonal shows pairs of RMSE ratios for all combinations of ($\Opt$, Adjust, $\Optcmin$, $\Optmin$, Adjust$_{\rm Xmin}$, Adjust$_{\rm min}$) for optimal configurations (left in blue) and non-optimal configurations (right in green). Values above 200\% are not shown. The dashed horizontal line denotes the median of the RMSE ratios, and the white plus their average. The letter-value plots are interpreted as follows: The largest box shows the 25\%--75\% range. The next smaller box above (below) shows the 75\%--87.5\% (12.5\%--25\%) range and so forth.
The numbers on best-ranked methods at the top indicate the percentage of the $12{,}000$ randomly created configurations where the method had the lowest variance. The highest percentage is marked in bold. Note that the highest ranked method may outperform others only by a small margin. The results in the letter-value plots provide a more quantitative picture. See also Fig.~\ref{fig:linear_cardinality_lineargaussian} where the ranks are further distinguished by the $\Opt$-set cardinality.}
\label{fig:experiments_linear_30}
\end{figure*}

\clearpage
\begin{figure*}[h]  
\centering
\includegraphics[width=1.\linewidth]{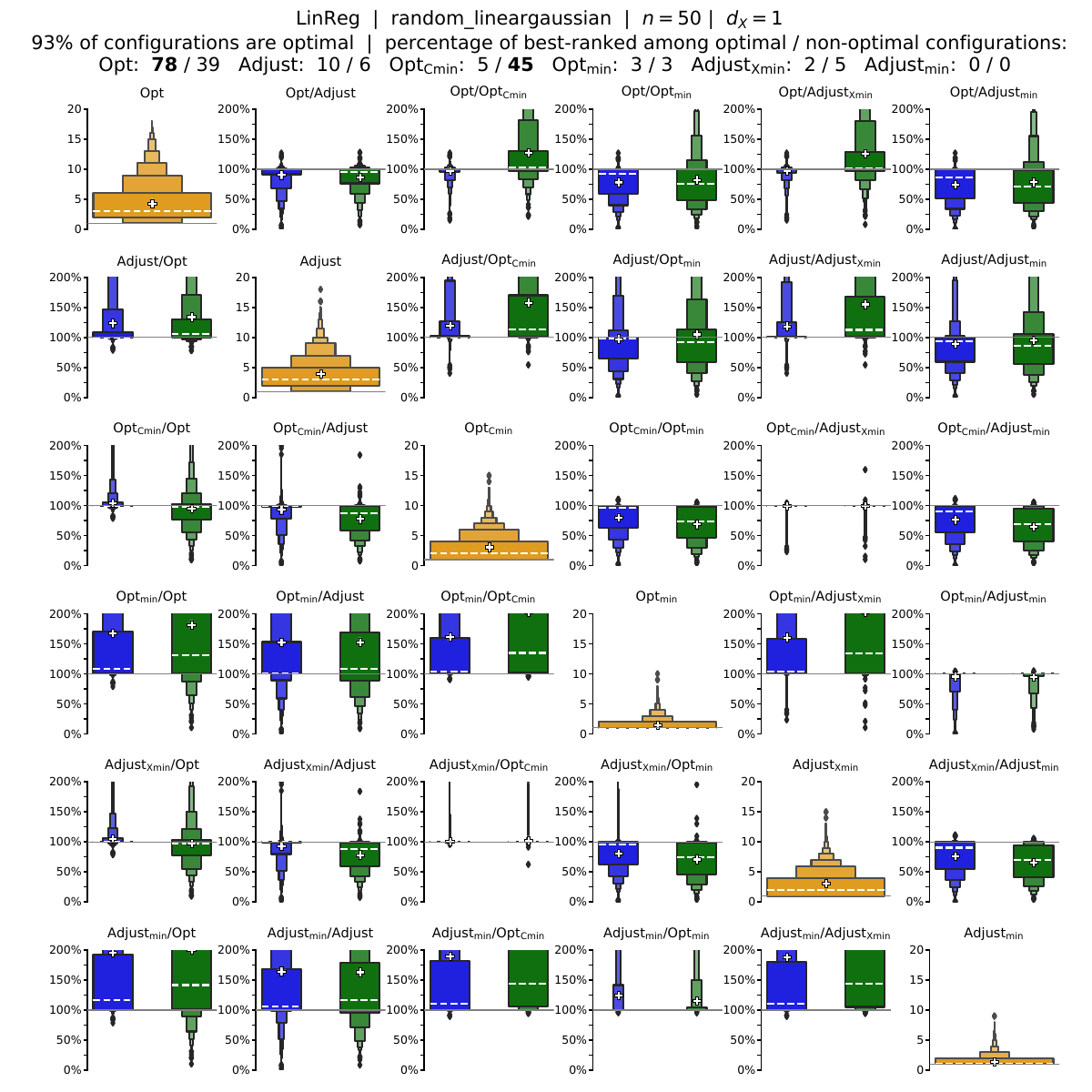}%
\caption{
As in Fig.~\ref{fig:experiments_linear_30} but for $n=50$.}
\label{fig:experiments_linear_50}
\end{figure*}

\clearpage
\begin{figure*}[h]  
\centering
\includegraphics[width=1.\linewidth]{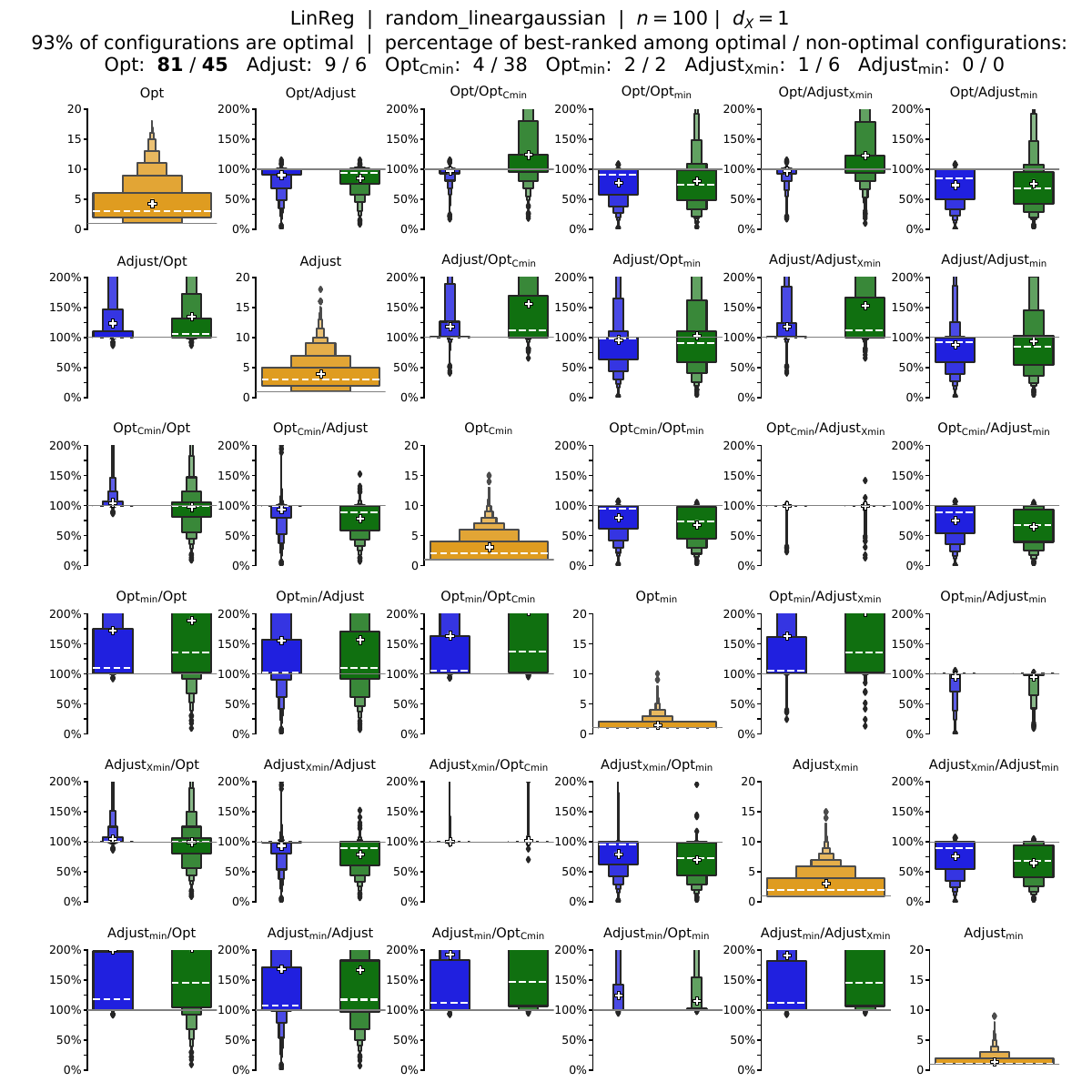}%
\caption{ 
As in Fig.~\ref{fig:experiments_linear_30} but for $n=100$.}
\label{fig:experiments_linear_100}
\end{figure*}

\clearpage
\begin{figure*}[h]  
\centering
\includegraphics[width=1.\linewidth]{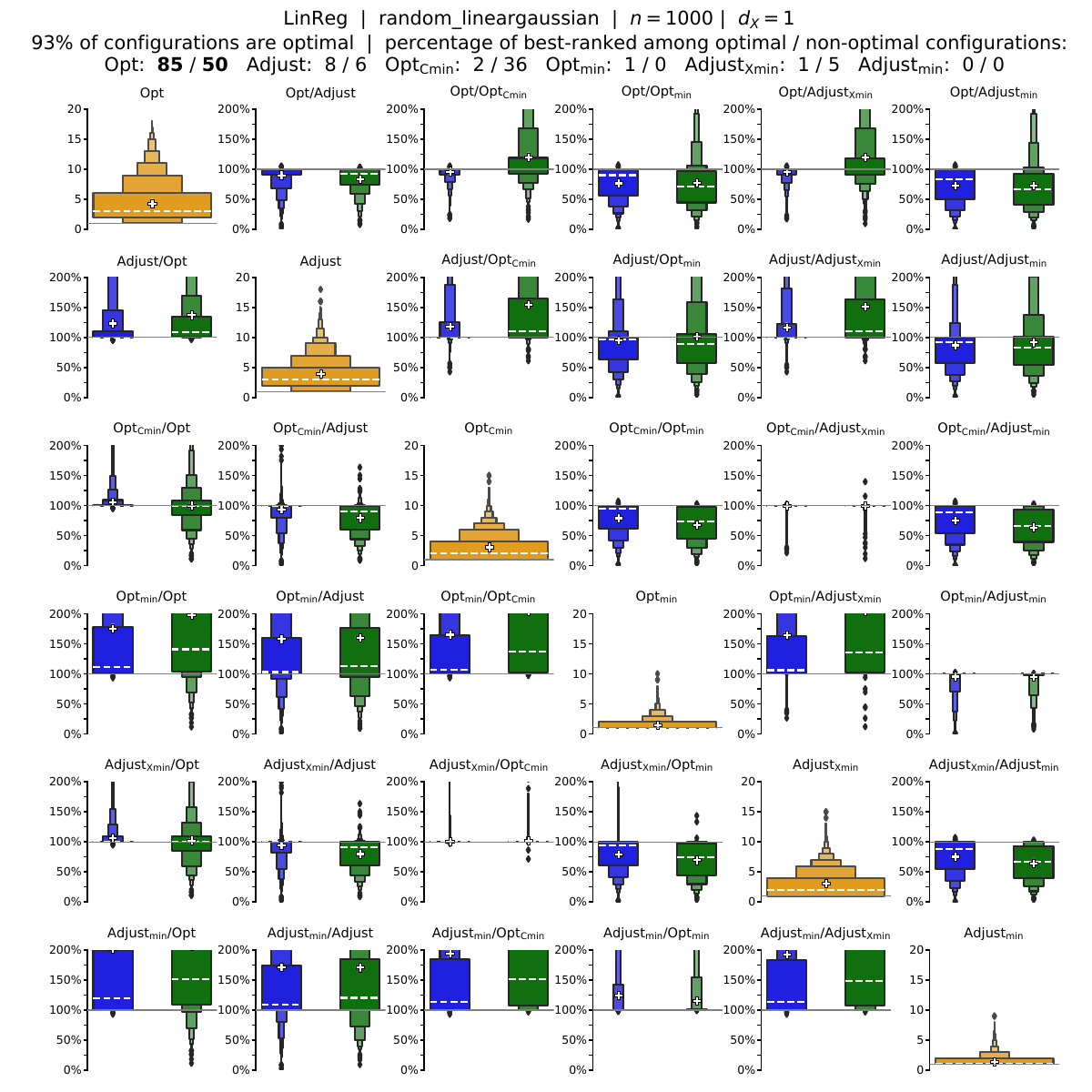}%
\caption{
As in Fig.~\ref{fig:experiments_linear_30} but for $n=1000$.}
\label{fig:experiments_linear_1000}
\end{figure*}

\clearpage
\begin{figure*}[h]  
\centering
\includegraphics[width=1.\linewidth]{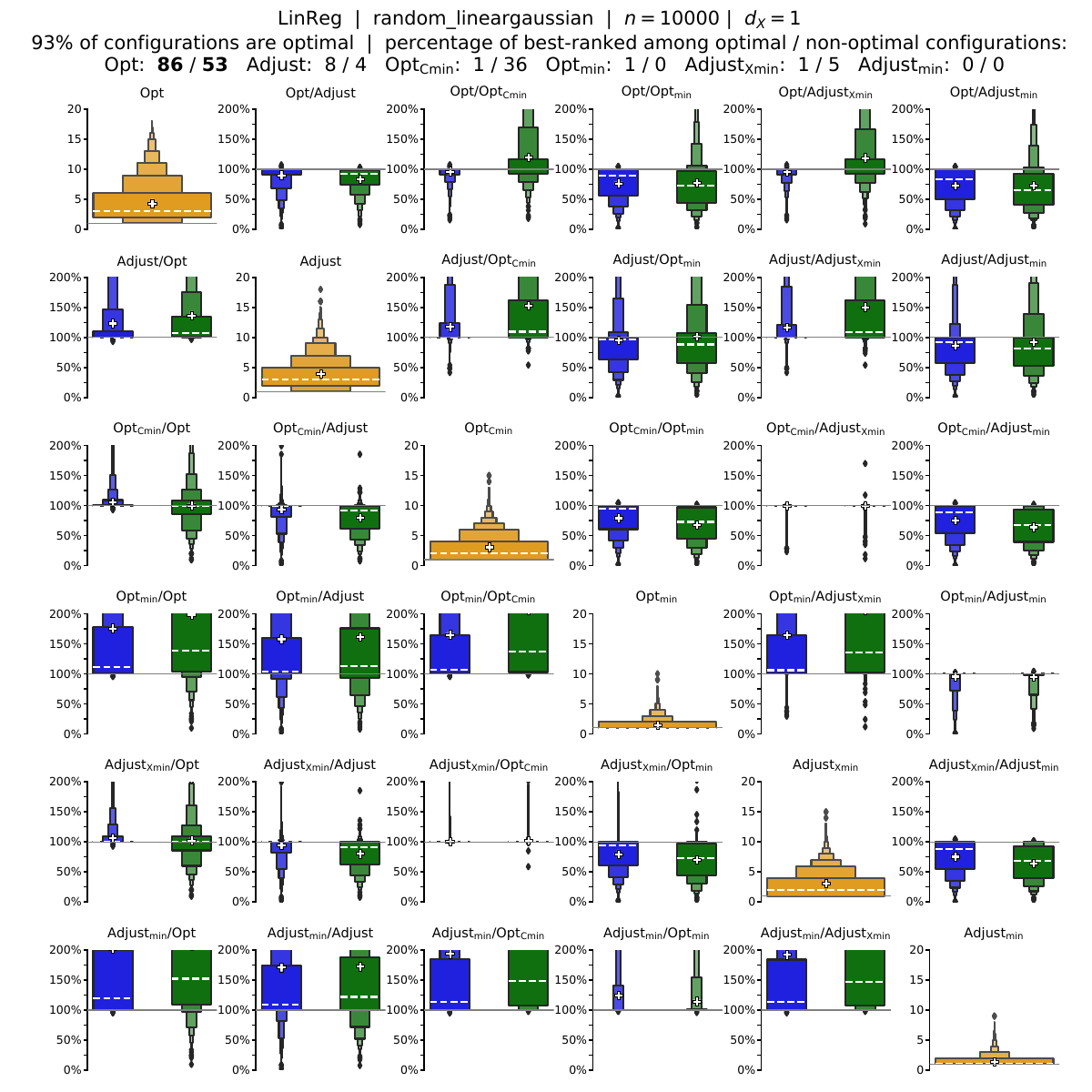}%
\caption{
As in Fig.~\ref{fig:experiments_linear_30} but for $n=10000$.}
\label{fig:experiments_linear_10000}
\end{figure*}

\clearpage
\begin{figure*}[h]  
\centering
\includegraphics[width=.85\linewidth]{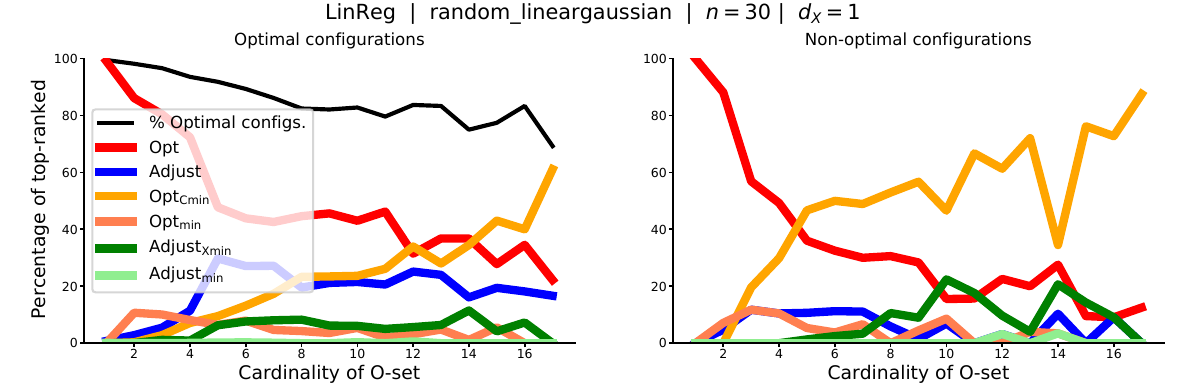}
\includegraphics[width=.85\linewidth]{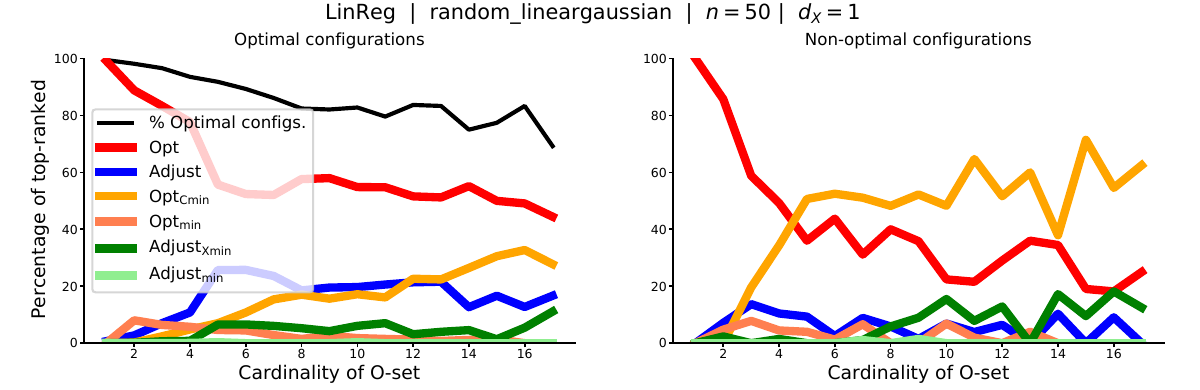}
\includegraphics[width=.85\linewidth]{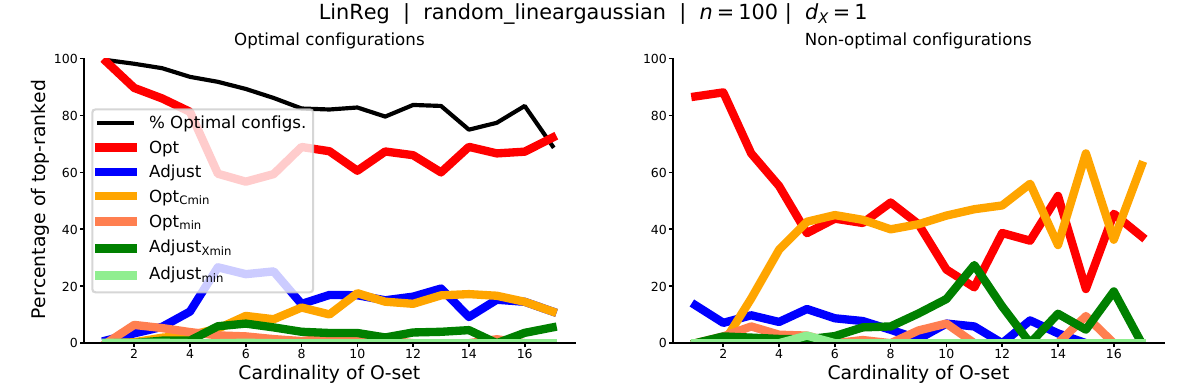}
\includegraphics[width=.85\linewidth]{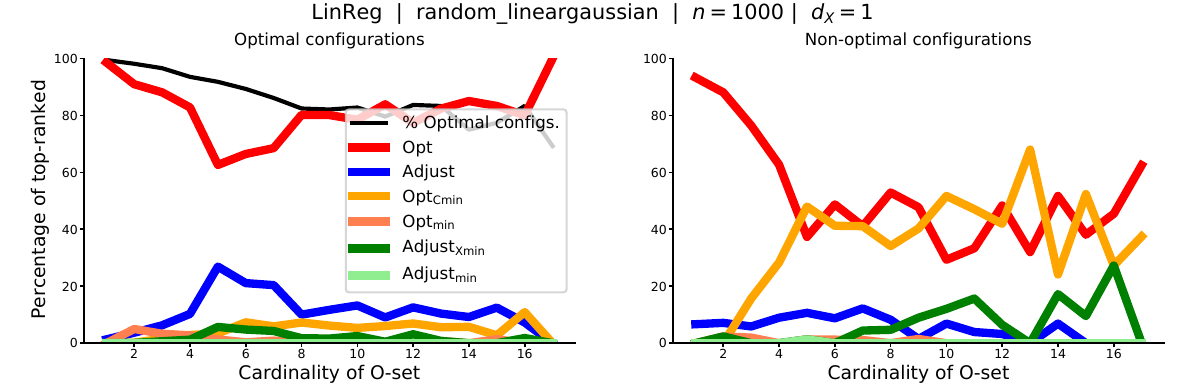}
\includegraphics[width=.85\linewidth]{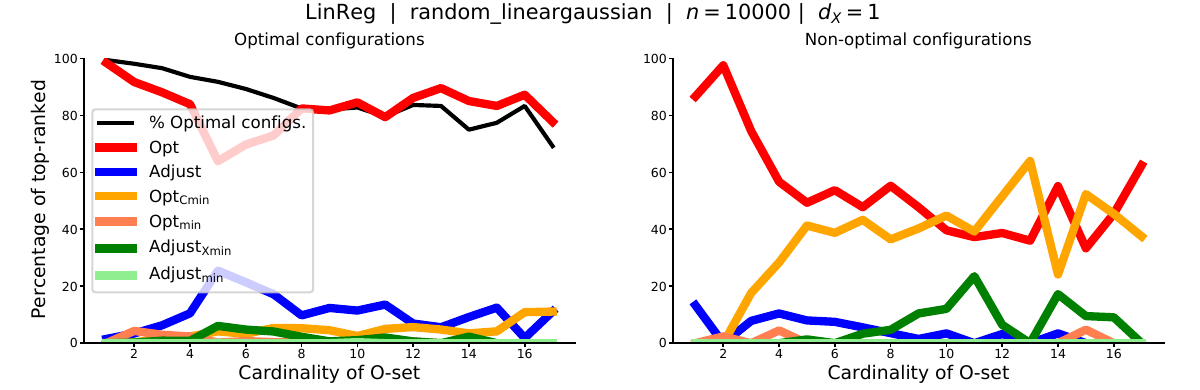}
\caption{
Percentage of configurations where each method has the lowest variance for linear experiments, stratified by the cardinality of the $\Opt$-set ($x$-axis) for $n=30$ (top) to $n=10{,}000$ (bottom).}
\label{fig:linear_cardinality_lineargaussian}
\end{figure*}

\clearpage
\subsection{Figures for non-parametric estimators} \label{sec:nonparametric_appendix}


\clearpage
\begin{figure*}[h]  
\centering
\includegraphics[width=1.\linewidth]{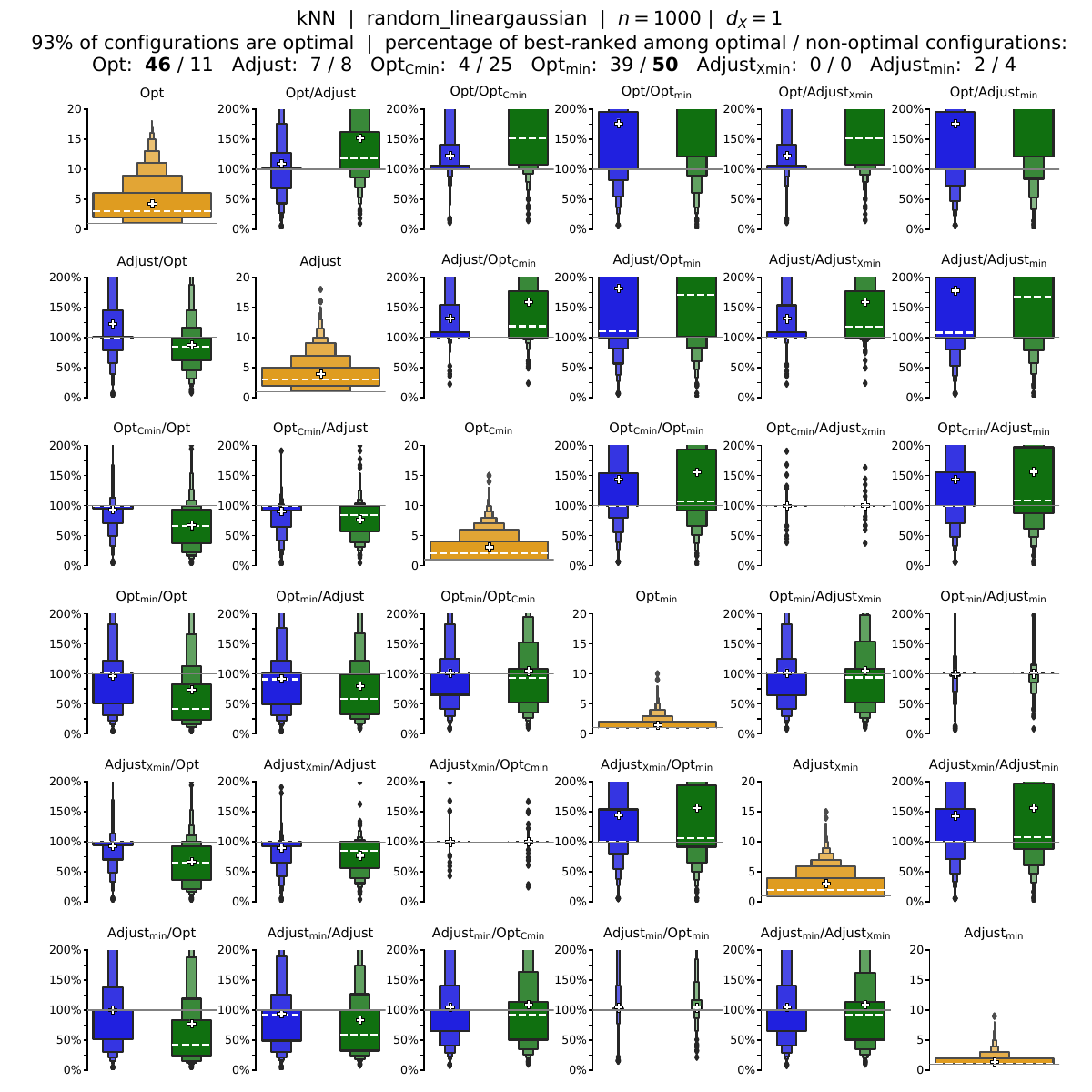}%
\caption{
As in Fig.~\ref{fig:experiments_linear_30} but with kNN estimator ($k=3$) and $n=1000$. See also Figs.~\ref{fig:cardinality_lineargaussian},\ref{fig:cardinality_nonlinearmixed} where the ranks are further distinguished by the $\Opt$-set cardinality.}
\label{fig:experiments_knn_1000}
\end{figure*}

\clearpage
\begin{figure*}[h]  
\centering
\includegraphics[width=1.\linewidth]{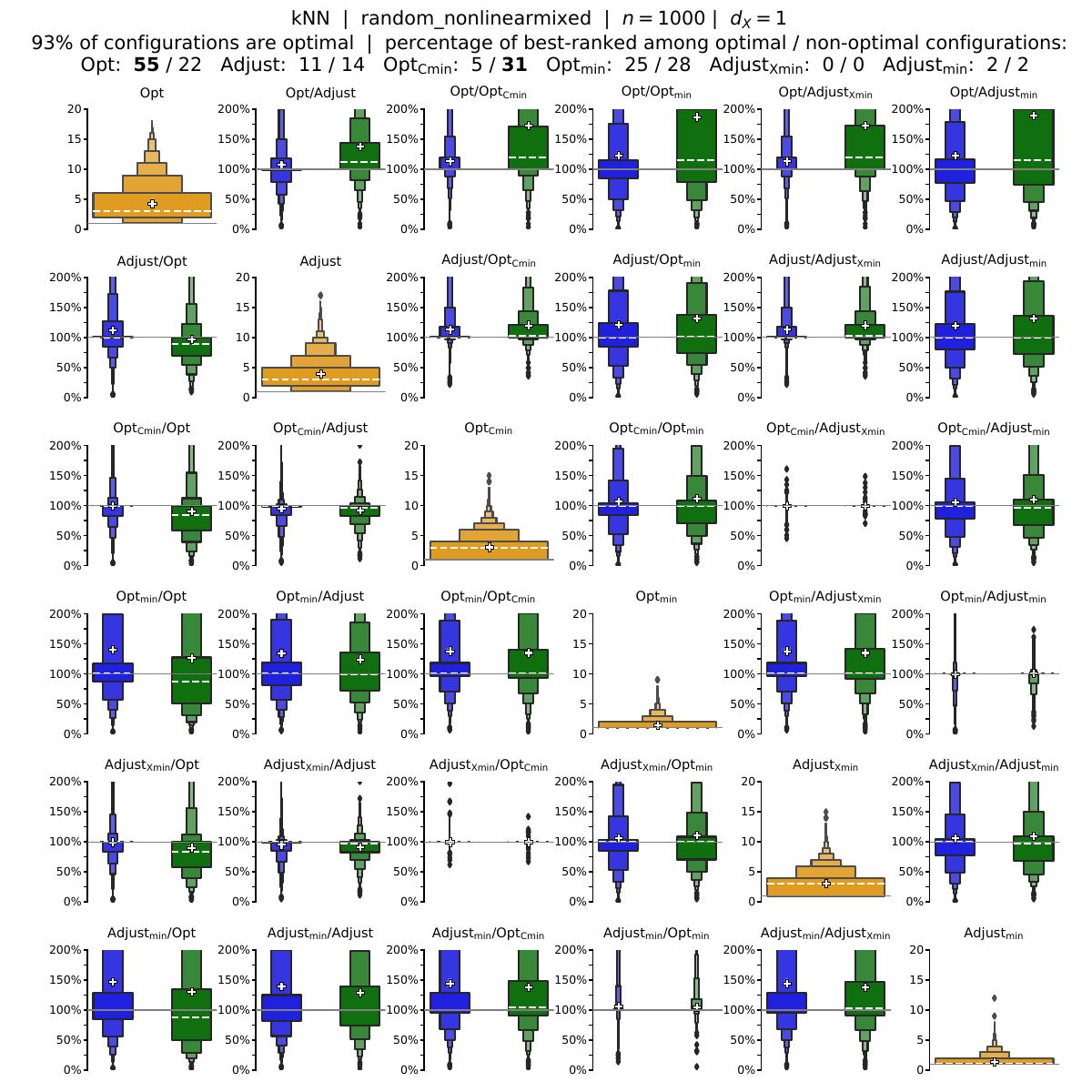}%
\caption{
As in Fig.~\ref{fig:experiments_linear_30} but for kNN estimator ($k=3$), the nonlinear model, and $n=1000$.}
\label{fig:experiments_knn_1000_nonlinear}
\end{figure*}

\clearpage
\begin{figure*}[h]  
\centering
\includegraphics[width=1.\linewidth]{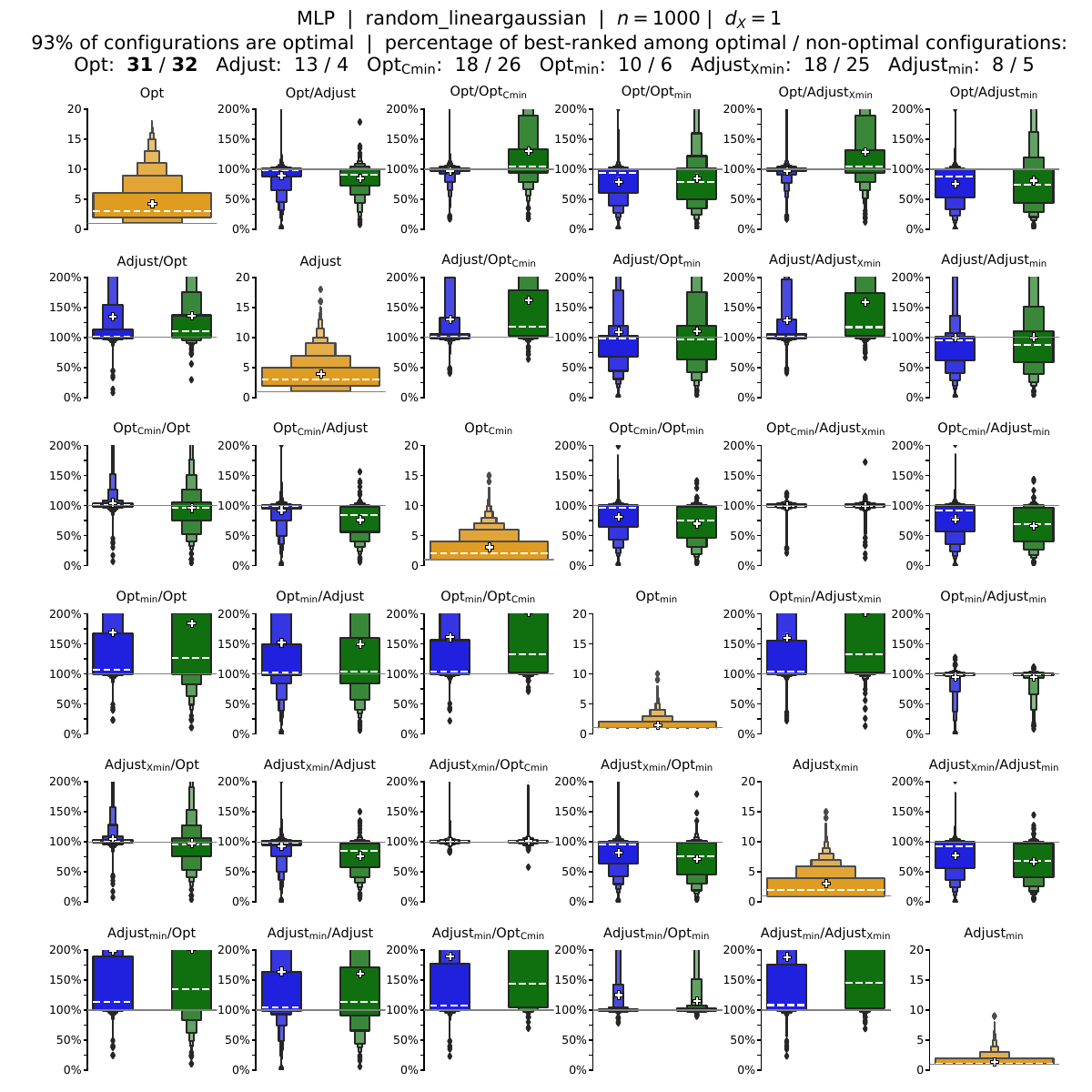}%
\caption{
As in Fig.~\ref{fig:experiments_linear_30} but for MLP estimator and $n=1000$.}
\label{fig:experiments_mlp_1000}
\end{figure*}

\clearpage
\begin{figure*}[h]  
\centering
\includegraphics[width=1.\linewidth]{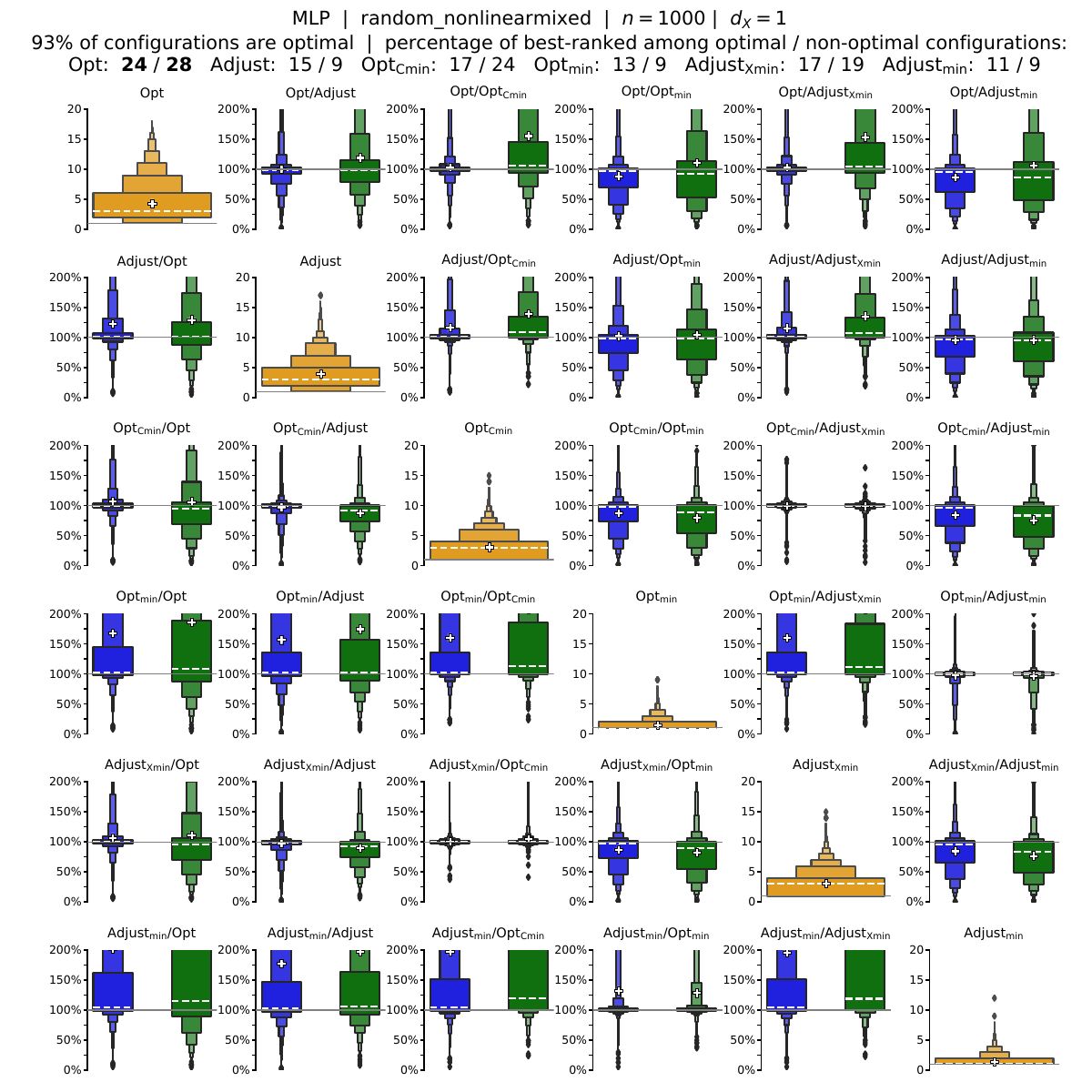}%
\caption{
As in Fig.~\ref{fig:experiments_linear_30} but for MLP estimator, the nonlinear model, and $n=1000$.}
\label{fig:experiments_mlp_1000_nonlinear}
\end{figure*}

\clearpage
\begin{figure*}[h]  
\centering
\includegraphics[width=1.\linewidth]{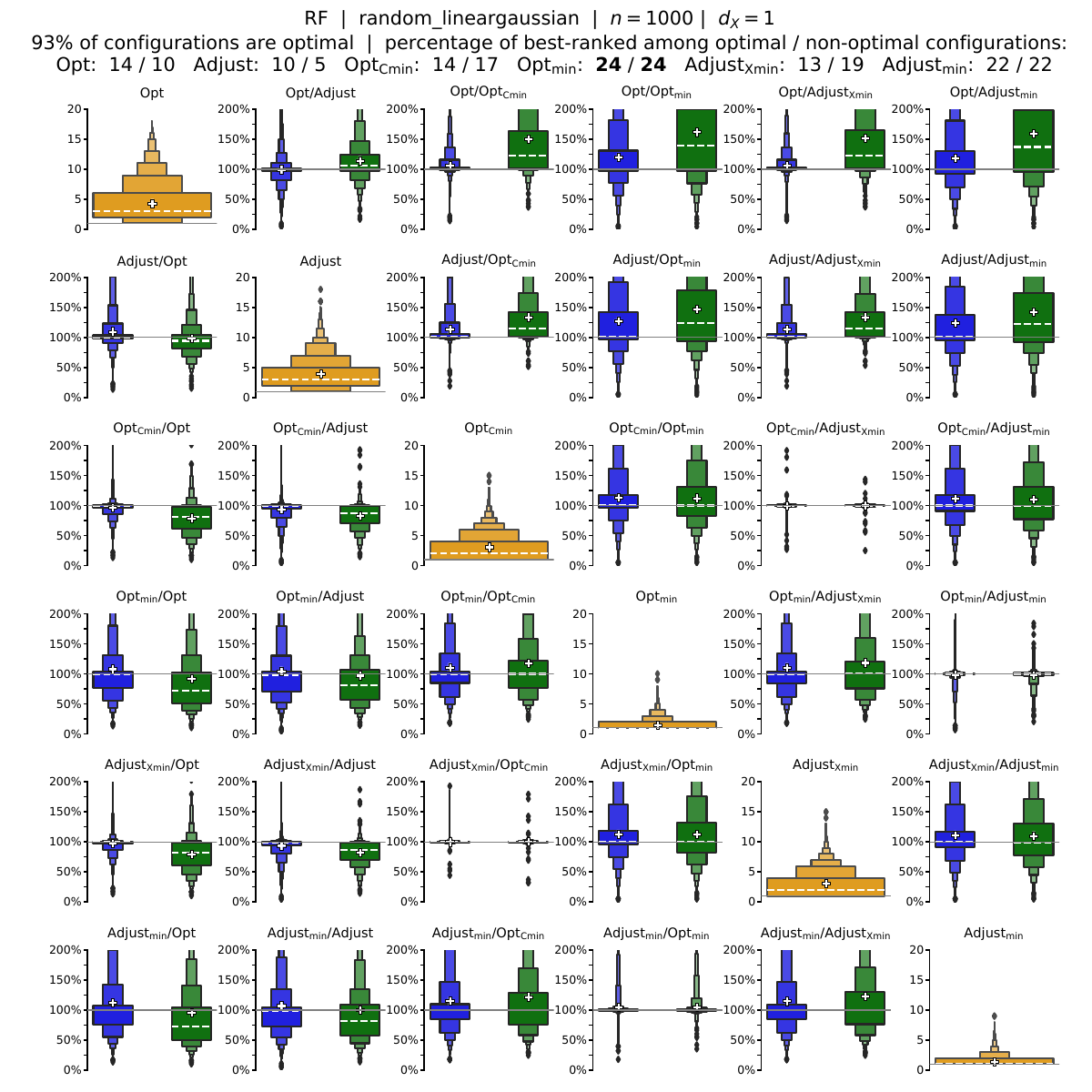}%
\caption{
As in Fig.~\ref{fig:experiments_linear_30} but for RF estimator and $n=1000$.}
\label{fig:experiments_rf_1000}
\end{figure*}

\clearpage
\begin{figure*}[h]  
\centering
\includegraphics[width=1.\linewidth]{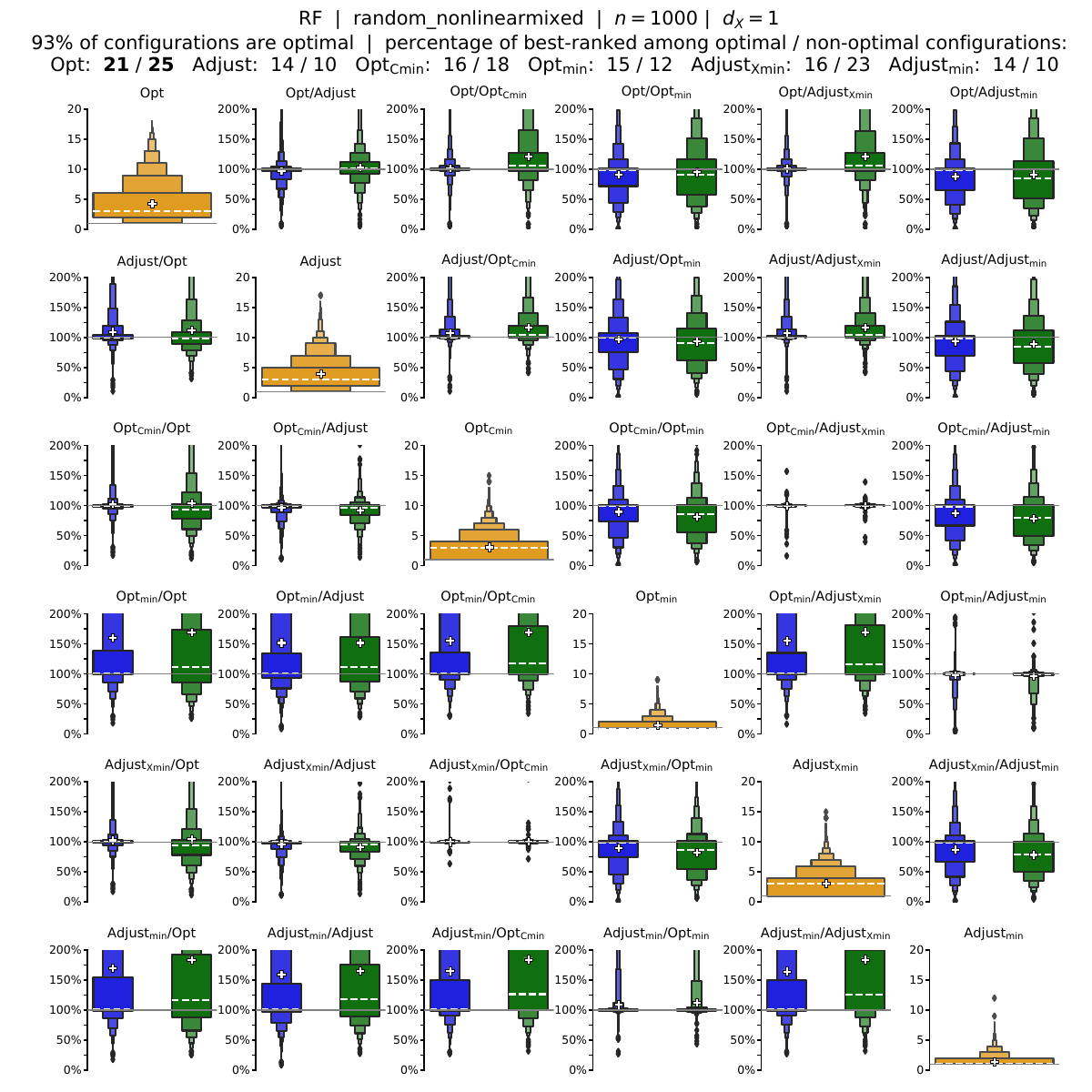}%
\caption{
As in Fig.~\ref{fig:experiments_linear_30} but for RF estimator, the nonlinear model, and $n=1000$.}
\label{fig:experiments_rf_1000_nonlinear}
\end{figure*}

\clearpage
\begin{figure*}[h]  
\centering
\includegraphics[width=1.\linewidth]{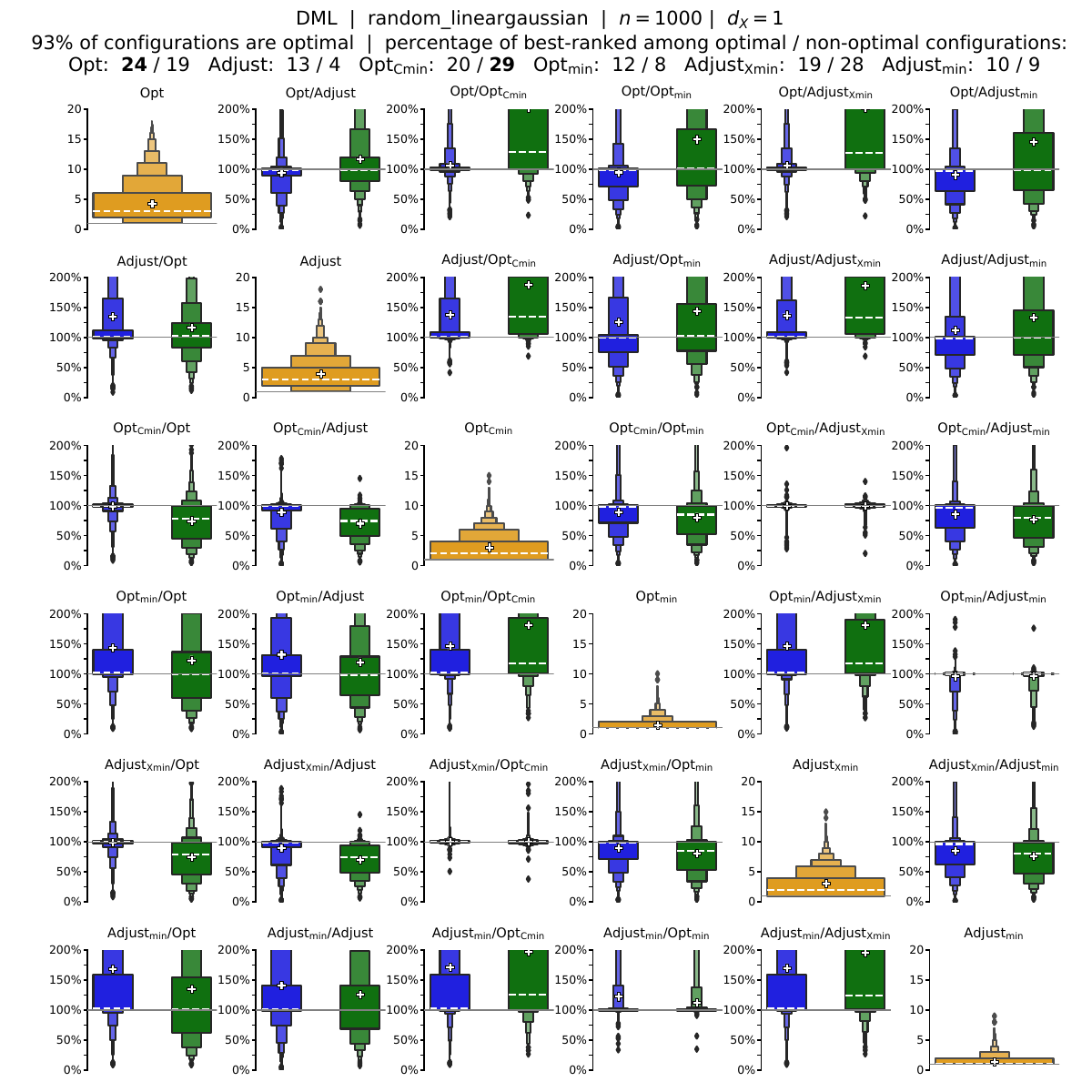}%
\caption{
As in Fig.~\ref{fig:experiments_linear_30} but for DML estimator and $n=1000$.}
\label{fig:experiments_doubleml_1000}
\end{figure*}





\clearpage
\begin{figure*}[h]  
\centering
\includegraphics[width=.85\linewidth]{figures/methods-allparasminimizedmatrixnewcardinality-1000-1-linear_cardinality.pdf}
\includegraphics[width=.85\linewidth]{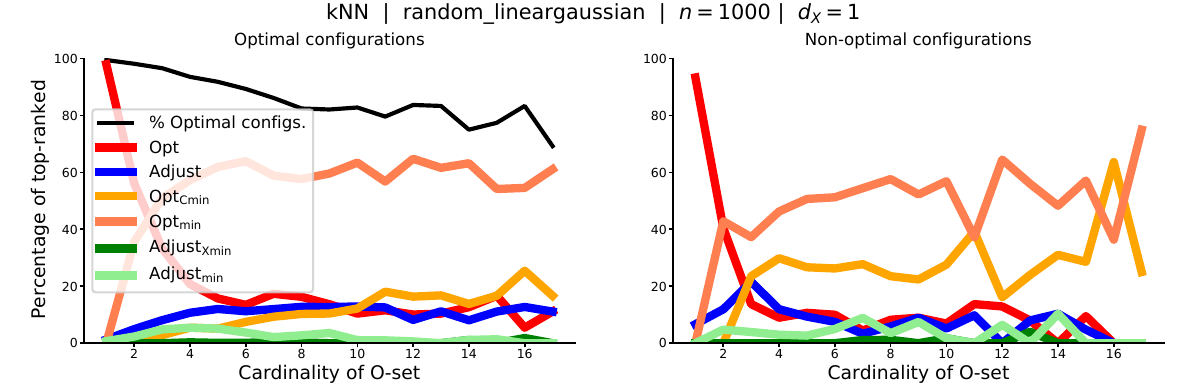}
\includegraphics[width=.85\linewidth]{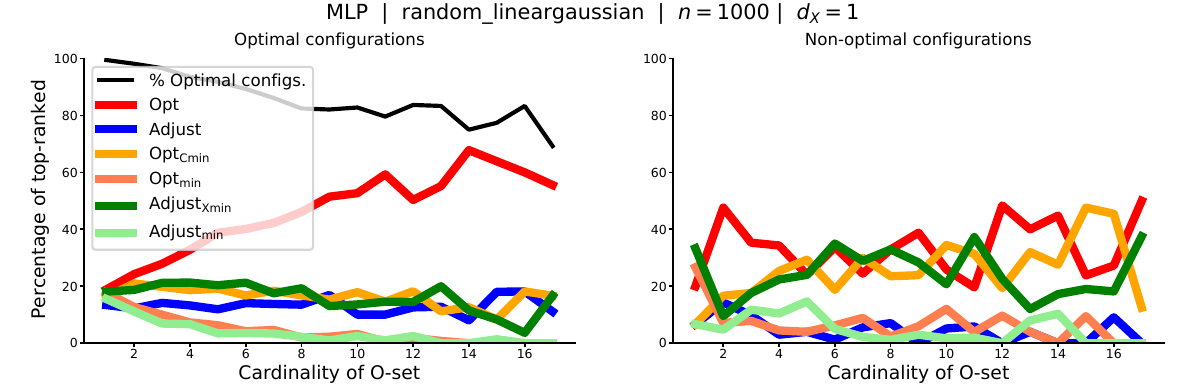}
\includegraphics[width=.85\linewidth]{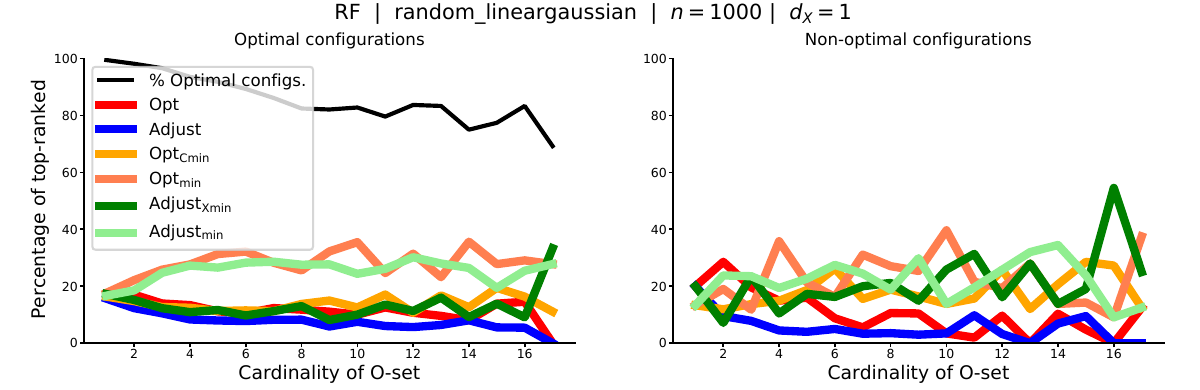}
\includegraphics[width=.85\linewidth]{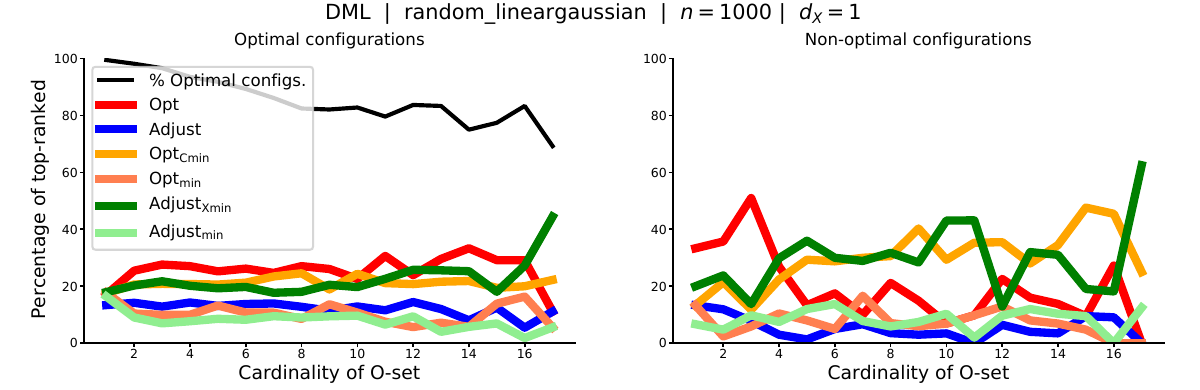}
\caption{As in Fig.~\ref{fig:linear_cardinality_lineargaussian} but including non-parametric estimators for $n=1000$.}
\label{fig:cardinality_lineargaussian}
\end{figure*}

\clearpage
\begin{figure*}[h]  
\centering
\phantom{
\includegraphics[width=.85\linewidth]{figures/methods-allparasminimizedmatrixnewcardinality-1000-1-linear_cardinality.pdf}}
\includegraphics[width=.85\linewidth]{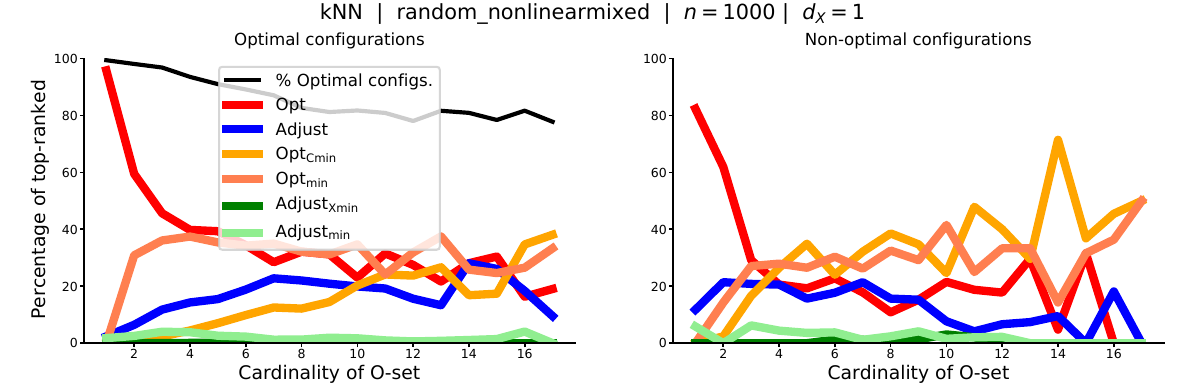}
\includegraphics[width=.85\linewidth]{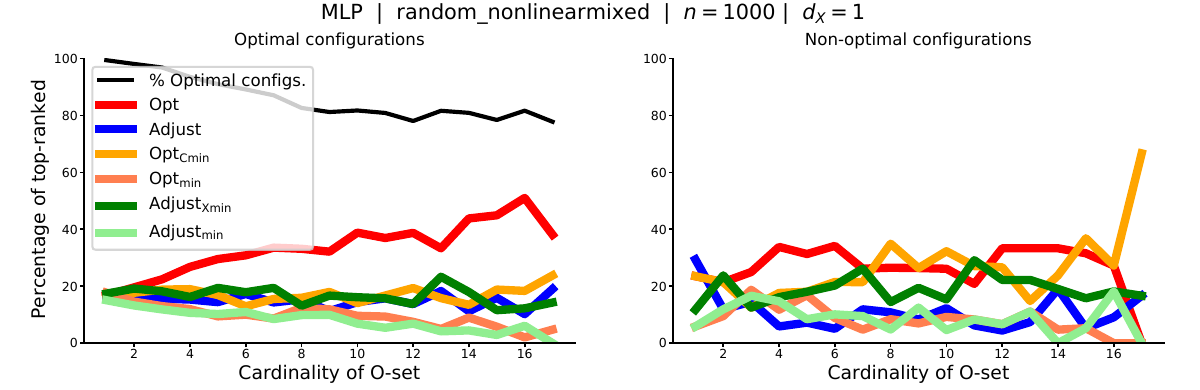}
\includegraphics[width=.85\linewidth]{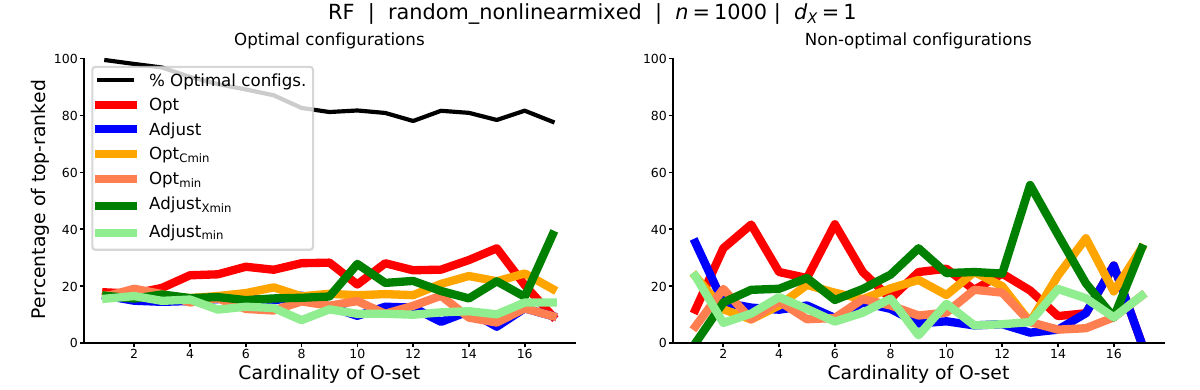}
\phantom{\includegraphics[width=.85\linewidth]{figures/methods-allparasminimizedmatrixnewcardinality-1000-1-doubleml_cardinality.pdf}}
\caption{
As in Fig.~\ref{fig:cardinality_lineargaussian} but for nonlinear experiments.}
\label{fig:cardinality_nonlinearmixed}
\end{figure*}


\begin{thebibliography}{22}
\providecommand{\natexlab}[1]{#1}
\providecommand{\url}[1]{\texttt{#1}}
\expandafter\ifx\csname urlstyle\endcsname\relax
  \providecommand{\doi}[1]{doi: #1}\else
  \providecommand{\doi}{doi: \begingroup \urlstyle{rm}\Url}\fi

\bibitem[Bach et~al.(2021)Bach, Chernozhukov, Kurz, and
  Spindler]{DoubleML2021Python}
Philipp Bach, Victor Chernozhukov, Malte~S. Kurz, and Martin Spindler.
\newblock {DoubleML} -- {A}n object-oriented implementation of double machine
  learning in {P}ython, 2021.
\newblock arXiv:\href{https://arxiv.org/abs/2104.03220}{2104.03220} [stat.ML].

\bibitem[Breiman(2001)]{breiman2001random}
Leo Breiman.
\newblock Random forests.
\newblock \emph{Machine learning}, 45\penalty0 (1):\penalty0 5--32, 2001.

\bibitem[Chernozhukov et~al.(2018)Chernozhukov, Chetverikov, Demirer, Duflo,
  Hansen, Newey, and Robins]{chernozhukov2018double}
Victor Chernozhukov, Denis Chetverikov, Mert Demirer, Esther Duflo, Christian
  Hansen, Whitney Newey, and James Robins.
\newblock Double/debiased machine learning for treatment and structural
  parameters.
\newblock \emph{The Econometrics Journal}, 21:\penalty0 C1--C68, 2018.

\bibitem[Cover and Thomas(2006)]{Cover2006}
Thomas~M. Cover and Joy~A. Thomas.
\newblock \emph{{Elements of Information Theory}}.
\newblock John Wiley {\&} Sons, Hoboken, 2006.

\bibitem[Evans and Richardson(2014)]{evans2014markovian}
Robin~J Evans and Thomas~S Richardson.
\newblock Markovian acyclic directed mixed graphs for discrete data.
\newblock \emph{The Annals of Statistics}, pages 1452--1482, 2014.

\bibitem[Henckel et~al.(2019)Henckel, Perkovi{\'c}, and
  Maathuis]{henckel2019graphical}
Leonard Henckel, Emilija Perkovi{\'c}, and Marloes~H Maathuis.
\newblock Graphical criteria for efficient total effect estimation via
  adjustment in causal linear models.
\newblock \emph{arXiv preprint arXiv:1907.02435}, 2019.

\bibitem[Hofmann et~al.(2017)Hofmann, Wickham, and Kafadar]{Hofmann2017}
Heike Hofmann, Hadley Wickham, and Karen Kafadar.
\newblock Letter-value plots: Boxplots for large data.
\newblock \emph{Journal of Computational and Graphical Statistics}, 26\penalty0
  (3):\penalty0 469--477, 2017.

\bibitem[Kuroki and Cai(2004)]{kuroki2004selection}
Manabu Kuroki and Zhihong Cai.
\newblock Selection of identifiability criteria for total effects by using path
  diagrams.
\newblock In \emph{Proceedings of the 20th conference on Uncertainty in
  artificial intelligence}, pages 333--340, 2004.

\bibitem[Kuroki and Miyakawa(2003)]{kuroki2003covariate}
Manabu Kuroki and Masami Miyakawa.
\newblock Covariate selection for estimating the causal effect of control plans
  by using causal diagrams.
\newblock \emph{Journal of the Royal Statistical Society: Series B (Statistical
  Methodology)}, 65\penalty0 (1):\penalty0 209--222, 2003.

\bibitem[Maathuis et~al.(2009)Maathuis, Kalisch, B{\"u}hlmann,
  et~al.]{maathuis2009estimating}
Marloes~H Maathuis, Markus Kalisch, Peter B{\"u}hlmann, et~al.
\newblock Estimating high-dimensional intervention effects from observational
  data.
\newblock \emph{The Annals of Statistics}, 37\penalty0 (6A):\penalty0
  3133--3164, 2009.

\bibitem[Maathuis et~al.(2010)Maathuis, Colombo, Kalisch, and
  B{\"{u}}hlmann]{Maathuis2010}
Marloes~H Maathuis, Diego Colombo, Markus Kalisch, and Peter B{\"{u}}hlmann.
\newblock {Predicting causal effects in large-scale systems from observational
  data.}
\newblock \emph{Nature Methods}, 7:\penalty0 247--8, 2010.

\bibitem[Pearl(1993)]{pearl1993}
Judea Pearl.
\newblock [bayesian analysis in expert systems]: Comment: Graphical models,
  causality and intervention.
\newblock \emph{Statistical Science}, 8\penalty0 (3):\penalty0 266--269, 08
  1993.
\newblock \doi{10.1214/ss/1177010894}.

\bibitem[Pearl(2009)]{Pearl2000}
Judea Pearl.
\newblock \emph{Causality: Models, reasoning, and inference.}
\newblock Cambridge University Press, 2009.

\bibitem[Pedregosa et~al.(2011)Pedregosa, Varoquaux, Gramfort, Michel, Thirion,
  Grisel, Blondel, Prettenhofer, Weiss, Dubourg, Vanderplas, Passos,
  Cournapeau, Brucher, Perrot, and Duchesnay]{Pedregosa2011}
Fabian Pedregosa, Ga{\"{e}}l Varoquaux, Alexandre Gramfort, Vincent Michel,
  Bertrand Thirion, Olivier Grisel, Mathieu Blondel, Peter Prettenhofer, Ron
  Weiss, Vincent Dubourg, Jake Vanderplas, Alexandre Passos, David Cournapeau,
  Matthieu Brucher, Matthieu Perrot, and {\'{E}}douard Duchesnay.
\newblock {Scikit-learn: Machine Learning in Python}.
\newblock \emph{Journal of Machine Learning Research}, 12\penalty0
  (Oct):\penalty0 2825--2830, 2011.
\newblock ISSN ISSN 1533-7928.

\bibitem[Perkovi{\'c} et~al.(2015)Perkovi{\'c}, Textor, Kalisch, and
  Maathuis]{perkovic2015complete}
Emilija Perkovi{\'c}, Johannes Textor, Markus Kalisch, and Marloes~H Maathuis.
\newblock A complete generalized adjustment criterion.
\newblock In \emph{Uncertainty in Artificial Intelligence-Proceedings of the
  Thirty-First Conference (2015)}, pages 682--691. AUAI Press, 2015.

\bibitem[Perkovi{\'c} et~al.(2018)Perkovi{\'c}, Textor, and
  Kalisch]{perkovic2018complete}
Emilija Perkovi{\'c}, Johannes Textor, and Markus Kalisch.
\newblock Complete graphical characterization and construction of adjustment
  sets in markov equivalence classes of ancestral graphs.
\newblock \emph{Journal of Machine Learning Research}, 18:\penalty0 1--62,
  2018.

\bibitem[Richardson and Spirtes(2002)]{richardson2002}
Thomas Richardson and Peter Spirtes.
\newblock Ancestral graph markov models.
\newblock \emph{The Annals of Statistics}, 30\penalty0 (4):\penalty0 962--1030,
  08 2002.

\bibitem[Rotnitzky and Smucler(2019)]{rotnitzky2019efficient}
Andrea Rotnitzky and Ezequiel Smucler.
\newblock Efficient adjustment sets for population average treatment effect
  estimation in non-parametric causal graphical models.
\newblock \emph{arXiv preprint arXiv:1912.00306}, 2019.

\bibitem[Shpitser et~al.(2010)Shpitser, VanderWeele, and
  Robins]{shpitser2010validity}
Ilya Shpitser, Tyler VanderWeele, and James~M Robins.
\newblock On the validity of covariate adjustment for estimating causal
  effects.
\newblock In \emph{Proceedings of the Twenty-Sixth Conference on Uncertainty in
  Artificial Intelligence}, pages 527--536, 2010.

\bibitem[Smucler et~al.(2021)Smucler, Sapienza, and
  Rotnitzky]{smucler2020efficient}
E~Smucler, F~Sapienza, and A~Rotnitzky.
\newblock {Efficient adjustment sets in causal graphical models with hidden
  variables}.
\newblock \emph{Biometrika}, 3 2021.
\newblock \doi{10.1093/biomet/asab018}.

\bibitem[van~der Zander et~al.(2019)van~der Zander, Li{\'s}kiewicz, and
  Textor]{van2019separators}
Benito van~der Zander, Maciej Li{\'s}kiewicz, and Johannes Textor.
\newblock Separators and adjustment sets in causal graphs: Complete criteria
  and an algorithmic framework.
\newblock \emph{Artificial Intelligence}, 270:\penalty0 1--40, 2019.

\bibitem[Witte et~al.(2020)Witte, Henckel, Maathuis, and
  Didelez]{witte2020efficient}
Janine Witte, Leonard Henckel, Marloes~H Maathuis, and Vanessa Didelez.
\newblock On efficient adjustment in causal graphs.
\newblock \emph{Journal of Machine Learning Research}, 21\penalty0
  (246):\penalty0 1--45, 2020.

\end{thebibliography}
\end{document}